\documentclass[DTMColor]{ROB-New}

\usepackage{algorithmic}
\usepackage{algorithm}
\usepackage{amsmath,amsfonts}
\usepackage{array}
\usepackage[caption=false,font=normalsize,labelfont=sf,textfont=sf]{subfig}
\usepackage{csquotes}
\usepackage{graphicx}
\usepackage{stfloats}
\usepackage{textcomp}
\usepackage{url}
\usepackage{verbatim}
\usepackage{xcolor}
\usepackage{appendix}

\begin{document}

\authormark{Yinnan Luo et al.}

\articletype{RESEARCH ARTICLE}

\jnlPage{1}{36}
\jyear{2023}
\jdoi{10.1017/xxxxx}

\title{Hybrid Zero Dynamics Control for Bipedal Walking with a Non-Instantaneous Double Support Phase}

\author[1]{Yinnan Luo\hyperlink{corr}{*}} 
\author[1]{Ulrich J. Römer}
\author[1]{Alexander Dyck}
\author[2]{Marten Zirkel}
\author[2]{Lena Zentner}
\author[1]{Alexander Fidlin}
\address[1]{Institute of Engineering Mechanics, Karlsruhe Institute of Technology (KIT), Karlsruhe, Germany}
\address[2]{Compliant Systems Group, Technische Universität Ilmenau, Ilmenau, Germany}
\address{\hypertarget{corr}{*}Corresponding author. \email{yinnan.luo@kit.edu}}

\received{xx xxx xxx}
\revised{xx xxx xxx}
\accepted{xx xxx xxx}


\abstract{The hybrid zero dynamics control concept for bipedal walking is extended to include a non-instantaneous double support phase. A symmetric robot that consists of five rigid body segments which are connected by four actuated revolute joints is considered. Periodic walking gaits with a constant average walking speed consists of alternating single (SSP) and double support phases (DSP). Hybrid zero dynamics control designs usually assume an instantaneous DSP, which is a severe limitation. The proposed controllers use continuous SSPs and DSPs. Transitions between both phases are modeled as instantaneous events, when the rear leg lifts off at the end of the DSP and the swing leg touches down at the end of the SSP. Due to the fact that the model during the DSP has more actuators (4) than degrees of freedom (3), the system is overactuated. In order to combine it with the underactuated SSP model and then formulate a periodic walking gait, we suggest three controller designs for different applications. One with the underactuated DSP, one with the fully actuated DSP, and one with the overactuated DSP. A numerical optimization is used to generate energy efficient gaits in an offline process. According to the optimization results, artificially creating an underactuated controller for the DSP results in the most efficient gaits. Adding control tasks utilizing the full actuation or overactuation during the DSP significantly improves the gait stability.}

\maketitle

\section{Introduction}\label{labIntro}
	Humanoid walking has obtained a lot of attention from interdisciplinary research communities including medicine, biomechanics and especially robotics. Inspired by nature, the development of bipedal robots and research into their application have advanced significantly in recent years. Due to the complexity of walking mechanisms, many simplified models such as the inverted pendulum \cite{InvertePend} and corresponding control strategies have been investigated \cite{Overview1,Overview2,Overview3}. The gait cycle of walking is commonly divided into two alternating phases: a single support phase (SSP, one leg contacts the ground) and a double support phase (DSP, both legs contact the ground). In fact, human walking gaits contain a DSP of non-negligible duration that makes up more than $20\,\%$ of the entire step according to experimental observations \cite{GaitMeasurement1,GaitMeasurement2}. Based on investigations of the inverted pendulum model in \cite{DSPBio1,DSPBio2}, one purpose of the DSP in human walking is the redirection of the centre of mass velocity during step-to-step transitions while the legs simultaneously perform positive and negative work.
	 
	From the perspective of robot designers, various DSP assumptions have been investigated in real robot systems. For example, controllers based on the zero moment point (ZMP) stability criterion treat the robot system as fully actuated since no relative motion is allowed between the stance leg foot and the ground \cite{ZMP1,ZMP2}. This makes it straightforward to create walking patterns that mimic humans with a continuous DSP, which can even be applied in different scenarios \cite{ZMP_DSP1,ZMP_DSP2,ZMP_DSP3}. As an extension, the contact wrench sum (CWS) criterion \cite{CWS1,CWS2,CWS3} has been proposed as a more promising concept for generating the motion of the fully actuated robots. According to this criterion, the movement is balanced when the CWS provided by the contact forces and torques is equal to the rate of change of the linear and angular momentum with respect to the robot's center of mass. One of the benefits thereof is the agility to realize arbitrary, reasonable motions and complex contact profiles, e.\,g.\ the DSP, during gait planing. Efficient numerical optimization methods, like sequential quadratic programming (SQP) \cite{Numeric3,Numeric4}, are used for generating feasible gaits. A clear trade-off is, however, that the controller requires powerful actuators in every joint, such as the hydraulic actuators in the Atlas robot \cite{CWS2,atlas1}. 
 
    Alternative approaches for control strategies use robot concepts that have far fewer actuators, such as passive walkers \cite{PassivWalk,PassivWalk2}. In this context, the DSP is treated as an impulsive, inelastic impact, that instantaneously redirects the velocity of the centre of mass. Impact losses need to be recovered thorough gravity or tiny contributions of motors in order to maintain a constant average velocity. Similar assumptions in terms of the DSP duration are also incorporated into control strategies that aim on utilizing the system's under-actuation, such as the hybrid zero dynamics (HZD) control which is developed for planar robots with point or curved feet \cite{HZDwalker1,HZDwalker2,HZDwalker3}, and also for 3D robots \cite{HZD_3D1,HZD_3D2,HumanHZD}. Based on this approach, a controller synchronizes each joint angle to a corresponding reference trajectory\footnote{In the related literature, these reference trajectories are also called holonomic virtual constraints. As no conditions in the manner of constraint forces exist in the robot joint, the constraint is enforced by the actuator due to the feedback controller. }. Assuming the control error vanishes, the analysis of the full system dynamics is reduced to the analysis of the zero dynamics---the underactuated degree of freedom. Periodic walking gaits are sequences of SSPs and DSPs that correspond to a limit cycle (stable periodic orbits) of the zero dynamics. After solving these low dimensional dynamics for the periodic orbits, the full system states can be reconstructed using the predefined reference trajectories. A major advantage is that the passive dynamics in the controlled system can be utilized to create highly efficient gaits \cite{OptimalEcoupling1,OptimalEcoupling2,OptimalEcoupling3} via a priori numerical optimization. The objective is to minimize the energy consumption of locomotion which is evaluated by the dimensionless cost of transport $CoT:=E_\mathrm{supp}/(\ell_{\mathrm{step}}\cdot m g)$, i.\,e.\ the supplied energy $E_\mathrm{supp}$ during one total step divided by the step length $\ell_{\mathrm{step}}$ and the weight force $m g$.

	The present manuscript introduces a variety of HZD control strategies with a continuous, non-instantaneous DSP, while HZD control strategies in the literature are almost exclusively based on an instantaneous DSP \cite{HZDwalker1}. The challenge in extending the established HZD control strategies for a non-instantaneous DSP is that the system is overactuated in the DSP, rather than underactuated as in the SSP. Since there are more actuators than degrees of freedom in the DSP, further control objectives can be specified and there is not necessarily any zero dynamics in the DSP. Corresponding extensions have already been proposed in \cite{DSP-Hamed,DSP-Buss}, but for special cases without addressing the general possibilities. In this contribution, we propose three different control concepts which are all designed so that periodic walking gaits of alternating SSPs and DSPs correspond to a limit cycle solution of the combined hybrid zero dynamics: Firstly, two independent virtual inputs are introduced to artificially create an underactuated DSP. The virtual inputs are produced from the simultaneous actuation of all four physical actuators, and are used for tracking the reference trajectory in each of the two independent joints during the DSP. As an extension, the second controller uses another virtual input to influence the stability of the limit cycle during the fully actuated DSP, which is regarded as an additional control objective. The third controller utilizes all four physical actuators not only to improve the stability, but also to enforce the collinearity of the contact forces on both feet. We expect this to be advantageous in experimental applications for validating the simulation results on a real robot prototype. 
	
	The paper is organized as follows: In section \ref{labSectionRobot}, models for the single and double support phases and for the transitions between them are derived. In section \ref{labSectionControl}, control strategies for the corresponding gaits are developed. In section \ref{labSectionNumeric}, the optimization process to generate stable periodic gaits of the controlled system is introduced. In section \ref{labSectionResult}, an efficiency study is carried out based on these optimizations. Furthermore, the developed controllers are validated via simulations. The paper ends with the conclusion in section \ref{labSectionConclusion}.

\section{Robot Model}\label{labSectionRobot}
\begin{figure}[ht]
	\centering
	\includegraphics[width=0.65\linewidth]{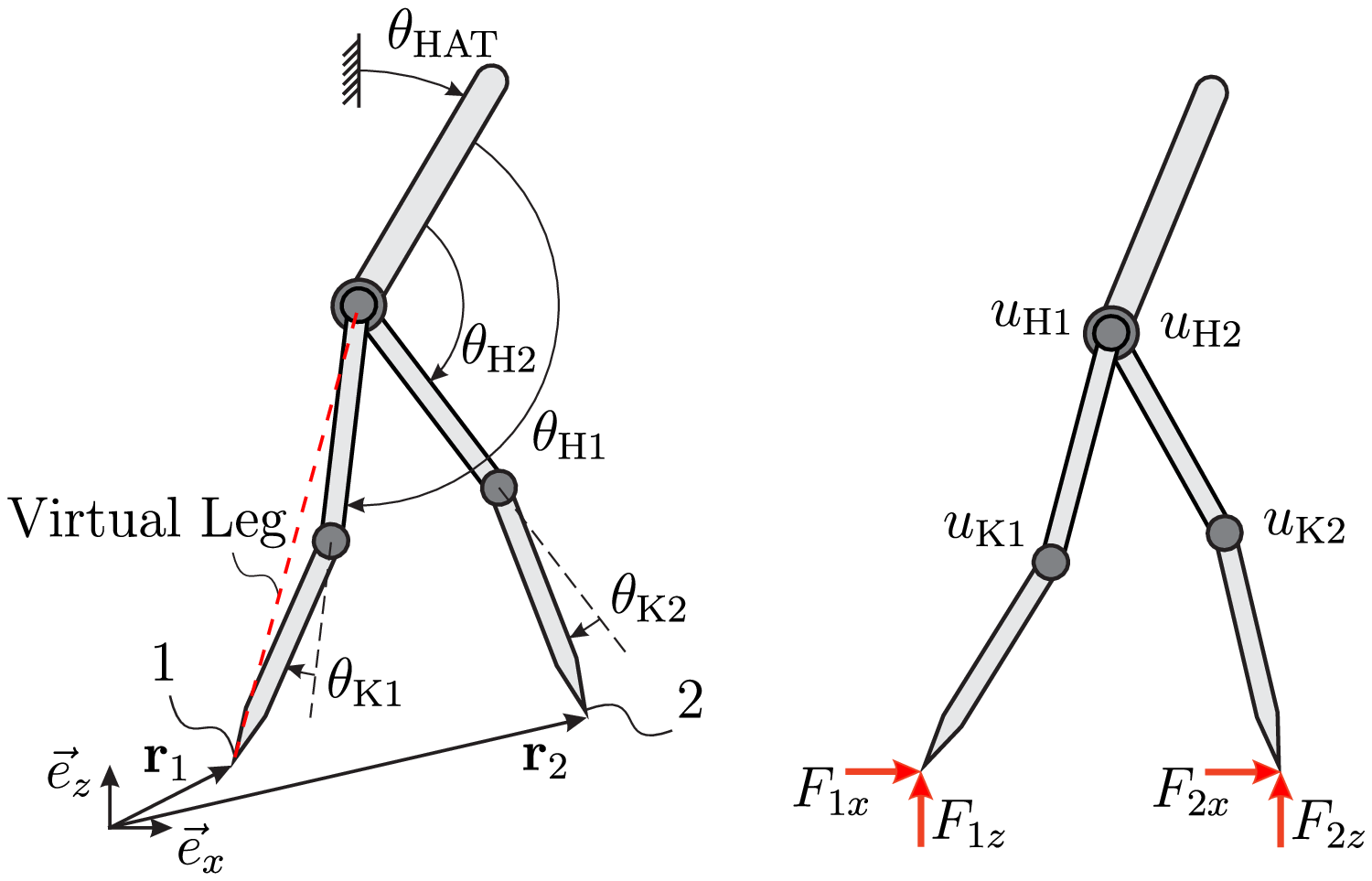}
	\caption{Left: free model with five segments and the coordinates. Right: four actuated joints and the reaction forces at both feet}
	\label{figRobotmodel}
	\end{figure}

	The robot model consists of five rigid body segments, which represent an upper body, two thighs and two shanks as illustrated in Figure \ref{figRobotmodel}. These segments are connected by four revolute joints in the hip and in the knees that are actuated by electric drive trains which provide driving torques for the motion. A planar model in the sagittal plane is assumed in accordance with studies on energy consumption in usual human walking, which state that the mechanical work in the frontal and transverse planes are of small magnitude \cite{EnergyPlaner1,EnergyPlaner2}. The lower end of each leg is modelled as a point foot. The non-slipping stance foot can therefore be regarded as an ideal revolute joint and does not transmit any torque to the ground.

	At a constant average walking speed---defined as the step length divided by the step duration---the gait is assumed to be a periodic sequence of alternating single support and double support phases. In both SSP and DSP, the system is characterized by continuous dynamics, while the phase transitions between them are described by discrete mappings: The SSP ends with an inelastic impact when the swing leg touches the ground (TD, touch down) and the DSP is terminated by the lift-off (LO) of the rear stance leg. This results in a hybrid dynamics model for the periodic walking gait. The degrees of freedom (DoF) of the system vary depending on the current walking phase: the total number of DoF of the robot model is 7, including 2 translational DoF for the absolute position and 5 DoF for the orientation of all segments, cf.\ Figure \ref{figRobotmodel} left. In the SSP, the model has 5 DoF, because the position of the stance leg foot is restricted by two constraints. In the DSP, there are 3 DoF, because both feet are restricted to contact the ground without slipping, meaning the distance between them remains constant. Accordingly, the reaction forces act on the single stance foot in the SSP and on both feet in the DSP. 

	The free model without ground contact is introduced in section \ref{labSubsectionFree}. Subsequently, the models for the SSP and DSP are derived in sections \ref{labSubsectionSSP} and \ref{labSubsectionDSP}, and for the phase transitions TD and LO in \ref{labSubsectionSSP2DSP} and \ref{labSubsectionDSP2SSP}. The continuous phases and the phase transitions are combined into a hybrid model for periodic walking gaits in section \ref{labSubsectionHybrid}. Since both legs of the robot are identical, periodic gaits are completely defined by one sequence of DSP$\rightarrow$LO$\rightarrow$SSP$\rightarrow$TD and a periodicity condition that maps the state after TD to the beginning of the DSP. This is achieved by swapping the legs, which is incorporated into the model for the phase transition TD in section \ref{labSubsectionSSP2DSP}.

\subsection{Free Model}\label{labSubsectionFree}
	As depicted in Figure \ref{figRobotmodel}, the generalized coordinates $\hat{\mathbf{q}}_\mathrm{f} = [\mathbf{r}_1^\top, \hat{\mathbf{q}}_\mathrm{s}^\top]^\top$ describe the system configuration in the $\{\vec{e}_x,\vec{e}_z\}$ plane, where $\mathbf{r}_1=[x_1,z_1]^\top$ is the position of foot 1 and $\hat{\mathbf{q}}_\mathrm{s} = [\theta_{\mathrm{HAT}}, \mathbf{q}_\mathrm{b}^\top]^\top$ contains the upper body orientation $\theta_{\mathrm{HAT}}$ and the four joint angles $\mathbf{q}_\mathrm{b} = [\theta_{\mathrm{H}1}, \theta_{\mathrm{H}2}, \theta_{\mathrm{K}1}, \theta_{\mathrm{K}2}]^\top$. With the actuator torques $\mathbf{u} = [u_{\mathrm{H}1}, u_{\mathrm{H}2}, u_{\mathrm{K}1}, u_{\mathrm{K}2}]^\top$, the position $\mathbf{r}_2=[x_2,z_2]^\top$ and the forces $\mathbf{F}_1=[F_{1x},F_{1z}]^\top$, $\mathbf{F}_2=[F_{2x},F_{2z}]^\top$ at both feet, the principle of virtual power yields the equations of motion
	\begin{equation}\label{eqODEfree}
		\underbrace{\begin{bmatrix}
			\hat{\mathbf{M}}_{\mathrm{f},00}\, \hat{\mathbf{M}}_{\mathrm{f},01}\\
			\hat{\mathbf{M}}_{\mathrm{f},10} \,\hat{\mathbf{M}}_{\mathrm{f},11}
		\end{bmatrix}}_{\hat{\mathbf{M}}_{\mathrm{f}}} \underbrace{\begin{bmatrix}
			\ddot{\mathbf{r}}_1\\ \ddot{\hat{\mathbf{q}}}_\mathrm{s}
		\end{bmatrix}}_{\ddot{\hat{\mathbf{q}}}_\mathrm{f}} + \underbrace{\begin{bmatrix}
			\hat{\mathbf{\Gamma}}_{\mathrm{f},0}\\ \hat{\mathbf{\Gamma}}_{\mathrm{f},1}
		\end{bmatrix}}_{\hat{\mathbf{\Gamma}}_{\mathrm{f}}} = 		\begin{bmatrix}
			\mathbf{0}_{2\times 4}\\ \hat{\mathbf{B}}_\mathrm{s}
		\end{bmatrix} \mathbf{u}
			+ \left(\frac{\partial \dot{\mathbf{r}}_1}{\partial \dot{\hat{\mathbf{q}}}_\mathrm{f}}\right)^{\!\!\!\top\!} \mathbf{F}_1
			+ \left(\frac{\partial \dot{\mathbf{r}}_2}{\partial \dot{\hat{\mathbf{q}}}_\mathrm{f}}\right)^{\!\!\!\top\!} \mathbf{F}_2
		\text{~,}
	\end{equation}
	where $\hat{\mathbf{M}}_{\mathrm{f}}$ is the mass matrix, the generalized forces $\hat{\mathbf{\Gamma}}_{\mathrm{f}}$ include gravity and Coriolis forces, $\mathbf{0}_{2\times 4}$ is a $2\times4$ matrix of zeros and $\hat{\mathbf{B}}_\mathrm{s}$ is defined in \eqref{eqODEssp}. 
	
\subsection{Single Support Phase Model}\label{labSubsectionSSP}
	In the SSP, the stance leg foot 1 remains in contact with the ground without slipping which is described by the two constraints $\dot{\mathbf{r}}_1=\mathbf{0}$. Since the swing leg foot 2 does not interact with the ground, $\mathbf{F}_2=\mathbf{0}$. The equations of motion for this 5 DoF system with coordinates $\hat{\mathbf{q}}_\mathrm{s}$ are
	\begin{equation}\label{eqODEssp}
		\underbrace{\begin{bmatrix}
			\hat{M}_{\mathrm{s},11} \,\hat{\mathbf{M}}_{\mathrm{s},12}\\
			\hat{\mathbf{M}}_{\mathrm{s},21} \,\hat{\mathbf{M}}_{\mathrm{s},22}
		\end{bmatrix}}_{\hat{\mathbf{M}}_{\mathrm{s}}} \underbrace{\begin{bmatrix}
			\ddot{\theta}_{\mathrm{HAT}}\\ \ddot{\mathbf{q}}_\mathrm{b}
		\end{bmatrix}}_{\ddot{\hat{\mathbf{q}}}_\mathrm{s}} + \underbrace{\begin{bmatrix}
			\Gamma_{\mathrm{s},1}\\ \hat{\mathbf{\Gamma}}_{\mathrm{s},2}
		\end{bmatrix}}_{\hat{\mathbf{\Gamma}}_{\mathrm{s}}}
= 		\underbrace{\begin{bmatrix}
			\mathbf{0}_{1\times 4}\\ \mathbf{I}_{4}
		\end{bmatrix}}_{\hat{\mathbf{B}}_\mathrm{s}} \mathbf{u}\text{~,}
	\end{equation}
	with mass matrix $\hat{\mathbf{M}}_{\mathrm{s}}$, generalized forces $\hat{\mathbf{\Gamma}}_{\mathrm{s}}$ and the $4\times4$ unit matrix $\mathbf{I}_{4}$. The interaction force $\mathbf{F}_1$ between the stance foot and the ground is derived from the first two rows\footnote{Notice that $\partial \dot{\mathbf{r}}_1/\partial \dot{\mathbf{r}}_1 = \mathbf{I}_{2}$ are the first two columns of the Jacobian $\partial \dot{\mathbf{r}}_1/\partial \dot{\hat{\mathbf{q}}}_\mathrm{f}$.} of \eqref{eqODEfree} as
	\begin{equation}\label{eqSspForce}
		\mathbf{F}_1 = \hat{\mathbf{M}}_{\mathrm{f},01} \ddot{\hat{\mathbf{q}}}_\mathrm{s} + \hat{\mathbf{\Gamma}}_{\mathrm{f},0} \text{~.}
	\end{equation}

\subsection{Double Support Phase Model}\label{labSubsectionDSP}
	In the DSP, both legs contact the ground without slipping which is expressed by the four constraints $\dot{\mathbf{r}}_1=\mathbf{0}$ and $\dot{\mathbf{r}}_2=\mathbf{0}$. This creates a closed kinematic chain with the relation $\mathbf{r}_1(\hat{\mathbf{q}}_\mathrm{d}) - \mathbf{r}_2(\hat{\mathbf{q}}_\mathrm{d}) = [\ell_{\mathrm{step}}, 0]^\top$ and the constrained system with 3 DoF is described by the independent coordinates $\hat{\mathbf{q}}_\mathrm{d} = [\theta_{\mathrm{HAT}},\mathbf{q}^\top_\mathrm{di}]^\top$ with $\mathbf{q}_\mathrm{di} = [\theta_{\mathrm{H}1},\theta_{\mathrm{K}1}]^\top$.\footnote{To make the implementation unique, we specify that foot 1 is in front of foot 2 in the DSP. In this manner, $\mathbf{q}_\mathrm{di}$ includes the joint angles in the hip $\theta_{\mathrm{H}1}$ and the knee $\theta_{\mathrm{K}1}$ of the front leg. The step length $\ell_{\mathrm{step}}$ follows from the configuration at the end of the SSP and is constant during the DSP.} Since there are four independent actuators, the DSP model is overactuated. To derive the equations of motion in the independent coordinates $\mathbf{q}_\mathrm{di}$, explicit expressions for the holonomic constraints $\mathbf{q}_\mathrm{dd} = [\theta_{\mathrm{H}2},\theta_{\mathrm{K}2}]^\top = \hat{\mathbf{\Omega}}(\hat{\mathbf{q}}_\mathrm{d})$ are derived in appendix \ref{labAppendixClosedLoop} to eliminate the dependent joint angles $\mathbf{q}_\mathrm{dd}$. The principle of virtual power then gives the equations of motion
	\begin{equation}\label{eqODEdsp}
		\underbrace{\begin{bmatrix}
			M_{\mathrm{d},11} \,\hat{\mathbf{M}}_{\mathrm{d},12}\\
			\hat{\mathbf{M}}_{\mathrm{d},21} \,\hat{\mathbf{M}}_{\mathrm{d},22}
		\end{bmatrix}}_{\hat{\mathbf{M}}_{\mathrm{d}}} \underbrace{\begin{bmatrix}
			\ddot{\theta}_{\mathrm{HAT}}\\ \ddot{\mathbf{q}}_\mathrm{di}
		\end{bmatrix}}_{\ddot{\hat{\mathbf{q}}}_\mathrm{d}} + \underbrace{\begin{bmatrix}
			\Gamma_{\mathrm{d},1}\\ \hat{\mathbf{\Gamma}}_{\mathrm{d},2}
		\end{bmatrix}}_{\hat{\mathbf{\Gamma}}_{\mathrm{d}}}
		= \underbrace{\begin{bmatrix}
			\mathbf{0}_{1\times 2}\\ \mathbf{I}_2
		\end{bmatrix}}_{\hat{\mathbf{B}}_\mathrm{di}} \underbrace{\begin{bmatrix}
			u_{\mathrm{H}1}\\ u_{\mathrm{K}1}
		\end{bmatrix}}_{\mathbf{u}_{\mathrm{i}}}
		+ {\underbrace{\left(\frac{\partial \hat{\mathbf{\Omega}}}{\partial \hat{\mathbf{q}}_\mathrm{d}}
			\right)^{\!\!\!\top}}_{\mathbf{J}_{\hat{\mathbf{\Omega}}}^\top}} \underbrace{\begin{bmatrix}
			u_{\mathrm{H}2}\\ u_{\mathrm{K}2}
		\end{bmatrix}}_{\mathbf{u}_{\mathrm{d}}}\text{.}
	\end{equation}
	As a result of overactuation, $\theta_{\mathrm{HAT}}$, which is not actuated in the SSP (cf.\ \eqref{eqODEssp}), becomes controllable by the inputs $\mathbf{u}_{\mathrm{d}}$. In order to calculate the reaction forces $\mathbf{F}_1$ and $\mathbf{F}_2$, equation \eqref{eqODEfree} is rearranged into
	\begin{equation}\label{eqODEfreeSort}
		\begin{bmatrix}
			\tilde{\mathbf{M}}_{\mathrm{f},00} \, \tilde{\mathbf{M}}_{\mathrm{f},01} \, \tilde{\mathbf{M}}_{\mathrm{f},02}\\
			\tilde{\mathbf{M}}_{\mathrm{f},10} \, \tilde{\mathbf{M}}_{\mathrm{f},11} \, \tilde{\mathbf{M}}_{\mathrm{f},12} \\
			\tilde{\mathbf{M}}_{\mathrm{f},20} \, \tilde{\mathbf{M}}_{\mathrm{f},21} \, \tilde{\mathbf{M}}_{\mathrm{f},22} 
		\end{bmatrix} \begin{bmatrix}
			\ddot{\mathbf{r}}_1 \\ \ddot{\mathbf{q}}_\mathrm{dd} \\ \ddot{\hat{\mathbf{q}}}_\mathrm{d}
		\end{bmatrix} + \begin{bmatrix}
			\tilde{\mathbf{\Gamma}}_{\mathrm{f},0} \\ \tilde{\mathbf{\Gamma}}_{\mathrm{f},1} \\ \tilde{\mathbf{\Gamma}}_{\mathrm{f},2}
		\end{bmatrix} =  \begin{bmatrix}
			\mathbf{0}_{2\times 4} \\ \tilde{\mathbf{B}}_{\mathrm{f},1} \\ \tilde{\mathbf{B}}_{\mathrm{f},2}
		\end{bmatrix} \mathbf{u}
			+ \begin{bmatrix} \mathbf{I}_{2} \\ \mathbf{0}_{2\times 2} \\ \left(\frac{\partial \dot{\mathbf{r}}_1}{\partial \dot{\hat{\mathbf{q}}}_\mathrm{d}}\right)^{\!\!\top}
		\end{bmatrix} \mathbf{F}_1
			+ \begin{bmatrix} \mathbf{I}_{2} \\ \left(\frac{\partial \dot{\mathbf{r}}_2}{\partial \dot{\mathbf{q}}_\mathrm{dd}}\right)^{\!\!\top}\\ \left(\frac{\partial \dot{\mathbf{r}}_2}{\partial \dot{\hat{\mathbf{q}}}_\mathrm{d}}\right)^{\!\!\top}
		\end{bmatrix} \mathbf{F}_2
	\end{equation}
	with $\tilde{\mathbf{B}}_{\mathrm{f},1} = \partial \mathbf{q}_{\mathrm{dd}}/\partial \mathbf{q}_{\mathrm{b}}$ and $\tilde{\mathbf{B}}_{\mathrm{f},2} = \partial \hat{\mathbf{q}}_{\mathrm{d}}/\partial \mathbf{q}_{\mathrm{b}}$. The first two rows yields the sum of the reaction forces which is independent of any actuation $\mathbf{u}$:
	\begin{equation}\label{eqForceSum}
		\mathbf{F}_1 + \mathbf{F}_2 = \tilde{\mathbf{M}}_{\mathrm{f},00} \ddot{\mathbf{r}}_1 + \tilde{\mathbf{M}}_{\mathrm{f},01} \ddot{\mathbf{q}}_\mathrm{dd} + \tilde{\mathbf{M}}_{\mathrm{f},02} \ddot{\hat{\mathbf{q}}}_\mathrm{d} + \tilde{\mathbf{\Gamma}}_{\mathrm{f},0} \text{~.}
	\end{equation}
    Subsequently, $\mathbf{F}_2 := \mathbf{f}_{\mathbf{F}}(\mathbf{u}_{\mathrm{d}})$ is derived from the third and fourth row as
    \begin{equation}\label{eqForce2}
		 \mathbf{F}_2 = \left(\frac{\partial \dot{\mathbf{r}}_2}{\partial \dot{\mathbf{q}}_\mathrm{dd}}\right)^{-\top} \left(\tilde{\mathbf{M}}_{\mathrm{f},10} \ddot{\mathbf{r}}_1 + \tilde{\mathbf{M}}_{\mathrm{f},11} \ddot{\mathbf{q}}_\mathrm{dd} + \tilde{\mathbf{M}}_{\mathrm{f},12} \ddot{\hat{\mathbf{q}}}_\mathrm{d} + \tilde{\mathbf{\Gamma}}_{\mathrm{f},1} - \mathbf{u}_{\mathrm{d}}\right) \text{~,}
	\end{equation}
    which is a function of the actuation $\mathbf{u}_{\mathrm{d}}=(\partial \mathbf{q}_{\mathrm{dd}}/\partial \mathbf{q}_{\mathrm{b}}) \mathbf{u}$.
	
\subsection{Touch Down of Swing Leg}\label{labSubsectionSSP2DSP}
	The transition from SSP to DSP is modeled as an inelastic impact where the swing leg foot touches the ground. Assuming an impact event of infinitesimal duration, the configurations before and after the impact are equal $\hat{\mathbf{q}}_\mathrm{f}^- = \hat{\mathbf{q}}_\mathrm{f}^+$. Time integration of equation \eqref{eqODEfree} yields the change of momentum
	\begin{equation}\label{eqImpactMomentum}
		\hat{\mathbf{M}}_{\mathrm{f}}(\hat{\mathbf{q}}_\mathrm{f}^+) \dot{\hat{\mathbf{q}}}_\mathrm{f}^+ - \hat{\mathbf{M}}_{\mathrm{f}}(\hat{\mathbf{q}}_\mathrm{f}^-)\dot{\hat{\mathbf{q}}}_\mathrm{f}^-
		= \left(\frac{\partial \dot{\mathbf{r}}_1}{\partial \dot{\hat{\mathbf{q}}}_\mathrm{f}}\right)^{\!\!\top\!}\hat{\mathbf{F}}_1
		+ \left(\frac{\partial \dot{\mathbf{r}}_2}{\partial \dot{\hat{\mathbf{q}}}_\mathrm{f}}\right)^{\!\!\top\!}\hat{\mathbf{F}}_2\text{~,}
	\end{equation}
	where $\hat{\mathbf{F}}_1$ and $\hat{\mathbf{F}}_2$ are impulsive reaction forces at the feet to impose the constraints that the velocities of both feet after the inelastic impact have to be zero
	\begin{equation}\label{eqImpactVelocity}
		\dot{\mathbf{r}}_1^+ = \left(\frac{\partial \dot{\mathbf{r}}_1}{\partial \dot{\hat{\mathbf{q}}}_\mathrm{f}}\right) \dot{\hat{\mathbf{q}}}_\mathrm{f}^+ =  \mathbf{0}\text{~,}\quad
		\dot{\mathbf{r}}_2^+ = \left(\frac{\partial \dot{\mathbf{r}}_2}{\partial \dot{\hat{\mathbf{q}}}_\mathrm{f}}\right) \dot{\hat{\mathbf{q}}}_\mathrm{f}^+ =  \mathbf{0}\text{~.}
	\end{equation}
	The impulsive reaction forces and the generalized velocities after the impact as functions of the state at the end of the SSP follow from the linear system of equations
	\begin{equation}\label{eqImpact}
		\underbrace{\begin{bmatrix}
			\hat{\mathbf{M}}_{\mathrm{f}}(\hat{\mathbf{q}}_\mathrm{f}^-)\,& -\left(\frac{\partial \dot{\mathbf{r}}_1}{\partial \dot{\hat{\mathbf{q}}}_\mathrm{f}}\right)^{\!\!\top}\,& -\left(\frac{\partial \dot{\mathbf{r}}_2}{\partial \dot{\hat{\mathbf{q}}}_\mathrm{f}}\right)^{\!\!\top}\\
			\left(\frac{\partial \dot{\mathbf{r}}_1}{\partial \dot{\hat{\mathbf{q}}}_\mathrm{f}}\right)\,&\mathbf{0}_{2\times 2}\,&\mathbf{0}_{2\times 2} \\
			\left(\frac{\partial \dot{\mathbf{r}}_2}{\partial \dot{\hat{\mathbf{q}}}_\mathrm{f}}\right)\,&\mathbf{0}_{2\times 2}\,&\mathbf{0}_{2\times 2}
		\end{bmatrix}}_{\mathbf{\Pi}_{\mathrm{lhs}}} \begin{bmatrix}
			\dot{\hat{\mathbf{q}}}_\mathrm{f}^+\\ \hat{\mathbf{F}}_{1}\\ \hat{\mathbf{F}}_{2}
		\end{bmatrix} \\ = \underbrace{\begin{bmatrix}
			\hat{\mathbf{M}}_{\mathrm{f}}(\hat{\mathbf{q}}_\mathrm{f}^-)\\\mathbf{0}_{2\times 7}\\\mathbf{0}_{2\times 7}
		\end{bmatrix}\left(\frac{\partial \dot{\hat{\mathbf{q}}}_\mathrm{f}}{\partial \dot{\hat{\mathbf{q}}}_\mathrm{s}}
			\right)}_{\mathbf{\Pi}_{\mathrm{rhs}}}\dot{\hat{\mathbf{q}}}_\mathrm{s}^-\text{.}
	\end{equation}
	In order to describe each walking gait by one periodic sequence \ldots$\rightarrow$DSP$\rightarrow$LO$\rightarrow$SSP$\rightarrow$TD$\rightarrow$\ldots, the legs are swapped after the touch down of the former swing leg which is achieved by reordering the joint angles and velocities via
	\begin{align*}
	    [\theta_{\mathrm{H}1}, \theta_{\mathrm{H}2}, \theta_{\mathrm{K}1}, \theta_{\mathrm{K}2}]^\top &\rightarrow [\theta_{\mathrm{H}2}, \theta_{\mathrm{H}1}, \theta_{\mathrm{K}2}, \theta_{\mathrm{K}1}]^\top\text{,}\\
	    [\dot{\theta}_{\mathrm{H}1}, \dot{\theta}_{\mathrm{H}2}, \dot{\theta}_{\mathrm{K}1}, \dot{\theta}_{\mathrm{K}2}]^\top &\rightarrow [\dot{\theta}_{\mathrm{H}2}, \dot{\theta}_{\mathrm{H}1}, \dot{\theta}_{\mathrm{K}2}, \dot{\theta}_{\mathrm{K}1}]^\top\text{.}
	\end{align*}
	This is included in the mappings of the independent coordinates and velocities from the end of the SSP to the beginning of the DSP
	\begin{equation}\label{eqSwitchLegsCoord}
		\begin{aligned}
		\hat{\mathbf{q}}_\mathrm{d}^+ &= \mathbf{R}_\mathrm{s}^\mathrm{d}\hat{\mathbf{q}}_\mathrm{s}^- \text{,}\\
		\dot{\hat{\mathbf{q}}}_\mathrm{d}^+ &= \underbrace{\begin{bmatrix} \mathbf{0}_{3\times 2} & \mathbf{R}_\mathrm{s}^\mathrm{d} & \mathbf{0}_{3\times 4} \end{bmatrix} \mathbf{\Pi}_{\mathrm{lhs}}^{-1}\mathbf{\Pi}_{\mathrm{rhs}}}_{\mathbf{\Delta}_{\mathrm{s}}}\dot{\hat{\mathbf{q}}}_\mathrm{s}^- \text{,}
		\end{aligned}
	\end{equation}
	with
	\begin{equation}\label{eqSwitchMatrix}
		\mathbf{R}_\mathrm{s}^\mathrm{d} = \begin{bmatrix} 1 &0 &0 &0 &0\\ 0 &0 &1 &0 &0\\ 0 &0 &0 &0 &1
		\end{bmatrix}\text{.}
	\end{equation}
	
\subsection{Lift-off of Swing Leg}\label{labSubsectionDSP2SSP}
	According to the experimental validation in \cite{HZDwalker5}, electric motors have much faster dynamics than the attached mechanical system, which allows for abrupt torque changes that can be used to terminate the DSP and initiate the lift-off of the swing leg at the beginning of the SSP at any desired time. This is modeled as an instantaneous lift-off event with continuous configuration and velocities, while discontinuities occur at the acceleration level. The independent coordinates and velocities are mapped from the end of the DSP to the beginning of the SSP via
	\begin{equation}\label{eqLiftoff}
		\begin{aligned}
		\hat{\mathbf{q}}_\mathrm{s}^+ &= \mathbf{R}_\mathrm{d}^\mathrm{s}\begin{bmatrix}
			\hat{\mathbf{q}}_\mathrm{d}^-\\ \hat{\mathbf{\Omega}}(\hat{\mathbf{q}}_\mathrm{d}^-)
		\end{bmatrix} \text{,} \\
		\dot{\hat{\mathbf{q}}}_\mathrm{s}^+ &= \mathbf{R}_\mathrm{d}^\mathrm{s}\begin{bmatrix}
			\mathbf{I}_3\\ \mathbf{J}_{\hat{\mathbf{\Omega}}}(\hat{\mathbf{q}}_\mathrm{d}^-)
		\end{bmatrix}\dot{\hat{\mathbf{q}}}_\mathrm{d}^- \text{~,}
		\end{aligned}
	\end{equation}
	with the Jacobian $\mathbf{J}_{\hat{\mathbf{\Omega}}}(\hat{\mathbf{q}}_\mathrm{d}^-) = \partial \hat{\mathbf{\Omega}}(\hat{\mathbf{q}}_\mathrm{d}^-)/\partial \hat{\mathbf{q}}_\mathrm{d}$ and
	\begin{equation}\label{eqLiftoffMatrix}
		\mathbf{R}_\mathrm{d}^\mathrm{s} = \begin{bmatrix} 1 &0 &0 &0 &0\\ 0 &1 &0 &0 &0
			\\ 0 &0 &0 &1 &0\\ 0 &0 &1 &0 &0\\ 0 &0 &0 &0 &1
		\end{bmatrix}\text{.}
	\end{equation}

\subsection{Hybrid Model for Periodic Walking}\label{labSubsectionHybrid}
    The continuous SSP and DSP and the discrete phase transitions LO and TD are combined into a hybrid model for periodic walking gaits of the bipedal robot system. First, additional coordinate transformations are performed to define the phase transitions and as preparation for the control design in section \ref{labSectionControl}. The hybrid model is then defined in \eqref{eqHybridSystem}.
    
	For the purpose of control design, the independent coordinate $\theta = \theta_{\mathrm{HAT}} + \theta_{\mathrm{H}1} + \frac{1}{2}\theta_{\mathrm{K}1}$ \cite[section~\mbox{VII-B}]{HZDwalker1} is introduced to describe the absolute orientation of the robot in both the SSP and the DSP. If the shank and thigh lengths are equal, $\theta$~is the angle of the virtual leg (the straight line that connects the hip and the stance foot) as depicted in Figure \ref{figRobotmodel}.\footnote{The transformation is also used when the lengths of shanks and thighs are different. In this general case, however, the simple interpretation of $\theta$ as the angle of the virtual leg is no longer valid.} By requiring $\theta$ to increase monotonically during each step, state-dependent rather than time-dependent reference trajectories for the controller can be defined in section \ref{labSectionControl}. This transformation introduces new sets of generalized coordinates for the SSP and DSP, and the phase transitions are modified accordingly.
	
	The transformation defines new SSP coordinates
	\begin{equation}\label{eqCanoSsp}
		\mathbf{q}_\mathrm{s} = \begin{bmatrix} \theta \\ \mathbf{q}_\mathrm{b} \end{bmatrix} = \mathbf{H}_\mathrm{s}\hat{\mathbf{q}}_\mathrm{s} \text{~,}
	\end{equation}
	with
	\begin{equation}\label{eqCanoSspMat}
		\mathbf{H}_\mathrm{s}=\begin{bmatrix}
		    1 & 1 & 0 & \frac{1}{2} & 0\\
		    0 & 1 & 0 & 0 & 0 \\
		    0 & 0 & 1 & 0 & 0 \\
		    0 & 0 & 0 & 1 & 0 \\
		    0 & 0 & 0 & 0 & 1
		\end{bmatrix} \text{,~}
		\mathbf{H}_\mathrm{s}^{-1}=\begin{bmatrix}
		    1 & -1 & 0 & -\frac{1}{2} & 0\\
		    0 & 1 & 0 & 0 & 0 \\
		    0 & 0 & 1 & 0 & 0 \\
		    0 & 0 & 0 & 1 & 0 \\
		    0 & 0 & 0 & 0 & 1
		\end{bmatrix} \text{.}
	\end{equation}
 
	Substituting \eqref{eqCanoSsp} into \eqref{eqODEssp} gives the equations of motion
	\begin{equation}\label{eqODEsspTheta}
		\mathbf{M}_{\mathrm{s}} \ddot{\mathbf{q}}_\mathrm{s} + \mathbf{\Gamma}_\mathrm{s} = \mathbf{B}_\mathrm{s}\mathbf{u}\text{,}
	\end{equation}
	with $\mathbf{M}_{\mathrm{s}} = \mathbf{H}_\mathrm{s}^{-\top}\hat{\mathbf{M}}_{\mathrm{s}}\mathbf{H}_\mathrm{s}^{-1}$, $\mathbf{\Gamma}_{\mathrm{s}} = \mathbf{H}_\mathrm{s}^{-\top}\hat{\mathbf{\Gamma}}_{\mathrm{s}}$ and $\mathbf{B}_\mathrm{s} = \hat{\mathbf{B}}_\mathrm{s}$.

	The equation of motion \eqref{eqODEsspTheta} is then expressed in state space 
	\begin{equation}\label{eqStateSpacessp}
		\begin{aligned}
		\dot{\mathbf{x}}_\mathrm{s} &= \begin{bmatrix}
			\dot{\mathbf{q}}_\mathrm{s}\\
			\mathbf{M}_{\mathrm{s}}^{-1}\left(-\mathbf{\Gamma}_\mathrm{s}+\mathbf{B}_\mathrm{s}\mathbf{u}\right)
		\end{bmatrix} \\
		&= \underbrace{\begin{bmatrix}
			\dot{\mathbf{q}}_\mathrm{s}\\
			- \mathbf{M}_{\mathrm{s}}^{-1} \mathbf{\Gamma}_\mathrm{s}
			\end{bmatrix}}_{\mathbf{f}_{\mathrm{s}}(\mathbf{x}_\mathrm{s})} + \underbrace{\begin{bmatrix} \mathbf{0}_{5\times 4} \\ \mathbf{M}_{\mathrm{s}}^{-1}\mathbf{B}_\mathrm{s} 
			\end{bmatrix}}_{\mathbf{g}_{\mathrm{s}}(\mathbf{x}_\mathrm{s})} \mathbf{u}\text{~,}
		\end{aligned}
	\end{equation}
	on the manifold $\mathcal{X}_{\mathrm{s}} = T\mathcal{Q}_{\mathrm{s}} = \{\mathbf{x}_\mathrm{s} = [\mathbf{q}_\mathrm{s}^\top,\dot{\mathbf{q}}_\mathrm{s}^\top]^\top | \,\mathbf{q}_\mathrm{s} \in \mathcal{Q}_{\mathrm{s}}, \dot{\mathbf{q}}_\mathrm{s} \in \mathbb{R}^5 \}$, where $\mathcal{Q}_{\mathrm{s}}=\mathbb{T}^5 $ is the 5-torus.
	
	Analogously, the transformation defines new DSP coordinates\footnote{Notice that the simple expression for this transformation results from the choice of $\mathbf{q}_\mathrm{di} = [\theta_{\mathrm{H}1},\theta_{\mathrm{K}1}]^\top$ in section \ref{labSubsectionDSP}.}
	\begin{equation}\label{eqCanoDsp}
		\mathbf{q}_\mathrm{d} = \begin{bmatrix} \theta \\ \mathbf{q}_\mathrm{di} \end{bmatrix} = \mathbf{H}_\mathrm{d}\hat{\mathbf{q}}_\mathrm{d} \text{~,}
	\end{equation}
	with
	\begin{equation}\label{eqCanoDspMat}
		\mathbf{H}_\mathrm{d}=\begin{bmatrix}
		    1 & 1 & \frac{1}{2}\\
		    0 & 1 & 0 \\
		    0 & 0 & 1
		\end{bmatrix} \text{,~}
		\mathbf{H}_\mathrm{d}^{-1}=\begin{bmatrix}
		    1 & -1 & -\frac{1}{2}\\
		    0 & 1 & 0 \\
		    0 & 0 & 1
		\end{bmatrix} \text{.}
	\end{equation}
	Thus, equation \eqref{eqODEdsp} is transformed into 
	\begin{equation}\label{eqODEdspTheta}
		\mathbf{M}_{\mathrm{d}} \ddot{\mathbf{q}}_\mathrm{d} + \mathbf{\Gamma}_\mathrm{d} = \mathbf{B}_\mathrm{di}\mathbf{u}_{\mathrm{i}} + \mathbf{J}_{\mathbf{\Omega}}^\top\mathbf{u}_{\mathrm{d}} = \mathbf{B}_\mathrm{d}\mathbf{u} \text{~,}
	\end{equation}
	with $\mathbf{M}_{\mathrm{d}} = \mathbf{H}_\mathrm{d}^{-\top}\hat{\mathbf{M}}_{\mathrm{d}}\mathbf{H}_\mathrm{d}^{-1}$, $\mathbf{\Gamma}_{\mathrm{d}} = \mathbf{H}_\mathrm{d}^{-\top}\hat{\mathbf{\Gamma}}_{\mathrm{d}}$, $\mathbf{B}_\mathrm{di} = \hat{\mathbf{B}}_\mathrm{di}$, $\mathbf{q}_\mathrm{dd} = \mathbf{\Omega}(\mathbf{q}_\mathrm{d}) = \hat{\mathbf{\Omega}}\left(\mathbf{H}_\mathrm{d}^{-1}\mathbf{q}_\mathrm{d}\right)$ and Jacobian $\mathbf{J}_{\mathbf{\Omega}} = \mathbf{J}_{\hat{\mathbf{\Omega}}}\mathbf{H}_\mathrm{d}^{-1}$. The input matrix $\mathbf{B}_\mathrm{d}$ is obtained by reordering and combining $\mathbf{B}_\mathrm{di}$ and $\mathbf{J}_{\mathbf{\Omega}}^\top$.
	
	The corresponding state space expression for the equations of motion in the DSP is
	\begin{equation}\label{eqStateSpacedsp}
		\begin{aligned}
		\dot{\mathbf{x}}_\mathrm{d} &= \begin{bmatrix}
			\dot{\mathbf{q}}_\mathrm{d}\\
			\mathbf{M}_{\mathrm{d}}^{-1}\left(-\mathbf{\Gamma}_\mathrm{d}+
			\mathbf{B}_\mathrm{di}\mathbf{u}_{\mathrm{i}}+\mathbf{J}_{\mathbf{\Omega}}^\top\mathbf{u}_{\mathrm{d}}\right)
		\end{bmatrix} \\
		&= \underbrace{\begin{bmatrix}
			\dot{\mathbf{q}}_\mathrm{d}\\
			- \mathbf{M}_{\mathrm{d}}^{-1} \mathbf{\Gamma}_\mathrm{d}
			\end{bmatrix}}_{\mathbf{f}_{\mathrm{d}}(\mathbf{x}_\mathrm{d})} + \underbrace{\begin{bmatrix} \mathbf{0}_{3\times 2} \\ \mathbf{M}_{\mathrm{d}}^{-1}\mathbf{B}_\mathrm{di} 
			\end{bmatrix}}_{\mathbf{g}_{\mathrm{i}}(\mathbf{x}_\mathrm{d})} \mathbf{u}_{\mathrm{i}} + \underbrace{\begin{bmatrix} \mathbf{0}_{3\times 2} \\ 
			\mathbf{M}_{\mathrm{d}}^{-1}\mathbf{J}_{\mathbf{\Omega}}^\top \end{bmatrix}}_{\mathbf{g}_{\mathrm{d}}(\mathbf{x}_\mathrm{d})} \mathbf{u}_{\mathrm{d}}\text{~,}
		\end{aligned}
	\end{equation}
	where the state space is $\mathcal{X}_{\mathrm{d}} = T\mathcal{Q}_{\mathrm{d}} = \{\mathbf{x}_\mathrm{d} = [\mathbf{q}_\mathrm{d}^\top,\dot{\mathbf{q}}_\mathrm{d}^\top]^\top | \,\mathbf{q}_\mathrm{d} \in \mathcal{Q}_{\mathrm{d}}, \dot{\mathbf{q}}_\mathrm{d} \in \mathbb{R}^3 \}$ with $\mathcal{Q}_{\mathrm{d}}=\mathbb{T}^3$. 
	
	The discontinuous phase transitions are expressed in terms of the states $\mathbf{x}_\mathrm{s}=[\mathbf{q}_\mathrm{s}^\top,\dot{\mathbf{q}}_\mathrm{s}^\top]^\top$ and $\mathbf{x}_\mathrm{d}=[\mathbf{q}_\mathrm{d}^\top,\dot{\mathbf{q}}_\mathrm{d}^\top]^\top$, where the superscript $-$ ($+$) is used to label states before (after) the respective transition. The inelastic impact of the swing leg with the ground at the end of the SSP gives the relation $\mathbf{x}_\mathrm{d}^{+}=\mathbf{\Delta}_\mathrm{s}^{\mathrm{d}}(\mathbf{x}_\mathrm{s}^{-})$, which follows from the mappings \eqref{eqSwitchLegsCoord} as
	\begin{equation}\label{eqSwitchLegsCoordState}
		\begin{aligned}
		\mathbf{q}_\mathrm{d}^+ &= \mathbf{H}_\mathrm{d}\mathbf{R}_\mathrm{s}^\mathrm{d}\mathbf{H}_\mathrm{s}^{-1}\mathbf{q}_\mathrm{s}^- \text{~,} \\
		\dot{\mathbf{q}}_\mathrm{d}^+ &= \mathbf{H}_\mathrm{d}\mathbf{\Delta}_{\mathrm{s}}\mathbf{H}_\mathrm{s}^{-1}\dot{\mathbf{q}}_\mathrm{s}^- \text{~.}
		\end{aligned}
	\end{equation}
	
	The swing leg lift-off at the end of the DSP is described by $\mathbf{x}_\mathrm{s}^{+}=\mathbf{\Delta}_\mathrm{d}^{\mathrm{s}}\left(\mathbf{x}_\mathrm{d}^{-}\right)$, where \eqref{eqLiftoff} gives
	\begin{equation}\label{eqLiftoffState}
		\begin{aligned}
		\mathbf{q}_\mathrm{s}^+ &= \mathbf{H}_\mathrm{s}\mathbf{R}_\mathrm{d}^\mathrm{s}\begin{bmatrix}
			\mathbf{H}_\mathrm{d}^{-1}\mathbf{q}_\mathrm{d}^-\\ \mathbf{\Omega}(\mathbf{q}_\mathrm{d}^-)
		\end{bmatrix} \text{~,} \\
		\dot{\mathbf{q}}_\mathrm{s}^+ &= \mathbf{H}_\mathrm{s}\mathbf{R}_\mathrm{d}^\mathrm{s}\begin{bmatrix}
			\mathbf{H}_\mathrm{d}^{-1}\\ \mathbf{J}_{\mathbf{\Omega}}(\mathbf{q}_\mathrm{d}^-)
		\end{bmatrix}\dot{\mathbf{q}}_\mathrm{d}^- \text{~.}
		\end{aligned}
	\end{equation}
	
	Periodic walking gaits that consist of an infinite sequence \ldots$\rightarrow$DSP$\rightarrow$LO$\rightarrow$SSP$\rightarrow$TD$\rightarrow$\ldots are described by the hybrid dynamical system
	\begin{equation}\label{eqHybridSystem}
		\Sigma :
		\begin{cases}
			\dot{\mathbf{x}}_\mathrm{d}^{\hphantom{+}} = \mathbf{f}_{\mathrm{d}}(\mathbf{x}_\mathrm{d})+\mathbf{g}_{\mathrm{i}}(\mathbf{x}_\mathrm{d})
				\mathbf{u}_{\mathrm{i}}+\mathbf{g}_{\mathrm{d}}(\mathbf{x}_\mathrm{d})\mathbf{u}_{\mathrm{d}} \text{~,} &
				\mathbf{x}_\mathrm{d}^{\hphantom{-}} \notin \mathcal{S}_\mathrm{d}^{\mathrm{s}} \text{,} \\
			\mathbf{x}_\mathrm{s}^{+} = \mathbf{\Delta}_\mathrm{d}^{\mathrm{s}}(\mathbf{x}_\mathrm{d}^{-}) \text{~,} &
				\mathbf{x}_\mathrm{d}^{-} \in \mathcal{S}_\mathrm{d}^{\mathrm{s}} \text{,}\\
			\dot{\mathbf{x}}_\mathrm{s}^{\hphantom{+}} = \mathbf{f}_{\mathrm{s}}(\mathbf{x}_\mathrm{s})+
				\mathbf{g}_{\mathrm{s}}(\mathbf{x}_\mathrm{s})\mathbf{u} \text{~,} &
				\mathbf{x}_\mathrm{s}^{\hphantom{-}} \notin \mathcal{S}_\mathrm{s}^{\mathrm{d}} \text{,} \\
			\mathbf{x}_\mathrm{d}^{+} = \mathbf{\Delta}_\mathrm{s}^{\mathrm{d}}(\mathbf{x}_\mathrm{s}^{-}) \text{~,} &
				\mathbf{x}_\mathrm{s}^{-} \in \mathcal{S}_\mathrm{s}^{\mathrm{d}} \text{,}\\
			\mathcal{S}_\mathrm{d}^{\mathrm{s}} = \{\mathbf{x}_\mathrm{d} \in \mathcal{X}_{\mathrm{d}} |\, 
				H_\mathrm{d}^{\mathrm{s}}(\mathbf{x}_\mathrm{d})=0\}\text{~,}\\
			\mathcal{S}_\mathrm{s}^{\mathrm{d}} = \{\mathbf{x}_\mathrm{s} \in \mathcal{X}_{\mathrm{s}} |\, 
				H_\mathrm{s}^{\mathrm{d}}(\mathbf{x}_\mathrm{s})=0\} \text{~.}
		\end{cases}
	\end{equation}
	The SSP ends when the swing leg touches the ground, which is detected by the distance of the swing leg foot to ground $H_\mathrm{s}^{\mathrm{d}}(\mathbf{x}_\mathrm{s})=z_2(\mathbf{x}_\mathrm{s})$, thus the TD event is triggered when $\mathbf{x}_\mathrm{s} \in \mathcal{S}_\mathrm{s}^{\mathrm{d}}: H_\mathrm{s}^{\mathrm{d}}(\mathbf{x}_\mathrm{s})=0$. The transition from DSP to SSP is initiated by the controller if $\theta=\theta_{\mathrm{DSP}}$, thus $H_\mathrm{d}^{\mathrm{s}}(\mathbf{x}_\mathrm{d})=\theta-\theta_{\mathrm{DSP}}$\footnote{In contrast to the more intuitive condition of a vanishing reaction force $\mathbf{F}_2=\mathbf{0}$, the chosen definition depends only on the configuration, but not on the velocities and not on the accelerations or inputs. This is advantageous below when gaits are generated via optimization.}, and the LO event is triggered when $\mathbf{x}_\mathrm{d} \in \mathcal{S}_\mathrm{d}^{\mathrm{s}}: H_\mathrm{d}^{\mathrm{s}}(\mathbf{x}_\mathrm{d})=0$. $\theta_{\mathrm{DSP}}$ is a parameter of the controller. In the following section (\ref{labSectionControl}), controllers for stable periodic solutions of the hybrid dynamical system are designed. 
	
\section{Hybrid Zero Dynamics Controller}\label{labSectionControl}
	Studies in \cite{HZDwalker1,HZDwalker4} suggest using a hybrid zero dynamics (HZD) controller\footnote{Hybrid zero dynamics refers to the zero dynamics of the hybrid system.} for stabilizing periodic walking gaits of a similar model with continuous SSP and an instantaneous DSP. Below, this control design is extended and applied to the hybrid system (\ref{labSubsectionHybrid}) in order to generate and stabilize periodic gaits with some constant average walking speed. The controller for the SSP is adopted form \cite{HZDwalker1,HZDwalker4} with marginal modifications as summarized in section \ref{labSubsectionControlSSP} for the sake of completeness. Subsequently, different controllers for the non-instantaneous DSP are proposed: In section \ref{labSubsectionDSPunderact}, a controller with under-actuated DSP is designed by means of a virtual input $\tilde{\mathbf{u}}$. In section \ref{labSubsectionDSPfullact}, a controller that utilizes the full actuation in the DSP to increase the system's stability is proposed. Finally, in section \ref{labSubsectionDSPoveract}, the overactuated DSP is stabilized by the control design that includes the contact force as an additional control objective.
	
\subsection{SSP Controller}\label{labSubsectionControlSSP}
	The controller for the continuous SSP is based on parametric reference trajectories $\mathbf{q}_\mathrm{r,s}(\theta,\boldsymbol{\alpha}_\mathrm{s})$---functions of the state $\theta$ and constant parameters $\boldsymbol{\alpha}_\mathrm{s}$---for the four joint angles $\mathbf{q}_\mathrm{b}$.\footnote{Any parametric function may be used to define the reference trajectories. We use Bézier polynomials in our implementation thus $\boldsymbol{\alpha}_\mathrm{s}$ are the coefficients of the polynomials.} The objective of the controller is to zero the output
	\begin{equation}\label{eqFeedbackssp}
		\mathbf{y}_{\mathrm{s}} = \mathbf{h}_\mathrm{s}(\mathbf{x}_\mathrm{s}) = \mathbf{q}_\mathrm{b}- \mathbf{q}_\mathrm{r,s}(\theta,\boldsymbol{\alpha}_\mathrm{s})\text{~,}
	\end{equation}
	i.\,e.\ the deviation of the joint angles form their reference trajectories. This is achieved by feedback linearization for this multiple input multiple output (MIMO) control task. The output $\mathbf{y}_{\mathrm{s}}$ is differentiated twice (the system has the vector relative degree 2) using the Lie derivative (denoted by $\mathcal{L}$) until the control input $\mathbf{u}$ appears for the first time:
	\begin{subequations}\label{eqFeedbackSspDiff}
		\begin{align}
		\dot{\mathbf{y}}_{\mathrm{s}} &= \mathcal{L}_{\mathbf{f}_{\mathrm{s}}} \mathbf{h}_\mathrm{s}(\mathbf{x}_\mathrm{s}) \text{~,}\label{eqFeedbackSspDiffDot}\\
		\ddot{\mathbf{y}}_{\mathrm{s}} &= \mathcal{L}^2_{\mathbf{f}_{\mathrm{s}}} \mathbf{h}_\mathrm{s}(\mathbf{x}_\mathrm{s}) + 
			\mathcal{L}_{\mathbf{g}_{\mathrm{s}}} \mathcal{L}_{\mathbf{f}_{\mathrm{s}}} \mathbf{h}_\mathrm{s}(\mathbf{x}_\mathrm{s})\, \mathbf{u}\text{~.}\label{eqFeedbackSspDiffDdot}
		\end{align}
	\end{subequations}
	As discussed in \cite{HZDwalker1}, the local coordinate transformation $\mathbf{\Phi}_\mathrm{s}(\mathbf{q}_\mathrm{s}) = [\boldsymbol{\eta}_{\mathrm{s},1}^\top,\xi_{\mathrm{s},1}]^\top=[\mathbf{h}_\mathrm{s}^\top,\theta(\mathbf{q}_\mathrm{s})]^\top$ is a diffeomorphism which transforms the system dynamics into the Byrnes–Isidori form
	\begin{equation}\label{eqBInormalformSsp}
		\begin{bmatrix}
			\dot{\boldsymbol{\eta}}_{\mathrm{s},1} \\ \dot{\boldsymbol{\eta}}_{\mathrm{s},2} \\ \dot{\xi}_{\mathrm{s},1} \\ 
			\dot{\xi}_{\mathrm{s},2} \end{bmatrix} = \begin{bmatrix} \boldsymbol{\eta}_{\mathrm{s},2} \\ \mathcal{L}^2_{\mathbf{f}_{\mathrm{s}}} 
			\mathbf{h}_\mathrm{s} \\ 
			\mathcal{L}_{\mathbf{f}_{\mathrm{s}}} \theta \\ 
			\mathcal{L}_{\mathbf{f}_{\mathrm{s}}} \sigma
		\end{bmatrix}+\begin{bmatrix} \mathbf{0} \\ \mathcal{L}_{\mathbf{g}_{\mathrm{s}}} 
		\mathcal{L}_{\mathbf{f}_{\mathrm{s}}} \mathbf{h}_\mathrm{s} \\ 0 \\ 0
		\end{bmatrix}\mathbf{u}\text{~,}
	\end{equation}
	where $\boldsymbol{\eta}_{\mathrm{s},2}=\mathcal{L}_{\mathbf{f}_{\mathrm{s}}} \mathbf{h}_\mathrm{s}$, $\xi_{\mathrm{s},2}=\sigma_\mathrm{s}(\mathbf{q}_\mathrm{s},\dot{\mathbf{q}}_\mathrm{s})$ and
	\begin{equation}\label{eqMomentumSsp}
		\sigma_\mathrm{s}=\underbrace{\begin{bmatrix} 1&\mathbf{0}_{1\times 4} \end{bmatrix} \mathbf{M}_\mathrm{s}(\mathbf{q}_\mathrm{s})}_{\boldsymbol{\gamma}_{\mathrm{s},0}(\mathbf{q}_\mathrm{s})} \dot{\mathbf{q}}_\mathrm{s}
	\end{equation}
	is the constrained system's generalized momentum conjugate to $\theta$. Feedback linearization is achieved by introducing the new control input $\mathbf{v}_{\mathrm{s}}=\ddot{\mathbf{y}}_{\mathrm{s}}$ which is related to the output by a simple double integrator and to the original inputs $\mathbf{u}$ via \eqref{eqFeedbackSspDiffDdot}:
	\begin{equation}\label{eqBMotorssp}
		\mathbf{u} = \left(\mathcal{L}_{\mathbf{g}_{\mathrm{s}}} \mathcal{L}_{\mathbf{f}_{\mathrm{s}}} \mathbf{h}_\mathrm{s}(\mathbf{x}_\mathrm{s})\right)^{-1} \left(\mathbf{v}_\mathrm{s}-\mathcal{L}^2_{\mathbf{f}_{\mathrm{s}}} \mathbf{h}_\mathrm{s}(\mathbf{x}_\mathrm{s})\right) \text{~.}
	\end{equation}
	The control error $\mathbf{y}$ of the linearized model is zeroed via PD control $\mathbf{v}_{\mathrm{s}} = -\mathbf{K}_\mathrm{P,s\,}\mathbf{y}_{\mathrm{s}} - \mathbf{K}_\mathrm{D,s\,}\dot{\mathbf{y}}_{\mathrm{s}}$, where $\mathbf{K}_\mathrm{P,s}$ and $\mathbf{K}_\mathrm{D,s}$ are positive definite control gains to achieve asymptotic stability. Since $\mathcal{L}_{\mathbf{g}_{\mathrm{s}}}\theta=0$ and $\mathcal{L}_{\mathbf{g}_{\mathrm{s}}}\sigma_\mathrm{s}=0$, the dynamics that correspond to $\xi_{\mathrm{s},1}=\theta$ and $\xi_{\mathrm{s},2}=\sigma_\mathrm{s}$ are not controllable through the input $\mathbf{u}$. The zero dynamics of the controlled system in the SSP follow from \eqref{eqBInormalformSsp} and \eqref{eqBMotorssp} for $\mathbf{y}_{\mathrm{s}} \equiv \mathbf{0}$. In this case, $\mathbf{v}_{\mathrm{s}}^\ast=\mathbf{0}$,
	\begin{equation}\label{eqBMotorsspZd}
		\mathbf{u}^* = -\left(\mathcal{L}_{\mathbf{g}_{\mathrm{s}}} \mathcal{L}_{\mathbf{f}_{\mathrm{s}}} \mathbf{h}_\mathrm{s}(\mathbf{x}_\mathrm{s})\right)^{-1} \mathcal{L}^2_{\mathbf{f}_{\mathrm{s}}} \mathbf{h}_\mathrm{s}(\mathbf{x}_\mathrm{s})
	\end{equation}
	and the dynamics via \eqref{eqBInormalformSsp} are restricted to the zero dynamics manifold: the smooth two dimensional submanifold $\mathcal{Z}_\mathrm{s}=\{\mathbf{x}_\mathrm{s}\in T\mathcal{Q}_{\mathrm{s}}|\,\mathbf{h}_\mathrm{s}(\mathbf{x}_\mathrm{s})=\mathbf{0}, \mathcal{L}_{\mathbf{f}_{\mathrm{s}}}\mathbf{h}_\mathrm{s}(\mathbf{x}_\mathrm{s})=\mathbf{0}\}$. Since $\boldsymbol{\eta}_{\mathrm{s},1}=\mathbf{0}$ and $\boldsymbol{\eta}_{\mathrm{s},2}=\mathbf{0}$, the zero dynamics $\dot{\mathbf{z}}_\mathrm{s}=\mathbf{f}_{\mathrm{s,zero}}(\mathbf{z}_\mathrm{s})=[\mathcal{L}_{\mathbf{f}_{\mathrm{s}}}\theta\rvert_{\mathcal{Z}_\mathrm{s}},\mathcal{L}_{\mathbf{f}_{\mathrm{s}}}\sigma_\mathrm{s}\rvert_{\mathcal{Z}_\mathrm{s}}]^\top$ with $\mathbf{z}_\mathrm{s}=[\xi_{\mathrm{s},1},\xi_{\mathrm{s},2}]$ in the SSP are given by the restriction of the internal dynamics to the zero dynamics manifold $\mathcal{Z}_\mathrm{s}$.
	
	As shown in \cite[(28)\,--\,(33)]{HZDwalker1}, the zero dynamics in the SSP can be expressed as
	\begin{subequations}\label{eqZerodynamicsSsp}
    	\begin{align}
    		\dot{\xi}_{\mathrm{s},1} &= \kappa_{\mathrm{s},1}(\xi_{\mathrm{s},1})\, \xi_{\mathrm{s},2} \text{~,}\label{eqZerodynamicsSsp1}\\
    		\dot{\xi}_{\mathrm{s},2} &= \kappa_{\mathrm{s},2}(\xi_{\mathrm{s},1})\label{eqZerodynamicsSsp2}
    	\end{align}
	\end{subequations}
	with
	\begin{equation}\label{eqZerodynamicsDerive}
	    \begin{aligned}
            \kappa_{\mathrm{s},1}(\xi_{\mathrm{s},1}) &= \left.\frac{\partial\theta}{\partial\mathbf{q}_\mathrm{s}}\begin{bmatrix}
                \frac{\partial\mathbf{h}_\mathrm{s}}{\partial \mathbf{q}_\mathrm{s}}\\ \boldsymbol{\gamma}_{\mathrm{s},0}
			\end{bmatrix}^{-1}\begin{bmatrix}\mathbf{0}_{4\times 1}\\1\end{bmatrix}\right\rvert_{\mathcal{Z}_\mathrm{s}}\text{~,}\\
			\kappa_{\mathrm{s},2}(\xi_{\mathrm{s},1}) &= \left.-\frac{\partial V_\mathrm{s}}{\partial\theta}\right\rvert_{\mathcal{Z}_\mathrm{s}}\text{~,}
	    \end{aligned}
	\end{equation}
	where $V_\mathrm{s}$ is the potential energy of the SSP system due to gravity. The derivation of \eqref{eqZerodynamicsSsp} uses the fact that in the SSP, $\theta$ is a cyclic variable which means that $\partial\mathbf{M}_{\mathbf{s}}/\partial\theta=\mathbf{0}$.
	
\subsection{DSP Controller with Underactuation}\label{labSubsectionDSPunderact}
    The philosophy that underlies HZD control for bipedal locomotion is to design a controller for an underactuated system in such a way that (stable) limit cycles of the hybrid dynamic system with control correspond to the desired mode of locomotion. In this section, the idea is to design a controller for the DSP that also produces an underactuated system so that the periodic sequence of \ldots$\rightarrow$DSP$\rightarrow$LO$\rightarrow$SSP$\rightarrow$TD$\rightarrow$\ldots has one unactuated DoF in every continuous phase and there are limit cycle solutions for the zero dynamics of the hybrid dynamic system with control. To achieve a DSP with one unactuated DoF, two independent virtual inputs $\tilde{\mathbf{u}}=[\tilde{u}_1,\tilde{u}_2]^\top$ are introduced and mapped to the physical inputs via
	\begin{equation}\label{eqProjectActuatorUnder}
		\mathbf{u} = \mathbf{P}_\mathrm{u} \tilde{\mathbf{u}}\text{~.}
	\end{equation}
	The mapping is a projection based on a constant $4\times 2$ matrix $\mathbf{P}_\mathrm{u}=[\mathbf{P}_{\mathrm{u},1},\mathbf{P}_{\mathrm{u},2}]$ with rank 2, where we further require the norm of each column $\lVert\mathbf{P}_\mathrm{u,i}\rVert_2=1$, $i\in\{1,2\}$ to be one to reduce ambiguity due to the otherwise arbitrary scaling of the virtual inputs. The equation of motion \eqref{eqODEdspTheta} is then
	\begin{equation}\label{eqODEdspThetaProject}
		\mathbf{M}_{\mathrm{d}} \ddot{\mathbf{q}}_\mathrm{d} + \mathbf{\Gamma}_\mathrm{d} = \underbrace{\mathbf{B}_\mathrm{d} \mathbf{P}_\mathrm{u}}_{\tilde{\mathbf{B}}_\mathrm{d}} \tilde{\mathbf{u}} \text{~,}
	\end{equation}
	with $\mathrm{rank}(\tilde{\mathbf{B}}_\mathrm{d}) = 2$. Although there is one unactuated DoF since there are only two inputs $\tilde{\mathbf{u}}$, $\tilde{\mathbf{B}}_\mathrm{d}$ maps both inputs to all three lines of \eqref{eqODEdspThetaProject}. In order to uncover the dynamics of the unactuated DoF, an orthogonal matrix $\mathbf{Q}_\mathrm{d}=\mathbf{G}_3\mathbf{G}_2\mathbf{G}_1$ that performs three Givens rotations \cite[section~\mbox{V}]{MatrixCite} is multiplied to \eqref{eqODEdspThetaProject} from the left. By choice of the Givens rotations $\mathbf{G}_i$, $i\in\{1,2,3\}$, the input matrix is transformed into lower triangular form $\tilde{\mathbf{L}}_\mathrm{d}=\mathbf{Q}_\mathrm{d}\tilde{\mathbf{B}}_\mathrm{d}$ and the transformed equation is
	\begin{equation}\label{eqODEdspThetaProjectQR}
		\tilde{\mathbf{M}}_{\mathrm{d}} \ddot{\mathbf{q}}_\mathrm{d} + \tilde{\mathbf{\Gamma}}_{\mathrm{d}} = \begin{bmatrix} 0&0\\ 0&\tilde{L}_{\mathrm{d},22}\\ \tilde{L}_{\mathrm{d},31}&\tilde{L}_{\mathrm{d},32} 
		\end{bmatrix} \tilde{\mathbf{u}} \text{~,}
	\end{equation}
	with $\tilde{\mathbf{M}}_{\mathrm{d}}=\mathbf{Q}_\mathrm{d}\mathbf{M}_{\mathrm{d}}$ and $\tilde{\mathbf{\Gamma}}_{\mathrm{d}}=\mathbf{Q}_\mathrm{d}\mathbf{\Gamma}_{\mathrm{d}}$. The comparison of \eqref{eqODEdspThetaProjectQR} to \eqref{eqODEssp} shows that this equation of motion has the same structure that was used to design the SSP controller---namely there are no inputs in the first line and the unactuated DoF is again $\theta$. The DSP dynamics can be expressed in state space as
	\begin{equation}\label{eqStateSpacedspReduced}
		\dot{\mathbf{x}}_\mathrm{d} = \begin{bmatrix}\dot{\mathbf{q}}_\mathrm{d}\\\mathbf{M}_{\mathrm{d}}^{-1}(-\mathbf{\Gamma}_\mathrm{d}+\tilde{\mathbf{B}}_\mathrm{d}\tilde{\mathbf{u}})\end{bmatrix}
		= \underbrace{\begin{bmatrix}\dot{\mathbf{q}}_\mathrm{d}\\-\mathbf{M}_{\mathrm{d}}^{-1}\mathbf{\Gamma}_\mathrm{d}\end{bmatrix}}_{\mathbf{f}_{\mathrm{d}}(\mathbf{x}_\mathrm{d})}
		+ \underbrace{\begin{bmatrix}\mathbf{0}_{3\times 2}\\\mathbf{M}_{\mathrm{d}}^{-1}\tilde{\mathbf{B}}_\mathrm{d}\end{bmatrix}}_{\mathbf{g}_{\mathrm{d}}(\mathbf{x}_\mathrm{d})}\tilde{\mathbf{u}}\text{~.}
	\end{equation}
	
\subsubsection{Control Design}\label{labDSPunderactControl}
	\begin{figure}[ht]
	\centering
	\includegraphics[width=0.6\linewidth]{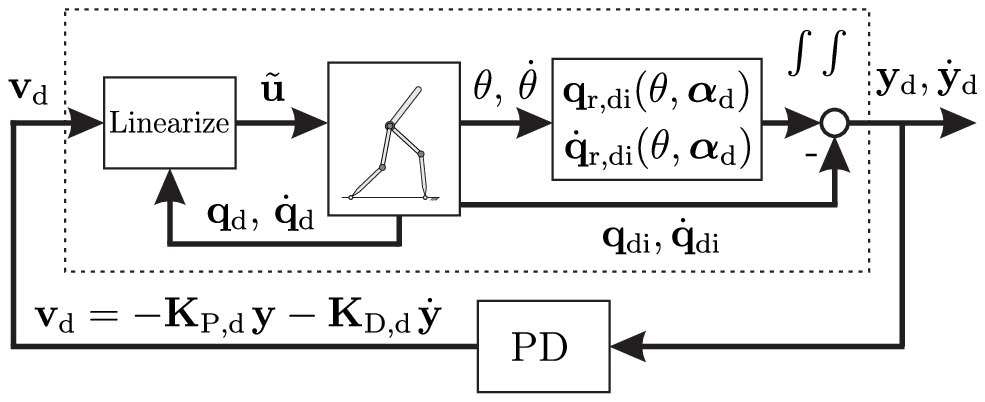}
	\caption{Feedback control in the DSP}
	\label{figFeedback}
	\end{figure}
	As depicted in Figure \ref{figFeedback}, the control output is defined as
	\begin{equation}\label{eqFeedbackunderactuatedDsp}
		\mathbf{y}_\mathrm{d} = \mathbf{h}_\mathrm{d}(\mathbf{x}_\mathrm{d})=\mathbf{q}_\mathrm{di} - \mathbf{q}_\mathrm{r,di}(\theta,\boldsymbol{\alpha}_\mathrm{d})\text{~,}
	\end{equation}
	where the reference trajectories for the independent joint angles $\mathbf{q}_\mathrm{r,di}(\theta,\boldsymbol{\alpha}_\mathrm{d})$ are parametric functions of the state $\theta$ and the constant parameters $\boldsymbol{\alpha}_\mathrm{d}$. Differentiating the output twice yields
	\begin{subequations}\label{eqFeedbackDspDiff}
		\begin{align}
		\dot{\mathbf{y}}_\mathrm{d} &= \mathcal{L}_{\mathbf{f}_{\mathrm{d}}}\mathbf{h}_\mathrm{d}(\mathbf{x}_\mathrm{d})\text{~,}\label{eqFeedbackDspDiffDot}\\
		\ddot{\mathbf{y}}_\mathrm{d} &= \mathcal{L}^2_{\mathbf{f}_{\mathrm{d}}} \mathbf{h}_\mathrm{d}(\mathbf{x}_\mathrm{d}) + 
			\mathcal{L}_{\mathbf{g}_{\mathrm{d}}} \mathcal{L}_{\mathbf{f}_{\mathrm{d}}} \mathbf{h}_\mathrm{d}(\mathbf{x}_\mathrm{d})\,\tilde{\mathbf{u}}\text{~,}\label{eqFeedbackDspDiffDdot}
		\end{align}
	\end{subequations}
	and the diffeomorphism\footnote{The local transformation is a diffeomorphism since its Jacobian $\frac{\partial \mathbf{\Phi}_\mathrm{d}}{\partial \mathbf{q}_\mathrm{d}}$ has full rank as can be verified via direct calculation of its determinant.} $\mathbf{\Phi}_\mathrm{d}(\mathbf{q}_\mathrm{d})=[\boldsymbol{\eta}_{\mathrm{d},1}^\top,\xi_{\mathrm{d},1}]^\top=[\mathbf{h}_\mathrm{d}^\top,\theta(\mathbf{q}_\mathrm{d})]^\top$ is used to transform the dynamics \eqref{eqStateSpacedspReduced} into the Byrnes–Isidori form
	\begin{equation}\label{eqBInormalformDsp}
		\begin{bmatrix}
			\dot{\boldsymbol{\eta}}_{\mathrm{d},1} \\ \dot{\boldsymbol{\eta}}_{\mathrm{d},2} \\ 
			\dot{\xi}_{\mathrm{d},1} \\ \dot{\xi}_{\mathrm{d},2}
		\end{bmatrix} = \begin{bmatrix} \boldsymbol{\eta}_{\mathrm{d},2} \\ \mathcal{L}^2_{\mathbf{f}_{\mathrm{d}}} \mathbf{h}_\mathrm{d} \\ \mathcal{L}_{\mathbf{f}_{\mathrm{d}}} \theta \\ 
			\mathcal{L}_{\mathbf{f}_{\mathrm{d}}} \tilde{\sigma}
		\end{bmatrix}+\begin{bmatrix} \mathbf{0} \\ \mathcal{L}_{\mathbf{g}_{\mathrm{d}}} \mathcal{L}_{\mathbf{f}_{\mathrm{d}}} \mathbf{h}_\mathrm{d} \\ 0 \\ 0
		\end{bmatrix} \tilde{\mathbf{u}} \text{~.}
	\end{equation}
	Here $\boldsymbol{\eta}_{\mathrm{d},2}=\mathcal{L}_{\mathbf{f}_{\mathrm{d}}} \mathbf{h}_\mathrm{d}$, $\xi_{\mathrm{d},2}=\tilde{\sigma}_\mathrm{d}(\mathbf{q}_\mathrm{d},\dot{\mathbf{q}}_\mathrm{d})$ and
	\begin{equation}\label{eqMomentumDsp}
		\tilde{\sigma}_\mathrm{d}=\underbrace{\begin{bmatrix} 1&\mathbf{0}_{1\times 2} \end{bmatrix} \tilde{\mathbf{M}}_\mathrm{d}(\mathbf{q}_\mathrm{d})}_{\tilde{\boldsymbol{\gamma}}_{\mathrm{d},0}(\mathbf{q}_\mathrm{d})} \dot{\mathbf{q}}_\mathrm{d}\text{~.}
	\end{equation}
	Feedback linearization is accomplished by introducing the new control input $\mathbf{v}_{\mathrm{d}}=\ddot{\mathbf{y}}_{\mathrm{d}}$ which is related to the output by a simple double integrator and to the virtual inputs $\tilde{\mathbf{u}}$ via \eqref{eqFeedbackDspDiffDdot}:
	\begin{equation}\label{eqBMotordsp}
		\tilde{\mathbf{u}} = \left(\mathcal{L}_{\mathbf{g}_{\mathrm{d}}} \mathcal{L}_{\mathbf{f}_{\mathrm{d}}} \mathbf{h}_\mathrm{d}(\mathbf{x}_\mathrm{d})\right)^{-1} \left(\mathbf{v}_\mathrm{d}-\mathcal{L}^2_{\mathbf{f}_{\mathrm{d}}} \mathbf{h}_\mathrm{d}(\mathbf{x}_\mathrm{d})\right) \text{~.}
	\end{equation}
	A PD controller with $\mathbf{v}_\mathrm{d} = -\mathbf{K}_\mathrm{P,d\,}\mathbf{y}-\mathbf{K}_\mathrm{D,d\,}\dot{\mathbf{y}}$ and positive definite gains $\mathbf{K}_\mathrm{P,d}$ and $\mathbf{K}_\mathrm{D,d}$ is introduced to achieve asymptotic stability for the linearized system. Analogously to the SSP controller, $\mathcal{L}_{\mathbf{g}_{\mathrm{d}}}\theta=0$ and $\mathcal{L}_{\mathbf{g}_{\mathrm{d}}}\tilde{\sigma}_\mathrm{d}=0$, the dynamics that correspond to $\xi_{\mathrm{d},1}=\theta$ and $\xi_{\mathrm{d},2}=\tilde{\sigma}_\mathrm{d}$ are not controllable through the virtual inputs $\tilde{\mathbf{u}}$. The zero dynamics of the controlled system in the DSP follow from \eqref{eqBInormalformDsp} and \eqref{eqBMotordsp} for $\mathbf{y}_{\mathrm{d}} \equiv \mathbf{0}$. In this case, $\mathbf{v}_{\mathrm{d}}^\ast=\mathbf{0}$,
	\begin{equation}\label{eqBMotordspZd}
		\tilde{\mathbf{u}}^* = -\left(\mathcal{L}_{\mathbf{g}_{\mathrm{d}}} \mathcal{L}_{\mathbf{f}_{\mathrm{d}}} \mathbf{h}_\mathrm{d}(\mathbf{x}_\mathrm{d})\right)^{-1} \mathcal{L}^2_{\mathbf{f}_{\mathrm{d}}} \mathbf{h}_\mathrm{d}(\mathbf{x}_\mathrm{d})
	\end{equation}
	and the dynamics via \eqref{eqBInormalformDsp} are restricted to the zero dynamics manifold: the smooth two dimensional submanifold $\mathcal{Z}_\mathrm{d}=\{\mathbf{x}_\mathrm{d}\in T\mathcal{Q}_{\mathrm{d}}|\,\mathbf{h}_\mathrm{d}(\mathbf{x}_\mathrm{d})=\mathbf{0}, \mathcal{L}_{\mathbf{f}_{\mathrm{d}}}\mathbf{h}_\mathrm{d}(\mathbf{x}_\mathrm{d})=\mathbf{0}\}$. Since $\boldsymbol{\eta}_{\mathrm{d},1}=\mathbf{0}$ and $\boldsymbol{\eta}_{\mathrm{d},2}=\mathbf{0}$, the zero dynamics $\dot{\mathbf{z}}_\mathrm{d}=\mathbf{f}_{\mathrm{d,zero}}(\mathbf{z}_\mathrm{d})=[\mathcal{L}_{\mathbf{f}_{\mathrm{d}}}\theta\rvert_{\mathcal{Z}_\mathrm{d}},\mathcal{L}_{\mathbf{f}_{\mathrm{d}}}\tilde{\sigma}_\mathrm{d}\rvert_{\mathcal{Z}_\mathrm{d}}]^\top$ with $\mathbf{z}_\mathrm{d}=[\xi_{\mathrm{d},1},\xi_{\mathrm{d},2}]$ in the DSP are given by the restriction of the internal dynamics to the zero dynamics manifold $\mathcal{Z}_\mathrm{d}$. Using the abbreviation
	\begin{equation*}
	    \mathbf{q}_\mathrm{r,di}^\prime(\theta,\boldsymbol{\alpha}_\mathrm{d}) = \frac{\partial}{\partial\theta}\mathbf{q}_\mathrm{r,di}(\theta,\boldsymbol{\alpha}_\mathrm{d})\text{~,}
	\end{equation*}
	this restriction gives
	\begin{equation}\label{eqZerodynamicsCoordinatesDsp}
		\begin{aligned}
		\left.\mathbf{q}_\mathrm{d}\right\rvert_{\mathcal{Z}_\mathrm{d}} &= \mathbf{\Phi}_\mathrm{d}^{-1} \left( \begin{bmatrix} \mathbf{0}_{2\times 1} \\ \xi_{\mathrm{d},1} \end{bmatrix} \right) \bigg\rvert_{\mathcal{Z}_\mathrm{d}}\\
		&= \begin{bmatrix} \xi_{\mathrm{d},1}\\ \mathbf{q}_\mathrm{r,di}(\xi_{\mathrm{d},1},\boldsymbol{\alpha}_\mathrm{d}) \end{bmatrix} \text{~,} \\
		\left.\tilde{\sigma}_\mathrm{d}\right\rvert_{\mathcal{Z}_\mathrm{d}} &= \left.\left(\tilde{\boldsymbol{\gamma}}_{\mathrm{d},0}(\mathbf{q}_\mathrm{d})\,\dot{\mathbf{q}}_\mathrm{d}\right)\right\rvert_{\mathcal{Z}_\mathrm{d}}\\
		&= \underbrace{\tilde{\boldsymbol{\gamma}}_{\mathrm{d},0}(\left.\mathbf{q}_\mathrm{d}\right\rvert_{\mathcal{Z}_\mathrm{d}}) \begin{bmatrix} 1\\ \mathbf{q}_\mathrm{r,di}^\prime(\xi_{\mathrm{d},1},\boldsymbol{\alpha}_\mathrm{d}) \end{bmatrix}}_{\kappa_{\mathrm{d},1}^{-1}(\xi_{\mathrm{d},1})}\dot{\xi}_{\mathrm{d},1}\text{~,}\\
		\left.\dot{\mathbf{q}}_\mathrm{d}\right\rvert_{\mathcal{Z}_\mathrm{d}} &= \begin{bmatrix} 1\\ \mathbf{q}_\mathrm{r,di}^\prime(\xi_{\mathrm{d},1},\boldsymbol{\alpha}_\mathrm{d}) \end{bmatrix}\dot{\xi}_{\mathrm{d},1}\\
		&= \kappa_{\mathrm{d},1}(\xi_{\mathrm{d},1})\begin{bmatrix} 1\\ \mathbf{q}_\mathrm{r,di}^\prime(\xi_{\mathrm{d},1},\boldsymbol{\alpha}_\mathrm{d}) \end{bmatrix}\xi_{\mathrm{d},2} \text{~,}
		\end{aligned}
	\end{equation}
	which in combination with \eqref{eqODEdspThetaProjectQR} yields the zero dynamics
	\begin{subequations}\label{eqZerodynamicsDsp}
		\begin{align}
		\dot{\xi}_{\mathrm{d},1} &= \kappa_{\mathrm{d},1}(\xi_{\mathrm{d},1})\, \xi_{\mathrm{d},2} \text{~,} \label{eqZerodynamicsDspDerive1}\\
		\dot{\xi}_{\mathrm{d},2} &= \left.\left(\dot{\mathbf{q}}_\mathrm{d}^\top\,\dot{\tilde{\boldsymbol{\gamma}}}_{\mathrm{d},0}^\top(\mathbf{q}_\mathrm{d}) + \tilde{\boldsymbol{\gamma}}_{\mathrm{d},0}(\mathbf{q}_\mathrm{d})\,\ddot{\mathbf{q}}_\mathrm{d}\right)\right\rvert_{\mathcal{Z}_\mathrm{d}}\notag\\
		    &= \left.\left(\dot{\mathbf{q}}_\mathrm{d}^\top\frac{\partial\tilde{\boldsymbol{\gamma}}_{\mathrm{d},0}^\top}{\partial\mathbf{q}_\mathrm{d}}\dot{\mathbf{q}}_\mathrm{d} - \begin{bmatrix} 1&\mathbf{0}_{1\times 2} \end{bmatrix} \tilde{\boldsymbol{\Gamma}}_\mathrm{d}(\mathbf{q}_\mathrm{d},\dot{\mathbf{q}}_\mathrm{d})\right)\right\rvert_{\mathcal{Z}_\mathrm{d}}\notag\\
			&= \kappa_{\mathrm{d},2}(\xi_{\mathrm{d},1})+\kappa_{\mathrm{d},3}(\xi_{\mathrm{d},1})\,\xi_{\mathrm{d},2}^2 \text{~.}\label{eqZerodynamicsDspDerive2}
		\end{align}
	\end{subequations}
	The expressions for $\kappa_{\mathrm{d},2}(\xi_{\mathrm{d},1})$ and $\kappa_{\mathrm{d},3}(\xi_{\mathrm{d},1})$ are rather long and not stated explicitly. The structure of \eqref{eqZerodynamicsDspDerive2} is not as simple as that of \eqref{eqZerodynamicsSsp2} for the SSP since $\theta$ is not a cyclic variable in the DSP. However, the structure is analogous to that of the SSP zero dynamics of a system with curved feet as e.\,g.\ described in \cite{HZDwalker3}.
	
\subsubsection{Hybrid Zero Dynamics}\label{labDSPunderactHZD}
    The hybrid zero dynamics, the zero dynamics of the hybrid system \eqref{eqHybridSystem}, are obtained by combining the zero dynamics of the continuous SSP and DSP via \eqref{eqZerodynamicsSsp} and \eqref{eqZerodynamicsDsp} in such a way, that the invariant zero dynamics manifolds $\mathcal{Z}_\mathrm{s}$ and $\mathcal{Z}_\mathrm{d}$ are mapped onto each other at the discrete transition events. This is achieved by choosing appropriate parametrizations for the SSP and DSP reference trajectories $\mathbf{q}_\mathrm{r,s}(\theta,\boldsymbol{\alpha}_\mathrm{s})$ and $\mathbf{q}_\mathrm{r,di}(\theta,\boldsymbol{\alpha}_\mathrm{d})$. Evaluating the mappings \eqref{eqSwitchLegsCoordState} and \eqref{eqLiftoffState} with states on both zero dynamics manifolds gives
    \begin{subequations}\label{eqInvariantMappings}
        \begin{align}
        \begin{bmatrix}\xi_{\mathrm{d},1}^+\\\mathbf{q}_\mathrm{r,di}(\xi_{\mathrm{d},1}^+,\boldsymbol{\alpha}_\mathrm{d})\end{bmatrix} &= \mathbf{H}_\mathrm{d}\mathbf{R}_\mathrm{s}^\mathrm{d}\mathbf{H}_\mathrm{s}^{-1}\begin{bmatrix}\xi_{\mathrm{s},1}^-\\\mathbf{q}_\mathrm{r,s}(\xi_{\mathrm{s},1}^-,\boldsymbol{\alpha}_\mathrm{s})\end{bmatrix}\text{~,}\label{eqInvariantMappings1}\\
		\begin{bmatrix}\xi_{\mathrm{s},1}^+\\\mathbf{q}_\mathrm{r,s}(\xi_{\mathrm{s},1}^+,\boldsymbol{\alpha}_\mathrm{s})\end{bmatrix} &= \mathbf{H}_\mathrm{s}\mathbf{R}_\mathrm{d}^\mathrm{s}\begin{bmatrix}
			\mathbf{H}_\mathrm{d}^{-1}\begin{bmatrix}\xi_{\mathrm{d},1}^-\\\mathbf{q}_\mathrm{r,di}(\xi_{\mathrm{d},1}^-,\boldsymbol{\alpha}_\mathrm{d})\end{bmatrix}\\ \mathbf{\Omega}\left(\begin{bmatrix}\xi_{\mathrm{d},1}^-\\\mathbf{q}_\mathrm{r,di}(\xi_{\mathrm{d},1}^-,\boldsymbol{\alpha}_\mathrm{d})\end{bmatrix}\right)
		\end{bmatrix}\text{~,}\label{eqInvariantMappings2}\\
		\begin{bmatrix} 1\\ \mathbf{q}_\mathrm{r,di}^\prime(\xi_{\mathrm{d},1}^+,\boldsymbol{\alpha}_\mathrm{d}) \end{bmatrix}\dot{\xi}_{\mathrm{d},1}^+ &= \mathbf{H}_\mathrm{d}\mathbf{\Delta}_{\mathrm{s}}\mathbf{H}_\mathrm{s}^{-1}\begin{bmatrix} 1\\ \mathbf{q}_\mathrm{r,s}^\prime(\xi_{\mathrm{s},1}^-,\boldsymbol{\alpha}_\mathrm{s}) \end{bmatrix}\dot{\xi}_{\mathrm{s},1}^-\text{~,}\label{eqInvariantMappings3}\\
		\begin{bmatrix} 1\\ \mathbf{q}_\mathrm{r,s}^\prime(\xi_{\mathrm{s},1}^+,\boldsymbol{\alpha}_\mathrm{s}) \end{bmatrix}\dot{\xi}_{\mathrm{s},1}^+&= \mathbf{H}_\mathrm{s}\mathbf{R}_\mathrm{d}^\mathrm{s}\begin{bmatrix}\mathbf{H}_\mathrm{d}^{-1}\\ \mathbf{J}_{\mathbf{\Omega}}\left(\begin{bmatrix}\xi_{\mathrm{d},1}^-\\\mathbf{q}_\mathrm{r,di}(\xi_{\mathrm{d},1}^-,\boldsymbol{\alpha}_\mathrm{d})\end{bmatrix}\right)\end{bmatrix}\!\!\begin{bmatrix} 1\\ \mathbf{q}_\mathrm{r,di}^\prime(\xi_{\mathrm{d},1}^-,\boldsymbol{\alpha}_\mathrm{d}) \end{bmatrix}\dot{\xi}_{\mathrm{d},1}^-\text{~.}\label{eqInvariantMappings4}
        \end{align}
    \end{subequations}
    Equations \eqref{eqInvariantMappings1} and \eqref{eqInvariantMappings2} give necessary conditions for the configuration at the beginning of the SSP (of the DSP) as functions of the configuration at the end of the DSP (of the SSP). The first lines of \eqref{eqInvariantMappings3} and \eqref{eqInvariantMappings4} give relations for the velocities on the zero dynamics manifold, which can be stated as
    \begin{subequations}\label{eqInvariantVelocityMappings}
        \begin{align}
            \dot{\xi}_{\mathrm{d},1}^+ &= \underbrace{\begin{bmatrix}1&\mathbf{0}_{1\times2}\end{bmatrix}\mathbf{H}_\mathrm{d}\mathbf{\Delta}_{\mathrm{s}}\mathbf{H}_\mathrm{s}^{-1}\begin{bmatrix} 1\\ \mathbf{q}_\mathrm{r,s}^\prime(\xi_{\mathrm{s},1}^-,\boldsymbol{\alpha}_\mathrm{d}) \end{bmatrix}}_{\tilde{\delta}_{\mathrm{s}}^{\mathrm{d}}}\dot{\xi}_{\mathrm{s},1}^-\text{~,}\label{eqInvariantVelocityMappings1}\\
            \dot{\xi}_{\mathrm{s},1}^+ &= \dot{\xi}_{\mathrm{d},1}^-\text{~.}\label{eqInvariantVelocityMappings2}
        \end{align}
    \end{subequations}
    Substituting these relations into \eqref{eqInvariantMappings3} and \eqref{eqInvariantMappings4} and eliminating $\dot{\theta}_{\mathrm{s}}^-$ and $\dot{\theta}_{\mathrm{d}}^-$, the respective second to last lines yield necessary conditions for the first derivatives of the reference trajectories at the beginning of the DSP and SSP, respectively. Given reference trajectories that fulfill all necessary conditions, the zero dynamics of the SSP and DSP are mapped onto each other at both discrete transition events. These mappings can also be formulated for the states used in \eqref{eqZerodynamicsSsp} and \eqref{eqZerodynamicsDsp}. Equations \eqref{eqInvariantMappings} and \eqref{eqInvariantVelocityMappings} give
    \begin{subequations}\label{eqInvariantZDStateMappings}
        \begin{align}
            \xi_{\mathrm{d},1}^+ &= \underbrace{\begin{bmatrix}1&\mathbf{0}_{1\times2}\end{bmatrix}\mathbf{H}_\mathrm{d}\mathbf{R}_\mathrm{s}^\mathrm{d}\mathbf{H}_\mathrm{s}^{-1}}_{\begin{bmatrix}1&-1&1&-\frac{1}{2}&\frac{1}{2}\end{bmatrix}}\begin{bmatrix}\xi_{\mathrm{s},1}^-\\\mathbf{q}_\mathrm{r,s}(\xi_{\mathrm{s},1}^-,\boldsymbol{\alpha}_\mathrm{s})\end{bmatrix}\text{~,}\label{eqInvariantZDStateMappings1}\\
            \xi_{\mathrm{s},1}^+ &= \underbrace{\begin{bmatrix}1&\mathbf{0}_{1\times4}\end{bmatrix}\mathbf{H}_\mathrm{s}\mathbf{R}_\mathrm{d}^\mathrm{s}}_{\begin{bmatrix}1&1&\frac{1}{2}&0&0\end{bmatrix}}\begin{bmatrix}
			\mathbf{H}_\mathrm{d}^{-1}\begin{bmatrix}\xi_{\mathrm{d},1}^-\\\mathbf{q}_\mathrm{r,di}(\xi_{\mathrm{d},1}^-,\boldsymbol{\alpha}_\mathrm{d})\end{bmatrix}\\ \mathbf{\Omega}\left(\begin{bmatrix}\xi_{\mathrm{d},1}^-\\\mathbf{q}_\mathrm{r,di}(\xi_{\mathrm{d},1}^-,\boldsymbol{\alpha}_\mathrm{d})\end{bmatrix}\right)\end{bmatrix}=\xi_{\mathrm{d},1}^-\text{~,}\label{eqInvariantZDStateMappings2}\\
			\xi_{\mathrm{d},2}^+ &= \underbrace{\frac{\kappa_{\mathrm{s},1}(\xi_{\mathrm{s},1}^-)}{\kappa_{\mathrm{d},1}(\xi_{\mathrm{d},1}^+)}\tilde{\delta}_{\mathrm{s}}^{\mathrm{d}}}_{\delta_{\mathrm{s}}^{\mathrm{d}}}\xi_{\mathrm{s},2}^-\text{~,}\label{eqInvariantZDStateMappings3}\\
			\xi_{\mathrm{s},2}^+ &= \underbrace{\frac{\kappa_{\mathrm{d},1}(\xi_{\mathrm{d},1}^-)}{\kappa_{\mathrm{s},1}(\xi_{\mathrm{s},1}^+)}}_{\delta_{\mathrm{d}}^{\mathrm{s}}}\xi_{\mathrm{d},2}^-\text{~.}\label{eqInvariantZDStateMappings4}
        \end{align}
    \end{subequations}
    By combining the SSP and DSP zero dynamics \eqref{eqZerodynamicsSsp} and \eqref{eqZerodynamicsDsp} and the mappings \eqref{eqInvariantZDStateMappings}, the hybrid zero dynamics that describe periodic walking movements of the robot system with underactuated DSP can be stated as
	\begin{equation}\label{eqHZDunderact}
		\Sigma_\mathrm{ZD} :
		\begin{cases}
			\dot{\mathbf{z}}_\mathrm{s} = \mathbf{f}_{\mathrm{s,zero}}(\mathbf{z}_\mathrm{s}) \text{~,} & \mathbf{z}_\mathrm{s} \notin 
				\mathcal{S}_\mathrm{s}^{\mathrm{d}} \cap \mathcal{Z}_\mathrm{s} \text{~,} \\
			\mathbf{z}_\mathrm{d}^{+} = \mathbf{\Delta}_\mathrm{s}^{\mathrm{d}}(\mathbf{z}_\mathrm{s}^{-}) \text{~,}  &
				\mathbf{z}_\mathrm{s}^{-} \in \mathcal{S}_\mathrm{s}^{\mathrm{d}} \cap \mathcal{Z}_\mathrm{s} \text{~,} \\
			\dot{\mathbf{z}}_\mathrm{d} = \mathbf{f}_{\mathrm{d,zero}}(\mathbf{z}_\mathrm{d}) \text{~,} &\mathbf{z}_\mathrm{d} \notin 
				\mathcal{S}_\mathrm{d}^{\mathrm{s}} \cap \mathcal{Z}_\mathrm{d} \text{~,} \\
			\mathbf{z}_\mathrm{s}^{+} = \mathbf{\Delta}_\mathrm{d}^{\mathrm{s}}(\mathbf{z}_\mathrm{d}^{-}) \text{~,} &
				\mathbf{z}_\mathrm{d}^{-} \in \mathcal{S}_\mathrm{d}^{\mathrm{s}} \cap \mathcal{Z}_\mathrm{d} \text{~.}
		\end{cases}
	\end{equation}

\subsubsection{Limit Cycle Solution and Stability}\label{labDSPunderactHZDPoincare}
    A limit cycle solution for the hybrid zero dynamics \eqref{eqHZDunderact} corresponds to an infinite sequence \ldots$\rightarrow$DSP$\rightarrow$LO$\rightarrow$SSP$\rightarrow$TD$\rightarrow$\ldots for periodic walking with underactuated DSP with zero control error. We presuppose that the phase variable $\theta(t)$ increases monotonically with time during the DSP and SSP until it is reset at the touch-down transition event. The limit cycle solution is then described by the time dependency of $\theta(t)$ or its inverse $t(\theta)$ (and derivatives thereof) that satisfies \eqref{eqHZDunderact}. The procedure from \cite{HZDwalker3} is adopted and expanded to calculate this solution and to determine its stability.
    
    Starting with the DSP, first the coordinate transformation
	\begin{equation}\label{eqCoordTransZetaDsp}
		\zeta_\mathrm{d}=\frac{1}{2}\xi_{\mathrm{d},2}^2 \quad\Rightarrow\quad \dot{\zeta}_\mathrm{d}=\xi_{\mathrm{d},2}\dot{\xi}_{\mathrm{d},2}
	\end{equation}
    is introduced and applied to the quotient of \eqref{eqZerodynamicsDspDerive2} and \eqref{eqZerodynamicsDspDerive1} which gives the linear differential equation
	\begin{equation}\label{eqSolDspZets}
		\frac{\mathrm{d} \zeta_\mathrm{d}}{\mathrm{d} \xi_{\mathrm{d},1}} = \frac{\kappa_{\mathrm{d},2}(\xi_{\mathrm{d},1})}{\kappa_{\mathrm{d},1}(\xi_{\mathrm{d},1})}+\frac{2\kappa_{\mathrm{d},3}(\xi_{\mathrm{d},1})}{\kappa_{\mathrm{d},1}(\xi_{\mathrm{d},1})}\zeta_\mathrm{d} \text{~.}
	\end{equation}
	Given the initial condition $\zeta_\mathrm{d}^+=\frac{1}{2}(\xi_{\mathrm{d},2}^+)^2$, variation of constants gives the solution
	\begin{equation}\label{eqSolDspZetsSol}
		\zeta_\mathrm{d}(\xi_{\mathrm{d},1}) = \iota_\mathrm{d}(\xi_{\mathrm{d},1})\left(\zeta_\mathrm{d}^{+}+\mu_\mathrm{d}(\xi_{\mathrm{d},1})\right)
	\end{equation}
	with 
	\begin{equation}\label{eqSolDspZetsSoliota}
		\iota_\mathrm{d}(\xi_{\mathrm{d},1}) = \exp\left(\int_{\xi_{\mathrm{d},1}^{+}}^{\xi_{\mathrm{d},1}} \frac{2\kappa_{\mathrm{d},3}(\theta)}{\kappa_{\mathrm{d},1}(\theta)} \mathrm{d} \theta \right)
	\end{equation}
	and
	\begin{equation}\label{eqSolDspZetsSolmu}
		\mu_\mathrm{d}(\xi_{\mathrm{d},1}) = \int_{\xi_{\mathrm{d},1}^{+}}^{\xi_{\mathrm{d},1}} \frac{\kappa_{\mathrm{d},2}(\theta)}{\kappa_{\mathrm{d},1}(\theta)\,\iota_\mathrm{d}(\theta)} \mathrm{d} \theta \text{~.}
	\end{equation}
	The inverse transformation gives $\xi_{\mathrm{d},2}(\xi_{\mathrm{d},1}) = \pm \sqrt{2\zeta_\mathrm{d}(\xi_{\mathrm{d},1})}$. Due to the presupposed monotonicity of $\theta(t)$, namely $\dot{\theta}=\dot{\xi}_{\mathrm{d},1} = \kappa_{\mathrm{d},1}\xi_{\mathrm{d},2}>0$ according to \eqref{eqZerodynamicsDspDerive1}, the already obtained $\kappa_{\mathrm{d},1}$ determines whether $\xi_{\mathrm{d},2}$ has a positive or negative sign. The desired solution $t(\theta)$ in the DSP is then
	\begin{equation}\label{eqIntTimeDsp}
		t_\mathrm{d}(\xi_{\mathrm{d},1}) = \int_{\xi_{\mathrm{d},1}^{+}}^{\xi_{\mathrm{d},1}} \frac{1}{\dot{\xi}_{\mathrm{d},1}(\theta)} \mathrm{d} \theta = \int_{\xi_{\mathrm{d},1}^{+}}^{\xi_{\mathrm{d},1}} \frac{1}{\kappa_{\mathrm{d},1}(\theta) \, \xi_{\mathrm{d},2}(\theta)} \mathrm{d} \theta \text{~.}
	\end{equation}
	
	The solution for the SSP is obtained in an analogous way by first introducing the transformation
	\begin{equation}\label{eqCoordTransZetaSsp}
		\zeta_\mathrm{s}=\frac{1}{2}\xi_{\mathrm{s},2}^2 \quad\Rightarrow\quad \dot{\zeta}_\mathrm{s}=\xi_{\mathrm{s},2}\dot{\xi}_{\mathrm{s},2}\text{~.}
	\end{equation}
	The quotient of \eqref{eqZerodynamicsSsp2} and \eqref{eqZerodynamicsSsp1} is then
	\begin{equation}\label{eqSolSspZets}
		\frac{\mathrm{d} \zeta_\mathrm{s}}{\mathrm{d} \xi_{\mathrm{s},1}} = \frac{\kappa_{\mathrm{s},2}(\xi_{\mathrm{s},1})}{\kappa_{\mathrm{s},1}(\xi_{\mathrm{s},1})}
	\end{equation}
	with initial condition $\zeta_\mathrm{s}^+=\frac{1}{2}(\xi_{\mathrm{s},2}^+)^2$ and solution
	\begin{align}
		\zeta_\mathrm{s}(\xi_{\mathrm{s},1}) &= \zeta_\mathrm{s}^+ + \mu_\mathrm{s}(\xi_{\mathrm{s},1}) \text{~,}\label{eqSolSspZetsSol}\\
		\mu_\mathrm{s}(\xi_{\mathrm{s},1}) &= \int_{\xi_{\mathrm{s},1}^{+}}^{\xi_{\mathrm{s},1}} \frac{\kappa_{\mathrm{s},2}(\theta)}{\kappa_{\mathrm{s},1}(\theta)} \mathrm{d} \theta \text{~.}\label{eqSolSspZetsSolmu}
	\end{align}
	The inverse transformation gives $\xi_{\mathrm{s},2}(\xi_{\mathrm{s},1}) = \sqrt{2\zeta_\mathrm{s}}(\xi_{\mathrm{s},1})$ and the desired solution $t(\theta)$ in the SSP is
	\begin{equation}\label{eqIntTimeSsp}
		t_\mathrm{s}(\xi_{\mathrm{s},1}) = \int_{\xi_{\mathrm{s},1}^{+}}^{\xi_{\mathrm{s},1}} \frac{1}{\dot{\xi}_{\mathrm{s},1}(\theta)} \mathrm{d} \theta = \int_{\xi_{\mathrm{s},1}^{+}}^{\xi_{\mathrm{s},1}} \frac{1}{\kappa_{\mathrm{s},1}(\theta) \, \xi_{\mathrm{s},2}(\theta)} \mathrm{d} \theta \text{~.}
	\end{equation}
	Since the expressions in \eqref{eqSolDspZetsSol} -- \eqref{eqIntTimeSsp} are large and complicated, we were not able to find any anti-derivatives using computer algebra systems. Therefore, the integrals are approximated numerically by quadrature with the trapezoidal rule. These two steps are summarized as a semi-analytical process. 
    
	The solution for the limit cycle---which requires appropriate values for the initial conditions $\zeta_\mathrm{d}^+$ and $\zeta_\mathrm{s}^+$---is obtained by combining the solutions for the whole SSP and DSP with the transformed mappings \eqref{eqInvariantZDStateMappings3} and \eqref{eqInvariantZDStateMappings4} for the phase transitions TD and LO:
	\begin{subequations}\label{eqZetaMappings}
	    \begin{align}
	        \zeta_{\mathrm{d}}^{+} &= (\delta_\mathrm{s}^{\mathrm{d}})^2 \zeta_{\mathrm{s}}^{-}\text{~,}\label{eqZetaMappings1}\\
	        \zeta_{\mathrm{s}}^{+} &= (\delta_\mathrm{d}^{\mathrm{s}})^2 \zeta_{\mathrm{d}}^{-}\text{~.}\label{eqZetaMappings2}
	    \end{align}
	\end{subequations}
	The initial condition for the SSP is then
	\begin{equation}\label{eqIniCondSsp}
	    \zeta_{\mathrm{s}}^{+}(\zeta_{\mathrm{d}}^{+}) = (\delta_\mathrm{d}^{\mathrm{s}})^2 \iota_\mathrm{d}(\xi_{\mathrm{d},1}^-)\left(\zeta_\mathrm{d}^{+}+\mu_\mathrm{d}(\xi_{\mathrm{d},1}^-)\right)\text{~,}
	\end{equation}
	and the mapping $P:\mathcal{Z}_\mathrm{d}\rightarrow\mathcal{Z}_\mathrm{d}, \zeta_{\mathrm{d}}^{+}\mapsto\zeta_{\mathrm{d}}^{+}$ of the initial condition $\zeta_{\mathrm{d}}^{+}$ of one period onto the beginning of the next DSP is
	\begin{equation}\label{eqIniCondDsp}
	    P\!\left(\zeta_{\mathrm{d}}^{+}\right) = (\delta_\mathrm{s}^{\mathrm{d}})^2\left((\delta_\mathrm{d}^{\mathrm{s}})^2 \iota_\mathrm{d}(\xi_{\mathrm{d},1}^-)\left(\zeta_\mathrm{d}^{+}+\mu_\mathrm{d}(\xi_{\mathrm{d},1}^-)\right) + \mu_\mathrm{s}(\xi_{\mathrm{s},1}^-)\right)\text{.}
	\end{equation}
	This mapping is a Poincaré map with one fixed point
	\begin{equation}\label{eqPmapInitialValue}
		\zeta_\mathrm{d}^{+} = \frac{(\delta_\mathrm{s}^{\mathrm{d}})^2\mu_\mathrm{s}(\xi_{\mathrm{s},1}^{-})+(\delta_\mathrm{s}^{\mathrm{d}}\delta_\mathrm{d}^{\mathrm{s}})^2\,\iota_\mathrm{d}(\xi_{\mathrm{d},1}^{-})\,\mu_\mathrm{d}(\xi_{\mathrm{d},1}^{-})}{1-(\delta_\mathrm{s}^{\mathrm{d}}\delta_\mathrm{d}^{\mathrm{s}})^2\,\iota_\mathrm{d}(\xi_{\mathrm{d},1}^{-})}\text{~,}
	\end{equation}
	which is the initial condition for the limit cycle solution. The stability of the limit cycle solution can be analyzed by the linearization of the Poincaré map at the fixed point
	\begin{equation}\label{eqFloquet}
		\Lambda = \frac{\partial P(\zeta_{\mathrm{d}}^{+})}{\partial\zeta_{\mathrm{d}}^{+}} = (\delta_\mathrm{s}^{\mathrm{d}}\delta_\mathrm{d}^{\mathrm{s}})^2 \,\iota_\mathrm{d}(\xi_{\mathrm{d},1}^{-}) \text{~,}
	\end{equation}
	where $\Lambda$ is equal to the Floquet multiplier of the limit cycle solution for \eqref{eqHZDunderact}.\footnote{Since the zero dynamics are two-dimensional, there are two Floquet multipliers. For a limit cycle solution, one of these is one, while the other one---$\Lambda$---determines the stability of the limit cycle solution.} The limit cycle is stable if $0 < \Lambda < 1$.
	
\subsection{DSP Controller with Full Actuation}\label{labSubsectionDSPfullact}
    The control design from the previous section for a DSP controller with underactuation can be expanded to use another (virtual) actuator to zero another output---which means an additional objective is added to the control strategy. The system is then fully actuated in the DSP, which means that strictly speaking there is no need for a hybrid zero dynamics controller and a different control design could be used in the DSP. However, the goal of the present manuscript is to extend the HZD control strategy for gaits with a non-instantaneous DSP which is the focus of this section. Since the use of the DSP controller with underactuation often results in unstable limit cycles, we seek an extension of the DSP controller with underactuation by use of another input that influences this stability via direct modification of the Floquet multiplier $\Lambda$. To achieve a DSP with full actuation, three independent virtual inputs $\tilde{\mathbf{u}}=[\tilde{u}_1,\tilde{u}_2,\tilde{u}_3]^\top$ are introduced and mapped to the physical inputs via
	\begin{equation}\label{eqProjectActuatorFull}
		\mathbf{u} = \mathbf{P}_\mathrm{f} \tilde{\mathbf{u}}\text{~.}
	\end{equation}
	The $4\times 3$ projection matrix $\mathbf{P}_\mathrm{f}=[\mathbf{P}_{\mathrm{f},1},\mathbf{P}_{\mathrm{f},2},\mathbf{P}_{\mathrm{f},3}]$ with rank~3 and $\lVert\mathbf{P}_\mathrm{f,i}\rVert_2=1$, $i\in\{1,2,3\}$ gives
	\begin{equation}\label{eqODEdspThetaProjectFull}
		\mathbf{M}_{\mathrm{d}} \ddot{\mathbf{q}}_\mathrm{d} + \mathbf{\Gamma}_\mathrm{d} = \underbrace{\mathbf{B}_\mathrm{d} \mathbf{P}_\mathrm{f}}_{\tilde{\mathbf{B}}_\mathrm{d}} \tilde{\mathbf{u}}
	\end{equation}
	with $\mathrm{rank}(\tilde{\mathbf{B}}_\mathrm{d}) = 3$. Again, an orthogonal matrix $\mathbf{Q}_\mathrm{d}=\mathbf{G}_3\mathbf{G}_2\mathbf{G}_1$ that performs three Givens rotations $\mathbf{G}_i$, $i\in\{1,2,3\}$ is multiplied to \eqref{eqODEdspThetaProjectFull} from the left to transform the input matrix into lower triangular shape
	\begin{equation}\label{eqODEdspThetaProjectQRFull}
		\tilde{\mathbf{M}}_{\mathrm{d}} \ddot{\mathbf{q}}_\mathrm{d} + \tilde{\mathbf{\Gamma}}_{\mathrm{d}} = \begin{bmatrix} 0&0&\tilde{L}_{\mathrm{d},13}\\ 0&\tilde{L}_{\mathrm{d},22}&\tilde{L}_{\mathrm{d},23}\\ \tilde{L}_{\mathrm{d},31}&\tilde{L}_{\mathrm{d},32}&\tilde{L}_{\mathrm{d},33}
		\end{bmatrix} \tilde{\mathbf{u}}\text{~,}
	\end{equation}
	where $\tilde{\mathbf{M}}_{\mathrm{d}}=\mathbf{Q}_\mathrm{d}\mathbf{M}_{\mathrm{d}}$ and $\tilde{\mathbf{\Gamma}}_{\mathrm{d}}=\mathbf{Q}_\mathrm{d}\mathbf{\Gamma}_{\mathrm{d}}$. The comparison with \eqref{eqODEdspThetaProjectQR} shows that only the virtual input $\tilde{u}_3$ acts on the first line and thereby directly on $\theta(t)$. The fully actuated DSP dynamics are expressed as
	\begin{equation}\label{eqStateSpacedspFull}
		\dot{\mathbf{x}}_\mathrm{d} = \underbrace{\begin{bmatrix}\dot{\mathbf{q}}_\mathrm{d}\\-\mathbf{M}_{\mathrm{d}}^{-1}\mathbf{\Gamma}_\mathrm{d}\end{bmatrix}}_{\mathbf{f}_{\mathrm{d}}(\mathbf{x}_\mathrm{d})}
		+ \underbrace{\begin{bmatrix}\mathbf{0}_{3\times 2}\\\mathbf{M}_{\mathrm{d}}^{-1}\tilde{\mathbf{B}}_\mathrm{d}\begin{bmatrix}\mathbf{I}_2\\\mathbf{0}_{1\times2}\end{bmatrix}\end{bmatrix}}_{\mathbf{g}_{\mathrm{d},12}(\mathbf{x}_\mathrm{d})}\begin{bmatrix}\tilde{u}_1\\\tilde{u}_2\end{bmatrix}
		+ \underbrace{\begin{bmatrix}\mathbf{0}_{3\times 1}\\\mathbf{M}_{\mathrm{d}}^{-1}\tilde{\mathbf{B}}_\mathrm{d}\begin{bmatrix}\mathbf{0}_{2\times1}\\1\end{bmatrix}\end{bmatrix}}_{\mathbf{g}_{\mathrm{d},3}(\mathbf{x}_\mathrm{d})}\tilde{u}_3
	\end{equation}
	in state space.

\subsubsection{Control Design}\label{labDSPfullyactControl}
    The control design consists of two steps. The first step is analogous to the DSP controller with underactuation, where only the virtual inputs $\tilde{u}_1$ and $\tilde{u}_2$ are used to zero the two outputs
	\begin{equation*}
		\mathbf{y}_\mathrm{d} = \mathbf{h}_\mathrm{d}(\mathbf{x}_\mathrm{d})=\mathbf{q}_\mathrm{di} - \mathbf{q}_\mathrm{r,di}(\theta,\boldsymbol{\alpha}_\mathrm{d})\text{~,}
	\end{equation*}
    given by \eqref{eqFeedbackunderactuatedDsp}. Differentiating this output twice gives
    \begin{subequations}\label{eqFeedbackDspDiffFull}
		\begin{align}
		\dot{\mathbf{y}}_\mathrm{d} &= \mathcal{L}_{\mathbf{f}_{\mathrm{d}}}\mathbf{h}_\mathrm{d}(\mathbf{x}_\mathrm{d})\text{~,}\label{eqFeedbackDspDiffFullDot}\\
		\ddot{\mathbf{y}}_\mathrm{d} &= \mathcal{L}^2_{\mathbf{f}_{\mathrm{d}}} \mathbf{h}_\mathrm{d}(\mathbf{x}_\mathrm{d}) + 
			\mathcal{L}_{\mathbf{g}_{\mathrm{d},12}} \mathcal{L}_{\mathbf{f}_{\mathrm{d}}} \mathbf{h}_\mathrm{d}(\mathbf{x}_\mathrm{d})\!\begin{bmatrix}\tilde{u}_1\\\tilde{u}_2\end{bmatrix} + \mathcal{L}_{\mathbf{g}_{\mathrm{d},3}} \mathcal{L}_{\mathbf{f}_{\mathrm{d}}} \mathbf{h}_\mathrm{d}(\mathbf{x}_\mathrm{d})\,\tilde{u}_3\text{~,}\label{eqFeedbackDspDiffFullDdot}
		\end{align}
	\end{subequations}
	and the diffeomorphism $\mathbf{\Phi}_\mathrm{d}(\mathbf{q}_\mathrm{d})=[\boldsymbol{\eta}_{\mathrm{d},1}^\top,\xi_{\mathrm{d},1}]^\top=[\mathbf{h}_\mathrm{d}^\top,\theta(\mathbf{q}_\mathrm{d})]^\top$ gives the Byrnes–Isidori form
	\begin{equation}\label{eqBInormalformDspFull}
		\begin{bmatrix}
			\dot{\boldsymbol{\eta}}_{\mathrm{d},1} \\ \dot{\boldsymbol{\eta}}_{\mathrm{d},2} \\ 
			\dot{\xi}_{\mathrm{d},1} \\ \dot{\xi}_{\mathrm{d},2}
		\end{bmatrix} = \begin{bmatrix} \boldsymbol{\eta}_{\mathrm{d},2} \\ \mathcal{L}^2_{\mathbf{f}_{\mathrm{d}}} \mathbf{h}_\mathrm{d} \\ \mathcal{L}_{\mathbf{f}_{\mathrm{d}}} \theta \\ 
			\mathcal{L}_{\mathbf{f}_{\mathrm{d}}} \tilde{\sigma}
		\end{bmatrix}+\begin{bmatrix} \mathbf{0} \\ \mathcal{L}_{\mathbf{g}_{\mathrm{d},12}} \mathcal{L}_{\mathbf{f}_{\mathrm{d}}} \mathbf{h}_\mathrm{d} \\ 0 \\ 0
		\end{bmatrix}\!\!\begin{bmatrix}\tilde{u}_1\\\tilde{u}_2\end{bmatrix}+\begin{bmatrix} \mathbf{0} \\ \mathcal{L}_{\mathbf{g}_{\mathrm{d},3}} \mathcal{L}_{\mathbf{f}_{\mathrm{d}}} \mathbf{h}_\mathrm{d} \\ 0 \\ \mathcal{L}_{\mathbf{g}_{\mathrm{d},3}} \tilde{\sigma}
		\end{bmatrix}\!\tilde{u}_3
	\end{equation}
	with $\boldsymbol{\eta}_{\mathrm{d},2}=\mathcal{L}_{\mathbf{f}_{\mathrm{d}}} \mathbf{h}_\mathrm{d}$, $\xi_{\mathrm{d},2}=\tilde{\sigma}_\mathrm{d}(\mathbf{q}_\mathrm{d},\dot{\mathbf{q}}_\mathrm{d})$ and
	\begin{equation}\label{eqMomentumDspFull}
		\tilde{\sigma}_\mathrm{d}=\underbrace{\begin{bmatrix} 1&\mathbf{0}_{1\times 2} \end{bmatrix} \tilde{\mathbf{M}}_\mathrm{d}(\mathbf{q}_\mathrm{d})}_{\tilde{\boldsymbol{\gamma}}_{\mathrm{d},0}(\mathbf{q}_\mathrm{d})} \dot{\mathbf{q}}_\mathrm{d}\text{~.}
	\end{equation}
	Feedback linearization is accomplished by introducing the new control input $\mathbf{v}_{\mathrm{d}}=\ddot{\mathbf{y}}_{\mathrm{d}}$ which is related to the output by a simple double integrator and to the virtual inputs $\tilde{\mathbf{u}}$ via \eqref{eqFeedbackDspDiffFullDdot}:
	\begin{equation}\label{eqBMotordspFull}
		\begin{bmatrix}\tilde{u}_1\\\tilde{u}_2\end{bmatrix} = \left(\mathcal{L}_{\mathbf{g}_{\mathrm{d},12}} \mathcal{L}_{\mathbf{f}_{\mathrm{d}}} \mathbf{h}_\mathrm{d}\right)^{-1} \left(\mathbf{v}_\mathrm{d}-\mathcal{L}^2_{\mathbf{f}_{\mathrm{d}}} \mathbf{h}_\mathrm{d}-\mathcal{L}_{\mathbf{g}_{\mathrm{d},3}} \mathcal{L}_{\mathbf{f}_{\mathrm{d}}} \mathbf{h}_\mathrm{d}\,\tilde{u}_3\right) \text{~.}
	\end{equation}
	A PD controller with $\mathbf{v}_\mathrm{d} = -\mathbf{K}_\mathrm{P,d\,}\mathbf{y}-\mathbf{K}_\mathrm{D,d\,}\dot{\mathbf{y}}$ and positive definite gains $\mathbf{K}_\mathrm{P,d}$ and $\mathbf{K}_\mathrm{D,d}$ is introduced to achieve asymptotic stability for the linearized system. Since $\mathcal{L}_{\mathbf{g}_{\mathrm{d},3}}\tilde{\sigma}_\mathrm{d}\neq0$, the dynamics that correspond to $\xi_{\mathrm{d},1}=\theta$ and $\xi_{\mathrm{d},2}=\tilde{\sigma}_\mathrm{d}$ are controllable through the input $\tilde{u}_3$, in contrast to the DSP controller with underactuation. The reference dynamics of the controlled system in the DSP follow from \eqref{eqBInormalformDspFull} and \eqref{eqBMotordspFull} for $\mathbf{y}_{\mathrm{d}} \equiv \mathbf{0}$. In this case, $\mathbf{v}_{\mathrm{d}}^\ast=\mathbf{0}$,
	\begin{equation}\label{eqBMotordspZdFull}
		\begin{bmatrix}\tilde{u}_1^*\\\tilde{u}_2^*\end{bmatrix} = -\left(\mathcal{L}_{\mathbf{g}_{\mathrm{d},12}} \mathcal{L}_{\mathbf{f}_{\mathrm{d}}} \mathbf{h}_\mathrm{d}\right)^{-1} \left(\mathcal{L}^2_{\mathbf{f}_{\mathrm{d}}} \mathbf{h}_\mathrm{d}+\mathcal{L}_{\mathbf{g}_{\mathrm{d},3}} \mathcal{L}_{\mathbf{f}_{\mathrm{d}}} \mathbf{h}_\mathrm{d}\,\tilde{u}_3\right)
	\end{equation}
	and the remaining dynamics from \eqref{eqBInormalformDspFull} are restricted to the smooth two dimensional submanifold $\mathcal{W}_\mathrm{d}=\{\mathbf{x}_\mathrm{d}\in T\mathcal{Q}_{\mathrm{d}}|\,\mathbf{h}_\mathrm{d}(\mathbf{x}_\mathrm{d})=\mathbf{0}, \mathcal{L}_{\mathbf{f}_{\mathrm{d}}}\mathbf{h}_\mathrm{d}(\mathbf{x}_\mathrm{d})=\mathbf{0}\}$. Since $\boldsymbol{\eta}_{\mathrm{d},1}=\mathbf{0}$ and $\boldsymbol{\eta}_{\mathrm{d},2}=\mathbf{0}$, the reference dynamics $\dot{\mathbf{w}}_\mathrm{d}=\mathbf{f}_{\mathrm{d,ref}}(\mathbf{w}_\mathrm{d})=[\mathcal{L}_{\mathbf{f}_{\mathrm{d}}}\theta\rvert_{\mathcal{W}_\mathrm{d}},\mathcal{L}_{\mathbf{f}_{\mathrm{d}}}\tilde{\sigma}_\mathrm{d}\rvert_{\mathcal{W}_\mathrm{d}}+\mathcal{L}_{\mathbf{g}_{\mathrm{d},3}}\tilde{\sigma}_\mathrm{d}\rvert_{\mathcal{W}_\mathrm{d}}\tilde{u}_3]^\top$ with $\mathbf{w}_\mathrm{d}=[\xi_{\mathrm{d},1},\xi_{\mathrm{d},2}]$ in the DSP are given by the restriction of the remaining dynamics to the manifold $\mathcal{W}_\mathrm{d}$. Analogous to \eqref{eqZerodynamicsCoordinatesDsp}, this restriction gives
	\begin{equation}\label{eqZerodynamicsCoordinatesDspFull}
		\begin{aligned}
		\left.\mathbf{q}_\mathrm{d}\right\rvert_{\mathcal{W}_\mathrm{d}} &= \begin{bmatrix} \xi_{\mathrm{d},1}\\ \mathbf{q}_\mathrm{r,di}(\xi_{\mathrm{d},1},\boldsymbol{\alpha}_\mathrm{d}) \end{bmatrix} \text{~,} \\
		\left.\tilde{\sigma}_\mathrm{d}\right\rvert_{\mathcal{W}_\mathrm{d}} &= \underbrace{\tilde{\boldsymbol{\gamma}}_{\mathrm{d},0}(\left.\mathbf{q}_\mathrm{d}\right\rvert_{\mathcal{W}_\mathrm{d}}) \begin{bmatrix} 1\\ \mathbf{q}_\mathrm{r,di}^\prime(\xi_{\mathrm{d},1},\boldsymbol{\alpha}_\mathrm{d}) \end{bmatrix}}_{\kappa_{\mathrm{d},1}^{-1}(\xi_{\mathrm{d},1})}\dot{\xi}_{\mathrm{d},1}\text{~,}\\
		\left.\dot{\mathbf{q}}_\mathrm{d}\right\rvert_{\mathcal{W}_\mathrm{d}} &= \kappa_{\mathrm{d},1}(\xi_{\mathrm{d},1})\begin{bmatrix} 1\\ \mathbf{q}_\mathrm{r,di}^\prime(\xi_{\mathrm{d},1},\boldsymbol{\alpha}_\mathrm{d}) \end{bmatrix}\xi_{\mathrm{d},2}\text{~,}
		\end{aligned}
	\end{equation}
	which in combination with \eqref{eqODEdspThetaProjectQRFull} yields the remaining dynamics
 	\begin{subequations}\label{eqZerodynamicsDspFull}
		\begin{align}
		\dot{\xi}_{\mathrm{d},1} &= \kappa_{\mathrm{d},1}(\xi_{\mathrm{d},1})\, \xi_{\mathrm{d},2} \text{~,} \label{eqZerodynamicsFullDspDerive1}\\
		\dot{\xi}_{\mathrm{d},2} &= \left.\left(\dot{\mathbf{q}}_\mathrm{d}^\top\frac{\partial\tilde{\boldsymbol{\gamma}}_{\mathrm{d},0}^\top}{\partial\mathbf{q}_\mathrm{d}}\dot{\mathbf{q}}_\mathrm{d} - \begin{bmatrix} 1&\mathbf{0}_{1\times 2} \end{bmatrix} \tilde{\boldsymbol{\Gamma}}_\mathrm{d}(\mathbf{q}_\mathrm{d},\dot{\mathbf{q}}_\mathrm{d})+\tilde{L}_{\mathrm{d},13}(\mathbf{q}_\mathrm{d})\,\tilde{u}_3\right)\right\rvert_{\mathcal{W}_\mathrm{d}}\notag\\
			&= \kappa_{\mathrm{d},2}(\xi_{\mathrm{d},1})+\kappa_{\mathrm{d},3}(\xi_{\mathrm{d},1})\,\xi_{\mathrm{d},2}^2+\kappa_{\mathrm{d},4}(\xi_{\mathrm{d},1})\,\tilde{u}_3 \text{~.}\label{eqZerodynamicsFullDspDerive2}
		\end{align}
	\end{subequations}
	We now define another output for the reference dynamics \eqref{eqZerodynamicsDspFull} to be zeroed by use of $\tilde{u}_3$. Since the aim is to influence the stability of the limit cycle solution of the hybrid dynamic system with control, we define a parametric function $\zeta_{\mathrm{r,d}}(\theta,\boldsymbol{\alpha}_\zeta)$ as a reference trajectory for $\zeta_\mathrm{d}=\frac{1}{2}\xi_{\mathrm{d},2}^2$ and the output
	\begin{equation}\label{eqControlDspZeta}
		y_{\zeta}=h_{\zeta}=\zeta_\mathrm{d}-\zeta_\mathrm{r,d}(\xi_{\mathrm{d},1},\boldsymbol{\alpha}_\zeta) \text{~.}
	\end{equation}
	The first derivative
	\begin{equation}
	    y_{\zeta}^\prime=h_{\zeta}^\prime=\frac{\mathrm{d}\zeta_\mathrm{d}}{\mathrm{d}\xi_{\mathrm{d},1}}-\zeta_\mathrm{r,d}^\prime(\xi_{\mathrm{d},1},\boldsymbol{\alpha}_\zeta)
	\end{equation}
	of this output with respect to $\xi_{\mathrm{d},1}$ depends on the input $\tilde{u}_3$ since
	\begin{equation}\label{eqSolDspZetsFull}
		\frac{\mathrm{d}\zeta_\mathrm{d}}{\mathrm{d}\xi_{\mathrm{d},1}} = \frac{\kappa_{\mathrm{d},2}(\xi_{\mathrm{d},1})}{\kappa_{\mathrm{d},1}(\xi_{\mathrm{d},1})}+\frac{2\kappa_{\mathrm{d},3}(\xi_{\mathrm{d},1})}{\kappa_{\mathrm{d},1}(\xi_{\mathrm{d},1})}\zeta_\mathrm{d}+\frac{\kappa_{\mathrm{d},4}(\xi_{\mathrm{d},1})}{\kappa_{\mathrm{d},1}(\xi_{\mathrm{d},1})}\tilde{u}_3 \text{~.}
	\end{equation}
	The reference trajectory for the remaining dynamics is stabilized by a P~controller with $y_{\zeta}^\prime=-K_\zeta\,y_{\zeta}$, $K_\zeta>0$, which gives the feedback
	\begin{equation}
	    \tilde{u}_3 = -\left(\frac{\kappa_{\mathrm{d},2}(\xi_{\mathrm{d},1})}{\kappa_{\mathrm{d},4}(\xi_{\mathrm{d},1})} + \frac{2\kappa_{\mathrm{d},3}(\xi_{\mathrm{d},1})}{\kappa_{\mathrm{d},4}(\xi_{\mathrm{d},1})}\zeta_\mathrm{d} + \frac{\kappa_{\mathrm{d},1}(\xi_{\mathrm{d},1})}{\kappa_{\mathrm{d},4}(\xi_{\mathrm{d},1})}\left(K_\zeta\left(\zeta_\mathrm{d}-\zeta_\mathrm{r,d}\right)-\zeta_\mathrm{r,d}^\prime\right)\right)
	\end{equation}
	and the dynamics
	\begin{equation}\label{eqControlledDynamcisZetaFull}
	    \frac{\mathrm{d}\zeta_\mathrm{d}}{\mathrm{d}\xi_{\mathrm{d},1}} = \zeta_\mathrm{r,d}^\prime - K_\zeta \left(\zeta_\mathrm{d} - \zeta_\mathrm{r,d}\right) \text{~.}
	\end{equation}
	
\subsubsection{Limit Cycle Solution and Stability}\label{labDSPoveractHZDPoincare}
    As for the DSP controller with underactuation, there is a limit cycle solution for the infinite sequence \ldots$\rightarrow$DSP$\rightarrow$LO$\rightarrow$SSP$\rightarrow$TD$\rightarrow$\ldots for periodic walking with with zero control error. Again, $\theta(t)$ increases monotonically with time during both DSP and SSP. We presuppose the same conditions via \eqref{eqInvariantMappings} and \eqref{eqInvariantZDStateMappings} for the reference trajectories $\mathbf{q}_\mathrm{r,s}$ and $\mathbf{q}_\mathrm{r,di}$ and the states of the zero/remaining dynamics at the transition events, where the consequences of \eqref{eqInvariantZDStateMappings3} and \eqref{eqInvariantZDStateMappings4} for $\zeta_\mathrm{r,d}$ are addressed below. The solution for the SSP is the same as in the previous section for the DSP controller with underactuation. The solution for the DSP follows from \eqref{eqControlledDynamcisZetaFull} for the initial condition $\zeta_\mathrm{d}^+$:
	\begin{align}
		\zeta_\mathrm{d}(\xi_{\mathrm{d},1}) &= \zeta_\mathrm{r,d} + \left(\zeta_\mathrm{d}^{+}-\zeta_\mathrm{r,d}^{+}\right)\nu_\mathrm{d}(\xi_{\mathrm{d},1}) \text{~,}\label{eqZerodynamicsOverDspControlledSol}\\
		\nu_\mathrm{d}(\xi_{\mathrm{d},1})&=\exp\left(-K_{\zeta}(\xi_{\mathrm{d},1} - \xi_{\mathrm{d},1}^{+})\right)\text{~.}\label{eqZerodynamicsDspNuFull}
	\end{align}
	Any deviation of $\zeta_\mathrm{d}$ from $\zeta_\mathrm{r,d}$ at the beginning of the DSP decreases exponentially with $\xi_{\mathrm{d},1}$ thanks to the P controller. To ensure that there is not control error $y_\zeta$ for the limit cycle solution, we require an additional condition for the reference trajectory $\zeta_\mathrm{r,d}$ by considering \eqref{eqZetaMappings} (which follow from \eqref{eqInvariantZDStateMappings3} and \eqref{eqInvariantZDStateMappings4}) and the SSP solution \eqref{eqSolSspZetsSol}. Evaluating these equations for the solution on the reference trajectory with $\zeta_{\mathrm{d},1}(\xi_{\mathrm{d},1}^+)=\zeta_\mathrm{r,d}^+$ and $\zeta_{\mathrm{d},1}(\xi_{\mathrm{d},1}^-)=\zeta_\mathrm{r,d}^-$ gives the desired relation
	\begin{equation}\label{eqZetaCondition}
     	\zeta_\mathrm{r,d}^+ = \left(\delta_\mathrm{s}^{\mathrm{d}}\delta_\mathrm{d}^{\mathrm{s}}\right)^2 \zeta_\mathrm{r,d}^- + \left(\delta_\mathrm{s}^{\mathrm{d}}\right)^2\mu_\mathrm{s}(\xi_{\mathrm{s},1}^{-}) 
	\end{equation}
	for $\zeta_\mathrm{r,d}(\theta,\boldsymbol{\alpha}_\zeta)$. This gives the Poincaré map $P:\mathcal{W}_\mathrm{d}\rightarrow\mathcal{W}_\mathrm{d}, \zeta_{\mathrm{d}}^+\mapsto\zeta_{\mathrm{d}}^+$ of the initial condition $\zeta_{\mathrm{d}}^+$ of one period onto the beginning of the next DSP as
	\begin{equation}\label{eqIniCondDspFull}
	    P\!\left(\zeta_{\mathrm{d}}^{+}\right) = (\delta_\mathrm{s}^{\mathrm{d}})^2\left((\delta_\mathrm{d}^{\mathrm{s}})^2 \left(\zeta_\mathrm{r,d}^-+(\zeta_\mathrm{d}^+-\zeta_\mathrm{r,d}^+)\nu_\mathrm{d}(\xi_{\mathrm{d},1}^-)\right) + \mu_\mathrm{s}(\xi_{\mathrm{s},1}^-)\right)
	\end{equation}
	with the fixed point $\zeta_\mathrm{d}^+ =\zeta_\mathrm{r,d}^+$ and the Floquet multiplier
	\begin{equation}\label{eqFloquetFull}
		\Lambda = \frac{\partial P(\zeta_{\mathrm{d}}^{+})}{\partial\zeta_{\mathrm{d}}^{+}} = (\delta_\mathrm{s}^{\mathrm{d}}\delta_\mathrm{d}^{\mathrm{s}})^2 \,\nu_\mathrm{d}(\xi_{\mathrm{d},1}^{-}) \text{~.}
	\end{equation}
	Since $\nu_\mathrm{d}(\xi_{\mathrm{d},1}^-)=\exp(-K_{\zeta}(\xi_{\mathrm{d},1}^- - \xi_{\mathrm{d},1}^{+}))$ via \eqref{eqZerodynamicsDspNuFull}, increasing the gain $K_{\zeta}$ decreases the multiplier $\Lambda$. Therefore, the proposed controller allows us to directly influence the stability of the limit cylce, as desired.
	
\subsection{DSP Controller with Overactuation}\label{labSubsectionDSPoveract}
    In the previous section, a controller that produces (arbitrarily) stable limit cycles for bipedal walking with a fully actuated DSP is introduced. Since the DSP system has three DoF but four actuators, this design can be extended further to add yet another objective. In our experience, the system is very sensitive to perturbations that influence the contact forces at both feet in the DSP. Since this may lead to slipping in tangential direction or even lift-off of one leg, small perturbations in the DSP can result in stumbling of the walking robot and catastrophic failures such as falling over. Since one input is not sufficient to independently control the forces at both foot contacts, we propose to instead require them to be parallel. While this does not prevent lift-off events due to perturbations, any loss of static friction will lead to slipping at both feet simultaneously. In this case, the distance between both legs stays constant and any such perturbation results in a sliding/skating motion until both feet stick to the ground again. 
    
    According to \eqref{eqForce2}, the reaction force $\mathbf{F}_2 := \mathbf{f}_{\mathbf{F}}(\mathbf{u}_{\mathrm{d}})$ solely depends on the actuators $\mathbf{u}_{\mathrm{d}}$ of the trailing leg. Utilizing this fact simplifies the formulation of the control task. Thus instead of introducing new virtual inputs, the physical actuators $\mathbf{u}$ are used in the control design of the overactuated model. Since the first line of the input matrix $\mathbf{B}_\mathrm{di} = [\mathbf{0}_{2\times 1}, \mathbf{I}_2]^\top$ in equation \eqref{eqODEdspTheta} is already zero (the underactuated DoF is not controllable by the input $\mathbf{u}_{\mathrm{i}}$), there is also no need to employ a matrix factorization to modify the equation of motion. The state space expression \eqref{eqStateSpacedsp} is used for the following control design. 

\subsubsection{Control Design}\label{labDSPoveractControl}
    The essential task \eqref{eqFeedbackunderactuatedDsp} remains unchanged for the independent joint coordinates $\mathbf{q}_\mathrm{di}$:
    \begin{equation*}
		\mathbf{y}_\mathrm{d} = \mathbf{h}_\mathrm{d}(\mathbf{x}_\mathrm{d})=\mathbf{q}_\mathrm{di} - \mathbf{q}_\mathrm{r,di}(\theta,\boldsymbol{\alpha}_\mathrm{d})\text{~,}
	\end{equation*}
    Again, differentiating $\mathbf{y}_\mathrm{d}$ twice yields
    \begin{subequations}\label{eqFeedbackDspDiffOver}
		\begin{align}
		\dot{\mathbf{y}}_\mathrm{d} &= \mathcal{L}_{\mathbf{f}_{\mathrm{d}}}\mathbf{h}_\mathrm{d}(\mathbf{x}_\mathrm{d})\text{~,}\label{eqFeedbackDspDiffOverDot}\\
		\ddot{\mathbf{y}}_\mathrm{d} &= \mathcal{L}^2_{\mathbf{f}_{\mathrm{d}}} \mathbf{h}_\mathrm{d}(\mathbf{x}_\mathrm{d}) + 
			\mathcal{L}_{\mathbf{g}_{\mathrm{i}}} \mathcal{L}_{\mathbf{f}_{\mathrm{d}}} \mathbf{h}_\mathrm{d}(\mathbf{x}_\mathrm{d})\mathbf{u}_{\mathrm{i}} + \mathcal{L}_{\mathbf{g}_{\mathrm{d}}} \mathcal{L}_{\mathbf{f}_{\mathrm{d}}} \mathbf{h}_\mathrm{d}(\mathbf{x}_\mathrm{d})\mathbf{u}_{\mathrm{d}}\text{~.}\label{eqFeedbackDspDiffOverDdot}
		\end{align}
	\end{subequations}
	Recalling the diffeomorphism $\mathbf{\Phi}_\mathrm{d}(\mathbf{q}_\mathrm{d})=[\boldsymbol{\eta}_{\mathrm{d},1}^\top,\xi_{\mathrm{d},1}]^\top=[\mathbf{h}_\mathrm{d}^\top,\theta(\mathbf{q}_\mathrm{d})]^\top$, the Byrnes–Isidori form is expressed as 
	\begin{equation}\label{eqBInormalformDspOver}
		\begin{bmatrix}
			\dot{\boldsymbol{\eta}}_{\mathrm{d},1} \\ \dot{\boldsymbol{\eta}}_{\mathrm{d},2} \\ 
			\dot{\xi}_{\mathrm{d},1} \\ \dot{\xi}_{\mathrm{d},2}
		\end{bmatrix} = \begin{bmatrix} \boldsymbol{\eta}_{\mathrm{d},2} \\ \mathcal{L}^2_{\mathbf{f}_{\mathrm{d}}} \mathbf{h}_\mathrm{d} \\ \mathcal{L}_{\mathbf{f}_{\mathrm{d}}} \theta \\ 
			\mathcal{L}_{\mathbf{f}_{\mathrm{d}}} \sigma_\mathrm{d}
		\end{bmatrix}+\begin{bmatrix} \mathbf{0} \\ \mathcal{L}_{\mathbf{g}_{\mathrm{i}}} \mathcal{L}_{\mathbf{f}_{\mathrm{d}}} \mathbf{h}_\mathrm{d} \\ 0 \\ 0
		\end{bmatrix}\!\mathbf{u}_{\mathrm{i}} + \begin{bmatrix} \mathbf{0} \\ \mathcal{L}_{\mathbf{g}_{\mathrm{d}}} \mathcal{L}_{\mathbf{f}_{\mathrm{d}}} \mathbf{h}_\mathrm{d} \\ 0 \\ \mathcal{L}_{\mathbf{g}_{\mathrm{d}}} \sigma_\mathrm{d}
		\end{bmatrix}\!\mathbf{u}_{\mathrm{d}} \text{~,}
	\end{equation}
    with the generalized momentum $\xi_{\mathrm{d},2} = \sigma_\mathrm{d}$, i.\,e.\
	\begin{equation}\label{eqMomentumDspOver}
		\sigma_\mathrm{d} = \underbrace{\begin{bmatrix} 1&\mathbf{0}_{1\times 2} \end{bmatrix} \mathbf{M}_\mathrm{d}(\mathbf{q}_\mathrm{d})}_{\boldsymbol{\gamma}_{\mathrm{d},0}(\mathbf{q}_\mathrm{d})} \dot{\mathbf{q}}_\mathrm{d}\text{~,}
	\end{equation}
    and $\mathcal{L}_{\mathbf{g}_{\mathrm{d}}} \sigma_\mathrm{d} := [1\quad \mathbf{0}_{1\times 2}] \left(\mathbf{J}_{\mathbf{\Omega}}^\top\right)$. After substituting the new control input $\mathbf{v}_\mathrm{d} = \ddot{\mathbf{y}}_\mathrm{d}$ into \eqref{eqFeedbackDspDiffOverDdot}, the actuation $\mathbf{u}_{\mathrm{i}}$ is determined 
    \begin{equation}\label{eqBMotordspOver}
		\mathbf{u}_{\mathrm{i}} = \left(\mathcal{L}_{\mathbf{g}_{\mathrm{i}}} \mathcal{L}_{\mathbf{f}_{\mathrm{d}}} \mathbf{h}_\mathrm{d}\right)^{-1} \left(\mathbf{v}_\mathrm{d}-\mathcal{L}^2_{\mathbf{f}_{\mathrm{d}}} \mathbf{h}_\mathrm{d}-\mathcal{L}_{\mathbf{g}_{\mathrm{d}}} \mathcal{L}_{\mathbf{f}_{\mathrm{d}}} \mathbf{h}_\mathrm{d}\mathbf{u}_{\mathrm{d}}\right) \text{~.}
	\end{equation}
    Also assuming the vanishing control error $\mathbf{v}_\mathrm{d}=\mathbf{0}$, the input is then expressed as
    \begin{equation}\label{eqBMotordspOverZeroErr}
		\mathbf{u}_{\mathrm{i}}^* = -\left(\mathcal{L}_{\mathbf{g}_{\mathrm{i}}} \mathcal{L}_{\mathbf{f}_{\mathrm{d}}} \mathbf{h}_\mathrm{d}\right)^{-1} \left(\mathcal{L}^2_{\mathbf{f}_{\mathrm{d}}} \mathbf{h}_\mathrm{d} + \mathcal{L}_{\mathbf{g}_{\mathrm{d}}} \mathcal{L}_{\mathbf{f}_{\mathrm{d}}} \mathbf{h}_\mathrm{d}\mathbf{u}_{\mathrm{d}}\right) \text{~.}
	\end{equation}
    In this manner, the full system's dynamics \eqref{eqBInormalformDspOver} is described by the reference dynamics $\dot{\mathbf{s}}_\mathrm{d}=\mathbf{f}_{\mathrm{d,ref}}(\mathbf{s}_\mathrm{d})=[\mathcal{L}_{\mathbf{f}_{\mathrm{d}}} \theta\rvert_{\mathcal{S}_\mathrm{d}},\mathcal{L}_{\mathbf{f}_{\mathrm{d}}} \sigma_\mathrm{d}\rvert_{\mathcal{S}_\mathrm{d}} + \mathcal{L}_{\mathbf{g}_{\mathrm{d}}} \sigma_\mathrm{d}\rvert_{\mathcal{S}_\mathrm{d}}\mathbf{u}_{\mathrm{d}}]^\top$ with $\mathbf{s}_\mathrm{d}=[\xi_{\mathrm{d},1},\xi_{\mathrm{d},2}]$ on the submanifold $\mathcal{S}_\mathrm{d}=\{\mathbf{x}_\mathrm{d}\in T\mathcal{Q}_{\mathrm{d}}|\,\mathbf{h}_\mathrm{d}(\mathbf{x}_\mathrm{d})=\mathbf{0}, \mathcal{L}_{\mathbf{f}_{\mathrm{d}}}\mathbf{h}_\mathrm{d}(\mathbf{x}_\mathrm{d})=\mathbf{0}\}$. The same derivation as for \eqref{eqZerodynamicsCoordinatesDspFull} is used to derive the remaining dynamics in the form 
	\begin{subequations}\label{eqZerodynamicsDspOver}
		\begin{align}
		\dot{\xi}_{\mathrm{d},1} &= \kappa_{\mathrm{d},1}(\xi_{\mathrm{d},1})\, \xi_{\mathrm{d},2} \text{~,} \label{eqZerodynamicsOverDspDerive1}\\
		\dot{\xi}_{\mathrm{d},2} &= \left(\dot{\mathbf{q}}_\mathrm{d}^\top\frac{\partial\boldsymbol{\gamma}_{\mathrm{d},0}^\top}{\partial\mathbf{q}_\mathrm{d}}\dot{\mathbf{q}}_\mathrm{d} - \begin{bmatrix} 1&\mathbf{0}_{1\times 2} \end{bmatrix} \boldsymbol{\Gamma}_\mathrm{d}(\mathbf{q}_\mathrm{d},\dot{\mathbf{q}}_\mathrm{d})\right.\notag\\[-1em]
		&\hspace*{4em}\left.\left.+\begin{bmatrix} 1&\mathbf{0}_{1\times 2} \end{bmatrix}\mathbf{J}_{\mathbf{\Omega}}^\top \begin{bmatrix} 1 \\ 0 \end{bmatrix} u_{\mathrm{H}2}  +\begin{bmatrix} 1&\mathbf{0}_{1\times 2} \end{bmatrix}\mathbf{J}_{\mathbf{\Omega}}^\top \begin{bmatrix} 0 \\ 1 \end{bmatrix} u_{\mathrm{K}2} \right)\right\rvert_{\mathcal{S}_\mathrm{d}}\notag\\
		&= \kappa_{\mathrm{d},2}(\xi_{\mathrm{d},1})+\kappa_{\mathrm{d},3}(\xi_{\mathrm{d},1})\,\xi_{\mathrm{d},2}^2 + 
                \kappa_{\mathrm{d},4}(\xi_{\mathrm{d},1})\,u_{\mathrm{H}2} +\kappa_{\mathrm{d},5}(\xi_{\mathrm{d},1})\,u_{\mathrm{K}2} \text{~.}\label{eqZerodynamicsDspOver2}
		\end{align}
	\end{subequations}
    Obviously the dynamics for $\dot{\xi}_{\mathrm{d},2}$ is controllable through the inputs $\mathbf{u}_{\mathrm{d}} = [u_{\mathrm{H}2},u_{\mathrm{K}2}]^\top$. Instead of solving \eqref{eqZerodynamicsDspOver2}, the inputs $\mathbf{u}_{\mathrm{d}}$ are first determined from two additional control objectives: The first is to enforce the collinearity condition of the contact forces evaluated by the cross product of \eqref{eqForceSum} and \eqref{eqForce2}, i.\,e.\
	\begin{equation}\label{eqcollinear}
		\left(\left(\mathbf{F}_1 + \mathbf{F}_2\right) \times \mathbf{F}_2 \right)\cdot \vec{e}_y = 0 \text{~,}
	\end{equation}
    which delivers the relation between two actuators $\mathbf{u}_{\mathrm{d}}$ in the form 
	\begin{equation}\label{eqcollinearEval}
		u_{\mathrm{H}2} = C_\mathrm{O}(\xi_{\mathrm{d},1}) + C_\mathrm{K}(\xi_{\mathrm{d},1}) u_{\mathrm{K}2} \text{~,}
	\end{equation}
    where the expressions $C_\mathrm{O}(\xi_{\mathrm{d},1})$ and $C_\mathrm{K}(\xi_{\mathrm{d},1})$ do not depend on the actuation; The second control task is to influence the stability of the limit cycle by introducing the control output
	\begin{equation*}
		y_{\zeta}=h_{\zeta}=\zeta_\mathrm{d}-\zeta_\mathrm{r,d}(\xi_{\mathrm{d},1},\boldsymbol{\alpha}_\zeta) \text{~.}
	\end{equation*}
    Again a P controller ($y_{\zeta}^\prime=-K_\zeta\,y_{\zeta}$, $K_\zeta>0$) is used to stabilize $y_{\zeta}$ which yields the feedback
	\begin{multline}\label{eqFeedbackOveract}
	    \kappa_{\mathrm{d},4}(\xi_{\mathrm{d},1})\,u_{\mathrm{H}2} + \kappa_{\mathrm{d},5}(\xi_{\mathrm{d},1})\,u_{\mathrm{K}2} = 
        \underbrace{- \kappa_{\mathrm{d},2}(\xi_{\mathrm{d},1}) - 2\kappa_{\mathrm{d},3}(\xi_{\mathrm{d},1})\zeta_\mathrm{d} 
        - \kappa_{\mathrm{d},1}(\xi_{\mathrm{d},1})\left(K_\zeta\left(\zeta_\mathrm{d}-\zeta_\mathrm{r,d}\right)-\zeta_\mathrm{r,d}^\prime\right)}_{u_{\mathrm{d},\,\mathrm{rhs}}} \text{~,}
	\end{multline}
    with the entire right hand side of \eqref{eqFeedbackOveract} shorthanded as $u_{\mathrm{d},\,\mathrm{rhs}}$. Substituting \eqref{eqcollinearEval} into the equation determines the first input 
	\begin{equation}\label{eqFeedbackOveractUk2}
	    u_{\mathrm{K}2} = \frac{u_{\mathrm{d},\,\mathrm{rhs}} - \kappa_{\mathrm{d},4}(\xi_{\mathrm{d},1}) C_\mathrm{O}(\xi_{\mathrm{d},1})}{\kappa_{\mathrm{d},4}(\xi_{\mathrm{d},1}) C_\mathrm{K}(\xi_{\mathrm{d},1}) +  \kappa_{\mathrm{d},5}(\xi_{\mathrm{d},1})} \text{~,}
	\end{equation}
    which is further set into \eqref{eqcollinearEval} to determine $u_{\mathrm{H}2}$. Thus the former dynamics \eqref{eqZerodynamicsDspOver2} is replaced by the time free formulation
	\begin{equation*}
	    \frac{\mathrm{d}\zeta_\mathrm{d}}{\mathrm{d}\xi_{\mathrm{d},1}} = \zeta_\mathrm{r,d}^\prime - K_\zeta \left(\zeta_\mathrm{d} - \zeta_\mathrm{r,d}\right) \text{~.}
	\end{equation*}
     The limit cycle solution and the related stability analysis can be derived by the same procedure as in \ref{labDSPoveractHZDPoincare}, which is not repeated again here. Notice that one of the major benefits of introducing the collinearity condition is resulting a dependency between the two actuators in $\mathbf{u}_{\mathrm{d}}$, which are involved in controlling the generalized momentum $\sigma_\mathrm{d}$. Indeed, the presented objective of collinear contact forces is not the only possible extension. Other control objectives could instead be added by following a similar approach for control design for the DSP as presented above. For instance, we can introduce again the $4\times 4$ projection matrix $\mathbf{P}_\mathrm{o}$ to map the four virtual inputs $\tilde{\mathbf{u}}=[\tilde{u}_1,\tilde{u}_2,\tilde{u}_3,\tilde{u}_4]^\top$ on the physical actuators via 
	\begin{equation*}
	    \mathbf{u} = \mathbf{P}_\mathrm{o} \tilde{\mathbf{u}} \text{~.}
	\end{equation*}
     Instead of requiring the collinearity of the contact forces, the fourth independent virtual input could be used to influence the stability of the robot in the frontal plane, since the real robot system indeed moves in three dimensional space. However, due to the large effort of extending the planar model, this control task is not discussed in the present manuscript. 
     
\section{Numerical Optimization}\label{labSectionNumeric}
    Gait generation via numerical optimization is a common approach with many variations in the specific implementations \cite{NumericMethod1,NumericMethod2,NumericMethod3}. We evaluate the integral using numerical approximations in the manner of the introduced semi-analytical process, due to the one DoF hybrid zero dynamics formulation, and the short integration periods in the SSP and DSP. A constrained optimization problem, which optimizes the virtual constraints and minimizes the energy consumption of locomotion, is then formulated. In contrast to e.\,g.\ \cite{NumericMethod2}, the optimization constraints only consist of the necessary physical conditions for simulating feasible walking motions. Indeed, most of the constraints are inactive, which makes the evaluation very efficient. This section firstly defines the reference trajectory by use of Bézier polynomials in \ref{labSubsectionRef} and \ref{labSubsectionRefOverDsp}; then the optimization framework including its objective and constraints is presented in \ref{labSubsectionFramework}. 
    
\subsection{Reference Trajectories}\label{labSubsectionRef}
	The major task of the HZD based controller is to synchronize the joint revolution to its reference trajectory, which are regarded as virtual holonomic constraints. Specifically, these are $M$-order\footnote{This work assumes $M=6$. According to \cite{DissBauer}, a higher order dose not change the qualitative behaviour of the optimized energy efficiency.} Bézier polynomials $b(s,\boldsymbol{\alpha})$, defined by the normalized independent variable $s\in[0,1]$ and the parameter set $\boldsymbol{\alpha}$ of dimension $(M+1)$ which determines the shape of one Bézier polynomial:  
	\begin{equation}\label{eqBezierDef}
		b(s,\boldsymbol{\alpha}) := \sum_{k=0}^{M} \alpha_k \frac{M!}{k!(M-k)!}s^k(1-s)^{M-k} \text{~.}
	\end{equation}
	The dimension of the reference trajectory in the SSP and DSP is determined in accordance to their control output in \eqref{eqFeedbackssp} and \eqref{eqFeedbackunderactuatedDsp}, respectively. The parameter $\boldsymbol{\alpha}_\mathrm{d}:=[\boldsymbol{\alpha}_{\mathrm{d},0}, \boldsymbol{\alpha}_{\mathrm{d},1}, \dots, \boldsymbol{\alpha}_{\mathrm{d},M}]$ in the DSP is a $(2 \times (M+1))$ matrix which defines 2 independent Bézier polynomials; and $\boldsymbol{\alpha}_{\mathrm{s}}:=[\boldsymbol{\alpha}_{\mathrm{s},0}, \boldsymbol{\alpha}_{\mathrm{s},1}, \dots, \boldsymbol{\alpha}_{\mathrm{s},M}]$ of dimension $(4 \times (M+1))$ in the SSP defines 4 Bézier polynomials. 
 
    The initial step to define the reference trajectory in the DSP is to normalize the corresponding phase variable $\theta = \xi_{\mathrm{d},1}$ to $s \in [0,1]$ via
	\begin{equation}\label{eqBezierDefNormDsp}
		s(\theta) := \frac{\theta - \xi_{\mathrm{d},1}^{+}}{\xi_{\mathrm{d},1}^{-} - \xi_{\mathrm{d},1}^{+}} \text{~.}
	\end{equation}
	Thus the virtual constraints containing two independent joints in \eqref{eqFeedbackunderactuatedDsp} are expressed as
	\begin{equation}\label{eqBezierDefDsp}
		\mathbf{q}_\mathrm{r,di}(\theta,\boldsymbol{\alpha}_\mathrm{d}) := \begin{bmatrix} b_{1}(s(\theta), &[1&0]\boldsymbol{\alpha}_\mathrm{d}) \\ b_{2}(s(\theta), &[0&1]\boldsymbol{\alpha}_\mathrm{d}) \end{bmatrix} \text{~.}
	\end{equation}
	Moreover, the boundary value of a Bézier polynomial, including its derivatives with respect to $\theta$, can be expressed as
	\begin{equation}\label{eqBezierBVDsp}
		\begin{aligned}
        \mathbf{q}_\mathrm{r,di}(\xi_{\mathrm{d},1}^+,\boldsymbol{\alpha}_\mathrm{d}) &= \boldsymbol{\alpha}_{\mathrm{d},0} \text{~,} \\
        \mathbf{q}_\mathrm{r,di}(\xi_{\mathrm{d},1}^-,\boldsymbol{\alpha}_\mathrm{d}) &= \boldsymbol{\alpha}_{\mathrm{d},M} \text{~,} \\
        \mathbf{q}_\mathrm{r,di}^\prime(\xi_{\mathrm{d},1}^+,\boldsymbol{\alpha}_\mathrm{d}) &= \frac{M(\boldsymbol{\alpha}_{\mathrm{d},1} - \boldsymbol{\alpha}_{\mathrm{d},0})}{\xi_{\mathrm{d},1}^{-} - \xi_{\mathrm{d},1}^{+}} \text{~,} \\
        \mathbf{q}_\mathrm{r,di}^\prime(\xi_{\mathrm{d},1}^-,\boldsymbol{\alpha}_\mathrm{d}) &= \frac{M(\boldsymbol{\alpha}_{\mathrm{d},M} - \boldsymbol{\alpha}_{\mathrm{d},M-1})}{\xi_{\mathrm{d},1}^{-} - \xi_{\mathrm{d},1}^{+}} \text{~.}
		\end{aligned}
	\end{equation}
    Analogously, the boundary of the reference trajectory in the SSP is expressed as
	\begin{equation}\label{eqBezierBVSsp}
		\begin{aligned}
        \mathbf{q}_\mathrm{r,s}(\xi_{\mathrm{s},1}^+,\boldsymbol{\alpha}_\mathrm{s}) &= \boldsymbol{\alpha}_{\mathrm{s},0} \text{~,} \\
        \mathbf{q}_\mathrm{r,s}(\xi_{\mathrm{s},1}^-,\boldsymbol{\alpha}_\mathrm{s}) &= \boldsymbol{\alpha}_{\mathrm{s},M} \text{~,} \\
        \mathbf{q}_\mathrm{r,s}^\prime(\xi_{\mathrm{s},1}^+,\boldsymbol{\alpha}_\mathrm{s}) &= \frac{M(\boldsymbol{\alpha}_{\mathrm{s},1} - \boldsymbol{\alpha}_{\mathrm{s},0})}{\xi_{\mathrm{s},1}^{-} - \xi_{\mathrm{s},1}^{+}} \text{~,} \\
        \mathbf{q}_\mathrm{r,s}^\prime(\xi_{\mathrm{s},1}^-,\boldsymbol{\alpha}_\mathrm{s}) &= \frac{M(\boldsymbol{\alpha}_{\mathrm{s},M} - \boldsymbol{\alpha}_{\mathrm{s},M-1})}{\xi_{\mathrm{s},1}^{-} - \xi_{\mathrm{s},1}^{+}} \text{~.}
		\end{aligned}
	\end{equation}
    Substituting \eqref{eqBezierBVDsp} and \eqref{eqBezierBVSsp} into the mapping \eqref{eqInvariantMappings} yields
    \begin{subequations}\label{eqInvariantMappingsParam}
        \begin{align}
        \begin{bmatrix}\xi_{\mathrm{d},1}^+\\ \boldsymbol{\alpha}_{\mathrm{d},0}\end{bmatrix} &= \mathbf{H}_\mathrm{d}\mathbf{R}_\mathrm{s}^\mathrm{d}\mathbf{H}_\mathrm{s}^{-1}\begin{bmatrix}\xi_{\mathrm{s},1}^-\\ \boldsymbol{\alpha}_{\mathrm{s},M}\end{bmatrix}\text{~,}\label{eqInvariantMappingsParam1}\\
		\begin{bmatrix}\xi_{\mathrm{s},1}^+\\ \boldsymbol{\alpha}_{\mathrm{s},0}\end{bmatrix} &= \mathbf{H}_\mathrm{s}\mathbf{R}_\mathrm{d}^\mathrm{s}\begin{bmatrix}
			\mathbf{H}_\mathrm{d}^{-1}\begin{bmatrix}\xi_{\mathrm{d},1}^-\\ \boldsymbol{\alpha}_{\mathrm{d},M}\end{bmatrix}\\ \mathbf{\Omega}\left(\begin{bmatrix}\xi_{\mathrm{d},1}^-\\ \boldsymbol{\alpha}_{\mathrm{d},M}\end{bmatrix}\right)\end{bmatrix}\text{~,}\label{eqInvariantMappingsParam2}\\
		\begin{bmatrix} 1\\ \frac{M(\boldsymbol{\alpha}_{\mathrm{d},1} - \boldsymbol{\alpha}_{\mathrm{d},0})}{\xi_{\mathrm{d},1}^{-} - \xi_{\mathrm{d},1}^{+}} \end{bmatrix}\dot{\xi}_{\mathrm{d},1}^+ &= \underbrace{\mathbf{H}_\mathrm{d}\mathbf{\Delta}_{\mathrm{s}}\mathbf{H}_\mathrm{s}^{-1}\begin{bmatrix} 1\\ \frac{M(\boldsymbol{\alpha}_{\mathrm{s},M} - \boldsymbol{\alpha}_{\mathrm{s},M-1})}{\xi_{\mathrm{s},1}^{-} - \xi_{\mathrm{s},1}^{+}} \end{bmatrix}}_{\mathbf{\Delta}_{\mathrm{ref}}} \dot{\xi}_{\mathrm{s},1}^-\text{~,}\label{eqInvariantMappingsParam3}\\
		\begin{bmatrix} 1\\ \frac{M(\boldsymbol{\alpha}_{\mathrm{s},1} - \boldsymbol{\alpha}_{\mathrm{s},0})}{\xi_{\mathrm{s},1}^{-} - \xi_{\mathrm{s},1}^{+}} \end{bmatrix}\dot{\xi}_{\mathrm{s},1}^+ &= \mathbf{H}_\mathrm{s}\mathbf{R}_\mathrm{d}^\mathrm{s}\begin{bmatrix}\mathbf{H}_\mathrm{d}^{-1}\\ \mathbf{J}_{\mathbf{\Omega}}\left(\begin{bmatrix}\xi_{\mathrm{d},1}^-\\ \boldsymbol{\alpha}_{\mathrm{d},M}\end{bmatrix}\right)\end{bmatrix}\!\!\begin{bmatrix} 1\\ \frac{M(\boldsymbol{\alpha}_{\mathrm{d},M} - \boldsymbol{\alpha}_{\mathrm{d},M-1})}{\xi_{\mathrm{d},1}^{-} - \xi_{\mathrm{d},1}^{+}} \end{bmatrix}\dot{\xi}_{\mathrm{d},1}^-\text{~.}\label{eqInvariantMappingsParam4}
        \end{align}
    \end{subequations}
    Due to the invariant zero dynamics manifolds of the SSP and DSP with respect to the discontinuous transitions, the parameter subset $[\boldsymbol{\alpha}_{\mathrm{d},0}, \boldsymbol{\alpha}_{\mathrm{d},1}, \boldsymbol{\alpha}_{\mathrm{s},0}, \boldsymbol{\alpha}_{\mathrm{s},1}]$ in \eqref{eqInvariantMappingsParam} can be expressed in dependency of the other remaining parameters. In particular, $\boldsymbol{\alpha}_{\mathrm{d},0}$ is derived by \eqref{eqInvariantMappingsParam1} as a function of $\boldsymbol{\alpha}_{\mathrm{s},M}$, since the phase variable $\xi_{\mathrm{s},1}^-$ can be derived by the kinematic condition when the swing leg touches the ground, which is also in dependency of $\boldsymbol{\alpha}_{\mathrm{s},M}$. Substituting the first row of \eqref{eqInvariantMappingsParam3}---which also corresponds to the expression \eqref{eqInvariantVelocityMappings1}---into the remaining rows gives 
	\begin{equation}\label{eqSwitchLegsCoordBezierRefdtAlpha1}
		\boldsymbol{\alpha}_{\mathrm{d},1} = \boldsymbol{\alpha}_{\mathrm{d},0} + \frac{\xi_{\mathrm{d},1}^{-} - \xi_{\mathrm{d},1}^{+}}{M}  \left(\tilde{\delta}_{\mathrm{s}}^{\mathrm{d}}\right)^{-1} \begin{bmatrix} \mathbf{0}_{2\times 1} & \mathbf{I}_{2} \end{bmatrix} \mathbf{\Delta}_{\mathrm{ref}} \text{~.}
	\end{equation}
	Analogously, $\boldsymbol{\alpha}_{\mathrm{s},0}$ is related to $\boldsymbol{\alpha}_{\mathrm{d},M}$ via equation \eqref{eqInvariantMappingsParam2}, which is then considered in \eqref{eqInvariantMappingsParam4} in order to determine $\boldsymbol{\alpha}_{\mathrm{s},1}$. 
 
    As mentioned in \ref{labSubsectionHybrid}, $\theta_{\mathrm{DSP}}$ is also regarded as a gait parameter, according to which the controller triggers the lift off of the stance leg in order to terminate the DSP. As a conclusion, the independent parameter set that is optimized is comprised of
	\begin{equation}\label{eqParameterSet}
		\boldsymbol{x}_{\mathrm{opt},1} := \begin{bmatrix} \boldsymbol{\alpha}_{\mathrm{d},2}^\top, \dots,\boldsymbol{\alpha}_{\mathrm{d},M}^\top, \boldsymbol{\alpha}_{\mathrm{s},2}^\top, \dots,\boldsymbol{\alpha}_{\mathrm{s},M}^\top, \theta_{\mathrm{DSP}} \end{bmatrix}^\top\text{~.}
	\end{equation}
	
\subsection{Reference of Generalized Momentum}\label{labSubsectionRefOverDsp}
	Section \ref{labSubsectionDSPfullact} and \ref{labSubsectionDSPoveract} introduced the control law in form of the reference trajectory $\zeta_\mathrm{r,d}$ in \eqref{eqControlDspZeta} in order to modify the associated Floquet multiplier. This is also described by an $M$-order Bézier polynomial with the parameters $\boldsymbol{\alpha}_{\zeta}:=[\alpha_{\zeta,0},\dots,\alpha_{\zeta,M}]$---meaning that the reference's boundary values are expressed by $\zeta_\mathrm{r,d}^+=\alpha_{\zeta,0}$ and $\zeta_\mathrm{r,d}^-=\alpha_{\zeta,M}$. Assuming the zeroed control error ($\zeta_\mathrm{d}=\zeta_\mathrm{r,d}$) and considering the periodic condition of the hybrid zero dynamics solution, $\alpha_{\zeta,0}$ is described in dependency\footnote{It is also possible to regard $\alpha_{\zeta,M}$ as dependent to the other parameters in $\boldsymbol{\alpha}_{\zeta}$. The method to determine this dependency according to the periodic condition is the same, thus it is not explicitly discussed again.} of the other parameters in $\boldsymbol{\alpha}_{\zeta}$. Specifically, setting $\zeta_\mathrm{r,d}^-=\alpha_{\zeta,M}$ in \eqref{eqZetaCondition} yields
	\begin{equation}\label{eqZetaPeriodic}
		\alpha_{\zeta,0} = \left(\delta_\mathrm{s}^{\mathrm{d}}\delta_\mathrm{d}^{\mathrm{s}}\right)^2 \alpha_{\zeta,M} + \left(\delta_\mathrm{s}^{\mathrm{d}}\right)^2\mu_\mathrm{s}(\xi_{\mathrm{s},1}^{-}) \text{~.}
	\end{equation}
	
	Hence, the independent parameters
 	\begin{equation}\label{eqParameterZetaSet}
		\boldsymbol{x}_{\mathrm{opt},2} := \begin{bmatrix} \alpha_{\zeta,1},\dots,\alpha_{\zeta,M} \end{bmatrix}^\top
	\end{equation}
    should be also considered in case of the control designs \ref{labSubsectionDSPfullact} and \ref{labSubsectionDSPoveract}. 
	
\subsection{Optimization Framework}\label{labSubsectionFramework}
    The optimization objective is to minimize the energy consumption (evaluated by the cost of transport $CoT$), while the independent gait parameters $\boldsymbol{x}_{\mathrm{opt},1}$ and $\boldsymbol{x}_{\mathrm{opt},2}$ are simultaneously optimized by means of a constrained optimization problem. 
    
    Assuming that the actuators are electric motors with negligible Ohmic heat losses ($I^2R=0$), the supplied energy $E_\mathrm{supp}$ during the entire step period is equivalent to the actuator's mechanical work\footnote{It is also assumed that no generated electrical energy can be reused while the motor works in its generator mode, i.\,e.\ decelerating the movement. So the negative mechanical power is neglected: $P = \max(0,\,P_\mathrm{mech})$.}, therefore the optimization has the objective 
	\begin{equation}\label{eqSuppEnergy}
		f(\boldsymbol{x}_{\mathrm{opt}}) := CoT = \frac{\sum_{i=1}^{4} \int_{0}^{t_\mathrm{step}} \max \left(0,\,P_{\mathrm{mech},i}\right) \mathrm{d} t}{\ell_{\mathrm{step}}\cdot m g} \text{~,}
	\end{equation}
	with the total time $t_\mathrm{step}=t_\mathrm{d}+t_\mathrm{s}$ combined by \eqref{eqIntTimeDsp} and \eqref{eqIntTimeSsp}. The mechanical power $P_{\mathrm{mech},i}=u_i \cdot \dot{q}_{\mathrm{b},i}$ with $i\in[1,4]$ is evaluated by the motor torque $u_i$ in the i-th joint multiplied with its velocity $\dot{q}_{\mathrm{b},i}$. Rather than studying the real time controller's performance against external perturbations, the optimization is conducted in an off-line process, which primarily aims on generating highly efficient gaits. Thus no control error is considered in the simulation. The motor torque $\mathbf{u}$ in the SSP is determined from \eqref{eqBMotorsspZd}; depending on the DSP controller, $\mathbf{u}$ is defined by either \eqref{eqBMotordspZd}, \eqref{eqBMotordspZdFull} or \eqref{eqBMotordspOverZeroErr}. 
	
	The first equality constraint enforces the target average velocity $v$ via
	\begin{equation}\label{eqEqconstraint}
		g_1 := v - \frac{\ell_{\mathrm{step}}}{t_\mathrm{step}} = 0 \text{~.}
	\end{equation}
    The orthogonality of the projection matrix $\mathbf{P}_\mathrm{u}$ in \eqref{eqProjectActuatorUnder} is ensured by the equality constraints
	\begin{equation}\label{eqEqconstraintProjUnder}
		\begin{aligned}
		g_2 &:= \lVert\mathbf{P}_\mathrm{u,1}\rVert_2-1 = 0 \text{~,} \qquad g_3 := \lVert\mathbf{P}_\mathrm{u,2}\rVert_2-1 = 0 \text{~,}\\
        g_4 &:= \mathbf{P}_\mathrm{u,1} \cdot\mathbf{P}_\mathrm{u,2} = 0 \text{~.}
	   \end{aligned}
    \end{equation}
    Analogously, the projection $\mathbf{P}_\mathrm{f}$ in \eqref{eqProjectActuatorFull} is constrained by
	\begin{equation}\label{eqEqconstraintProjOver}
		\begin{aligned}
		g_5 &:= \lVert\mathbf{P}_\mathrm{f,1}\rVert_2-1 = 0 \text{~,} & g_6 &:= \lVert\mathbf{P}_\mathrm{f,2}\rVert_2-1 = 0 \text{~,}\\
        g_7 &:= \lVert\mathbf{P}_\mathrm{f,3}\rVert_2-1 = 0 \text{~,}  &  g_8 &:= \mathbf{P}_\mathrm{f,1} \cdot\mathbf{P}_\mathrm{f,2} = 0 \text{~,}\\
        g_9 &:= \mathbf{P}_\mathrm{f,1} \cdot\mathbf{P}_\mathrm{f,3} = 0 \text{~,} & g_{10} &:= \mathbf{P}_\mathrm{f,2} \cdot\mathbf{P}_\mathrm{f,3} = 0 \text{~.}
	   \end{aligned}
    \end{equation}
    
	The inequality constraints $\mathbf{h}:=[h_1,\dots,h_{15}]^\top<\mathbf{0}$ include
	\begin{equation}\label{eqSspConstraints1}
		h_1 :=-F_{1,z} \text{~,} \qquad h_2 := |F_{1,x}| - \mu_0\cdot |F_{1,x}| \text{~,}
	\end{equation}
	\begin{equation}\label{eqSspConstraints2}
		h_3 :=-z_2 \text{~,} 
	\end{equation}
	\begin{equation}\label{eqSspConstraints3}
		\begin{aligned}
		h_4 := -\theta_{\mathrm{K}1} \text{~,} \qquad h_5 :=-\theta_{\mathrm{K}2} \text{~,}
		\end{aligned}
	\end{equation}
	in the SSP, 
	\begin{equation}\label{eqDspConstraints1}
		\begin{aligned}
		h_6 &:=-F_{1,z} \text{~,} \qquad h_7:=-F_{2,z} \text{~,} \\
		h_8 &:=|F_{1,x}| - \mu_0\cdot |F_{1,x}| \text{~,}  \\
		h_{9} &:=|F_{2,x}| - \mu_0\cdot |F_{2,x}| \text{~,} 
		\end{aligned}
	\end{equation}
	\begin{equation}\label{eqDspConstraints2}
		h_{10}:=-\theta_{\mathrm{K}1} \text{~,} \qquad h_{11}:=-\theta_{\mathrm{K}2} \text{~,}
	\end{equation}
	in the DSP and 
	\begin{equation}\label{eqDspConstraints3}
		\begin{aligned}
		h_{12} &:=|\hat{F}_{1,x}| - \mu_0\cdot |\hat{F}_{1,z}| \text{~,} \qquad h_{13} :=-\hat{F}_{1,z} \text{~,}\\
		h_{14} &:=|\hat{F}_{2,x}| - \mu_0\cdot |\hat{F}_{2,z}| \text{~,} \qquad h_{15} :=-\hat{F}_{2,z} \text{~,}
		\end{aligned}
	\end{equation}
	for the impact mappings. These can be cataloged as follows: \eqref{eqSspConstraints1}, \eqref{eqDspConstraints1} and \eqref{eqDspConstraints3} represent unilateral contact without slipping\footnote{$\mu_0=0.6$ is assumed as friction coefficient.}; \eqref{eqSspConstraints2} avoids scuffing of the swing leg to the ground; \eqref{eqSspConstraints3} and \eqref{eqDspConstraints2} ensure no hyper-extension in the knee. 
	
	The optimization problem 
	\begin{equation}\label{eqOptProblem}
		\begin{aligned}
		\underset{\boldsymbol{x}_{\mathrm{opt}}}{\text{min}} \quad &f(\boldsymbol{x}_{\mathrm{opt}}) \\
        \text{subject to} \quad &g_i = 0 \text{~,} \\
        \quad &h_j \leq 0 \\
		\end{aligned}
	\end{equation}
    is solved by a sequential quadratic programming algorithm, provided by the software \textit{Artelys Knitro}. The computer algebra system \textit{Maple} is used for modelling and deriving the equations of motion, which are exported into the programming language \textit{Julia} for the numerical evaluation and the optimization. All the related gradients are generated via automatic differentiation using the \textit{Julia} package \textit{ForwardDiff.jl}. 
	
\section{Results}\label{labSectionResult}
    The primary goal of the optimization process in \ref{labSectionNumeric} is to find valid periodic gaits created by different control strategies with a non-instantaneous DSP. Since the optimization objective is to minimize the energy consumption of locomotion, a preliminary efficiency study is carried out in this section for the speed range $\mathbf{v}:=[0.2,\,0.3,\,\dots,\,1.4]^\top \mathrm{m/s}$. The robot model has the height of $0.83 \,\mathrm{m}$ and the weight of $15.04 \,\mathrm{kg}$. Other mechanical parameters of the segments are given in Table \ref{tab1parameter}, where the center of mass position is measured from the hip joint for the upper body and the thighs, and from the knee for the shanks. 
	\begin{table}[ht]
	\begin{center}
	\caption{System parameters of the robot model}
	\label{tab1parameter}
	\begin{tabular}{ c | c  c  c }
	\hline
	& upper body	&	thigh	&	shank\\
	\hline
	moment of inertia (in $\mathrm{kg\, m^2}$)	&	0.04	&	0.03	&	0.01\\
	mass (in $\mathrm{kg}$)						&	7.00	&	2.69	&	1.33\\ 
	length (in $\mathrm{m}$)					&	0.23	&	0.30	&	0.30\\
	center of mass position (in $\mathrm{m}$)	&	0.14	&	0.16	&	0.09\\
	\hline 
	\end{tabular}
	\end{center}
	\end{table}
	
	This section has the following structure: At first, the optimized efficiency at different speeds is presented in \ref{labSubsectionEnergy}. Secondly, the optimum gaits generated by different controllers are discussed in \ref{labSubsectionGait}. Then, the stability of the periodic gait is discussed in \ref{labSubsectionStability}. Finally, simulations validate the control algorithms in \ref{labSubsectionSim}. 
	
\subsection{Energy Consumption}\label{labSubsectionEnergy}
\begin{figure}[ht]
	\centering
	\includegraphics[width=0.7\linewidth]{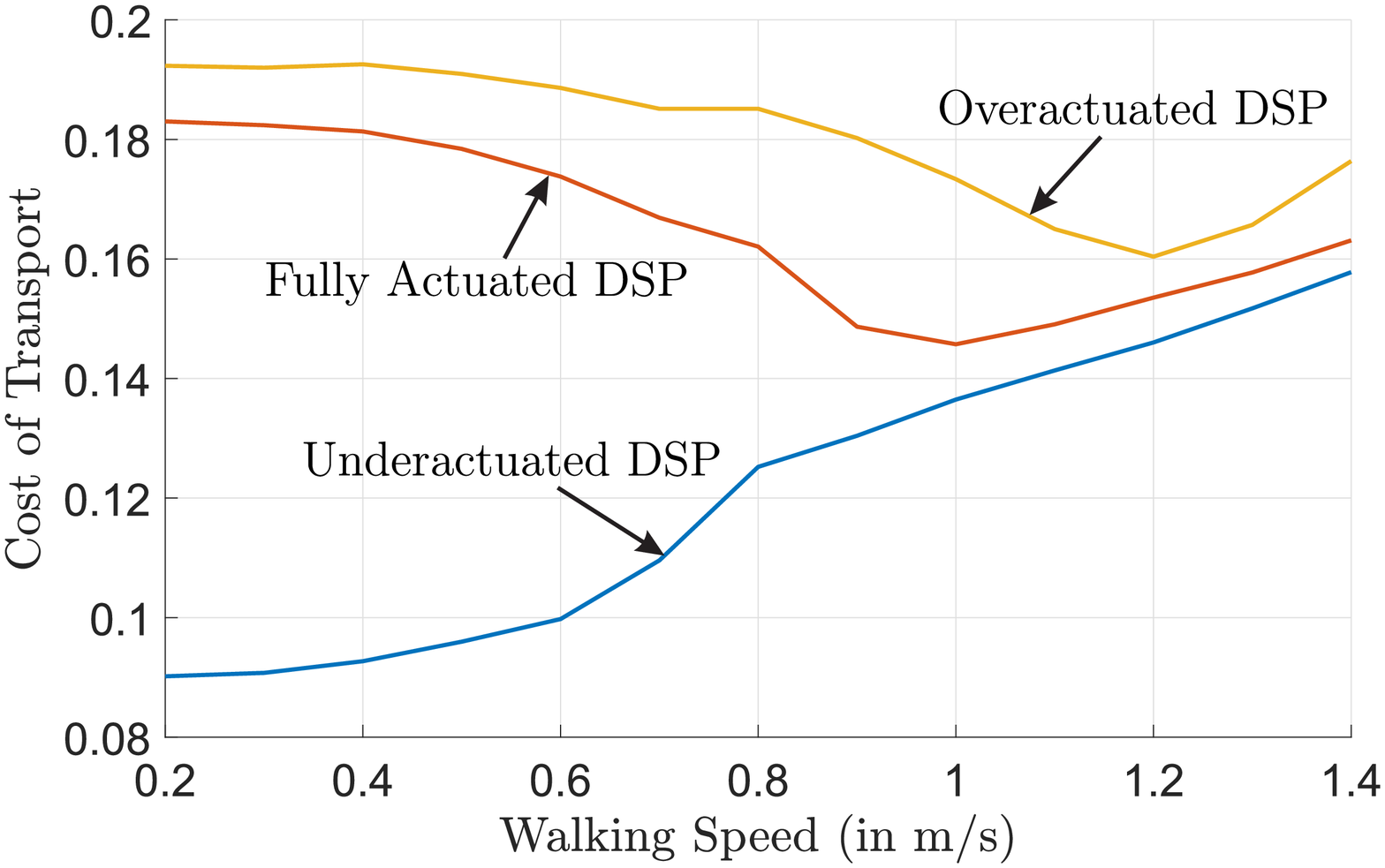}
	\caption{Optimized energy efficiency ($CoT$), resulted from different control strategies, over the speed range $v \in \{0.2,\,0.3,\,\dots,\,1.4\}\,\mathrm{m/s}$}
	\label{figFobj}
	\end{figure}

    Figure \ref{figFobj} shows a comparison of the optimal results for all three proposed controllers with a non-instantaneous DSP. As expected, adding further control objectives increases the energy consumption of the optimal gaits. In other words, the highest efficiency is achieved by creating an underactuated non-instantaneous DSP without any other control objectives; involving more control tasks, such as enforcing the collinearity of the contact forces and modifying the Floquet multiplier, results in the highest energy consumption. Note that none of the four physical actuators is deactivated in all of the controllers. 
 
    In case of the controller with underactuation during the DSP as derived in section \ref{labSubsectionDSPunderact}, the projection \eqref{eqProjectActuatorUnder} solely maps virtual inputs on four physical actuators, whose simultaneous actuation artificially creates an underactuated DSP. Although it would be straightforward and simple to make the DSP underactuated by deactivating some of the physical actuators, the concept of the virtual inputs and their projection provides more flexibility and robustness for creating highly efficient gaits. According to the optimum projection $\mathbf{P}_\mathrm{u}$, the driving torque that is provided from the motor in the trailing leg's knee joint plays a dominant role at all investigated walking speeds; the projection onto the other joint positions strongly depends on the average walking speed. The importance of the knee actuation can be an indication for designing the motor dimension in the knee joint of a real robot prototype, on which the developed controller should be experimentally validated. In the study of \cite{Nody21}\footnote{In the mentioned work, the DSP is modelled as an instantaneous inelastic impact of the swing leg with the ground. At the same time, the former stance leg lifts off without interactions to the environment. Then, the role of the legs is swapped and the next step follows. The efficiency using this model approach has a strong correlation to the speed, meaning that more energy consumption per distance is needed at a larger velocity.} on the other hand, the motor located in the knee joint produces an insignificant torque for the walking gait, which totally differs from the present results. 

\begin{figure}[ht]
	\centering
	\includegraphics[width=0.7\linewidth]{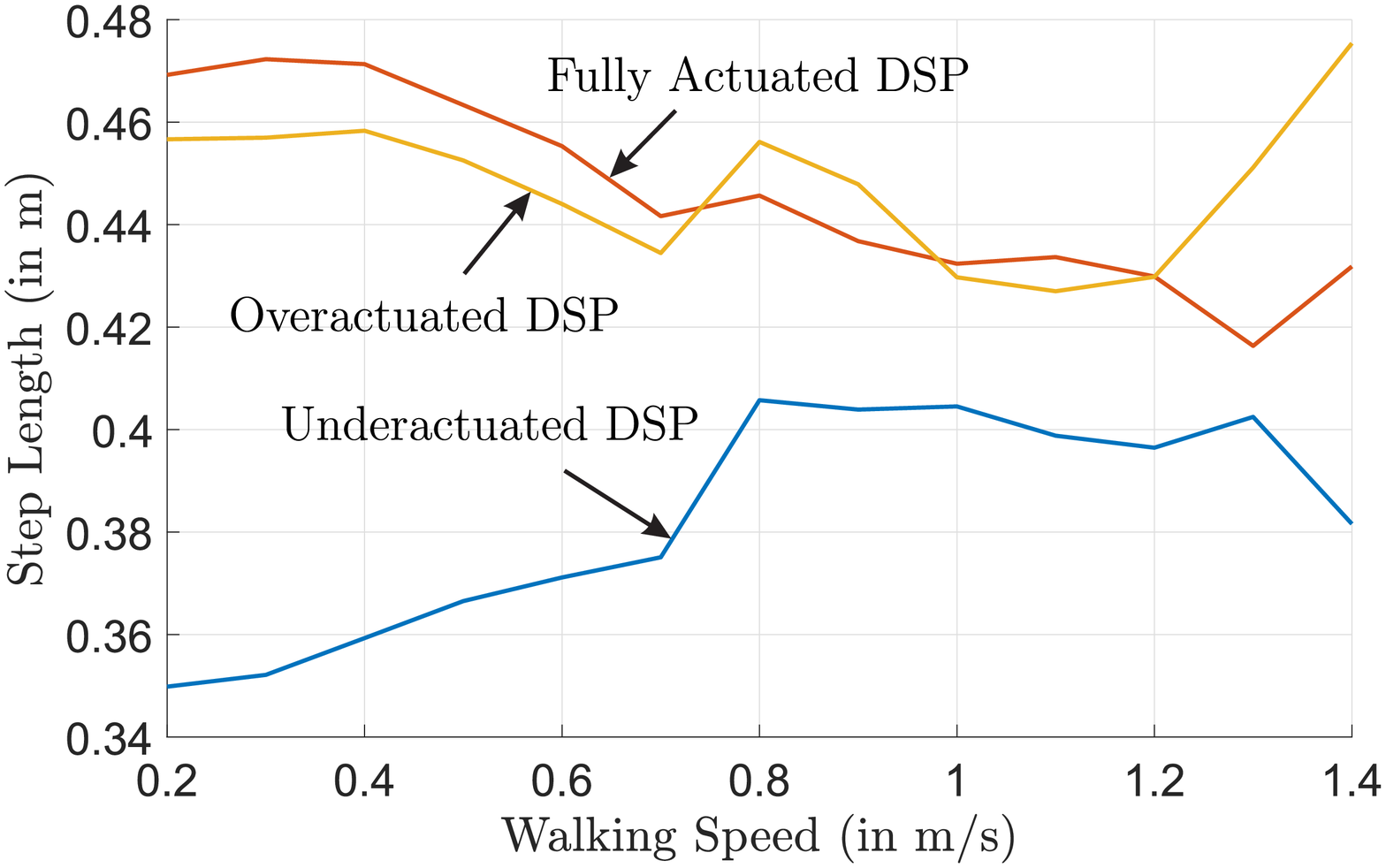}
	\caption{Optimized step length of different controllers}
	\label{figSteplen}
	\end{figure}

    Obviously, the controller associated with the full actuation or the overactuation approach in the DSP shows almost no advantages in terms of the efficiency, mainly due to the absence of the underactuation in the controlled system\footnote{A preliminary parameter study shows that changing the order of the reference trajectory $\zeta_\mathrm{r,d}$ from equation \eqref{eqControlDspZeta} has almost no influence on the resulting $CoT$ in the case of the fully actuated or overactuated controller. In other words, the high efficiency of the underactuated control design is indeed achieved by utilizing the underactuation as well as the passive dynamics. }. The $CoT$ is almost constant for all walking speeds. One of the main reasons is the large optimum step lengths $\ell_{\mathrm{step}}$, which are also almost independent of the walking speed, cf.\ Figure \ref{figSteplen}. 
 
\subsection{Walking Gaits}\label{labSubsectionGait}
\begin{figure}[ht]
	\centering
	\includegraphics[width=0.9\linewidth]{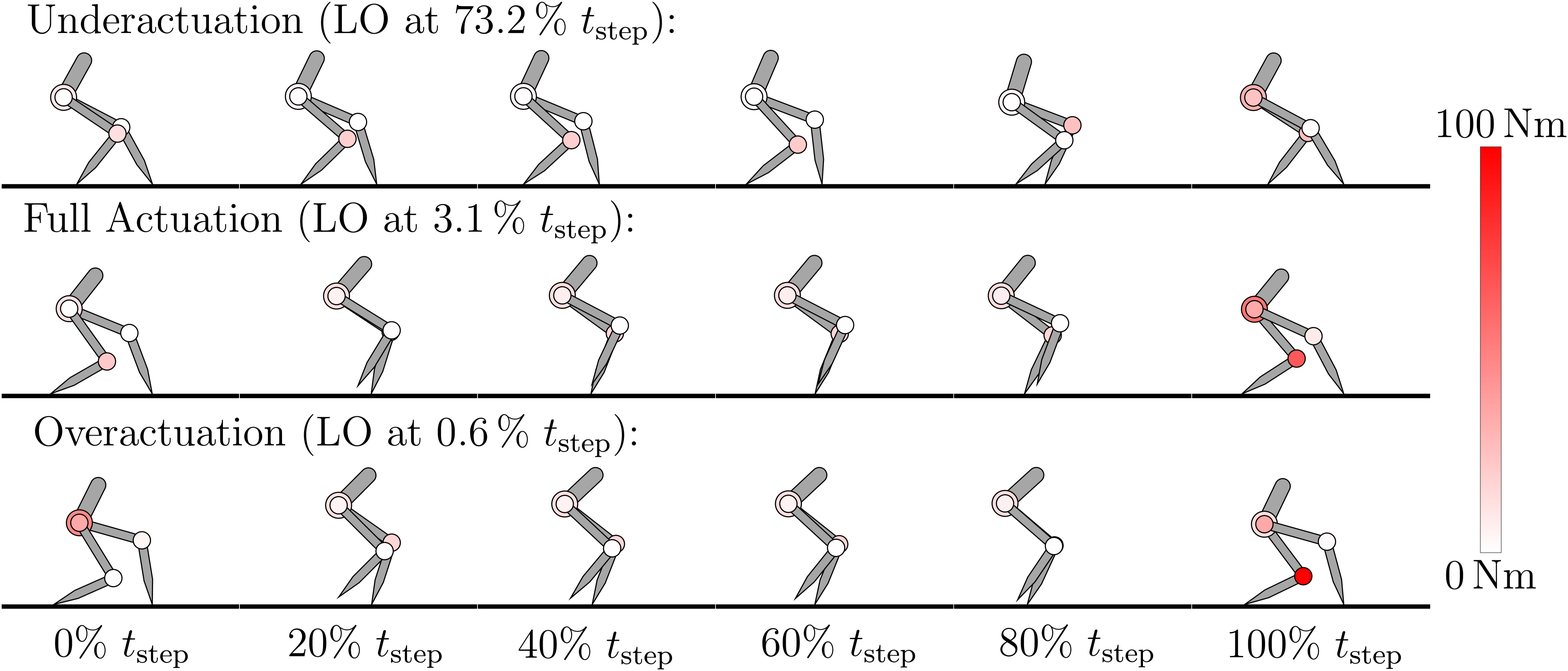}
	\caption{The optimal gait at $0.2\,\mathrm{m/s}$ starting with the DSP followed by the SSP}
	\label{figSnapshots_v020}
	\end{figure}

\begin{figure}[ht]
	\centering
	\includegraphics[width=0.9\linewidth]{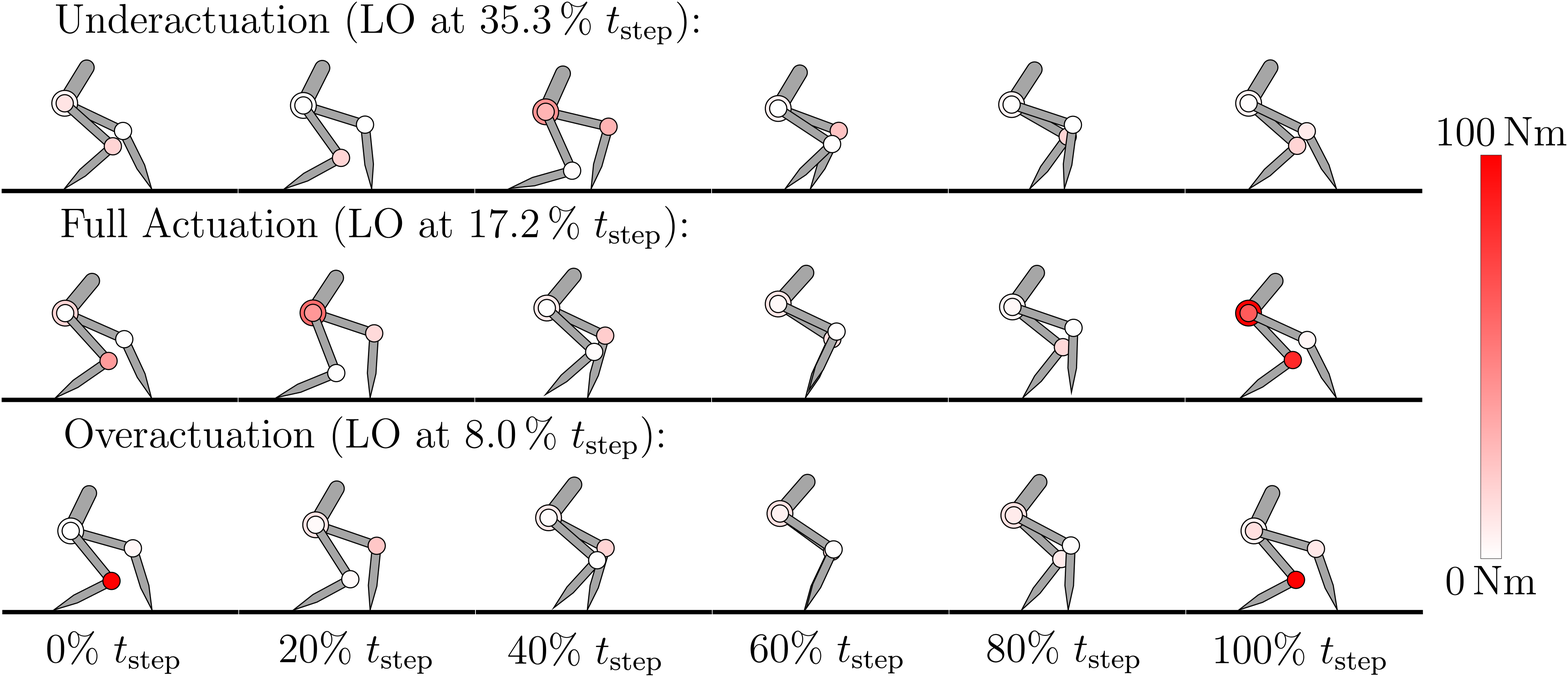}
	\caption{The optimal gait at $0.8\,\mathrm{m/s}$ starting with the DSP followed by the SSP}
	\label{figSnapshots_v080}
	\end{figure}

\begin{figure}[ht]
	\centering
	\includegraphics[width=0.9\linewidth]{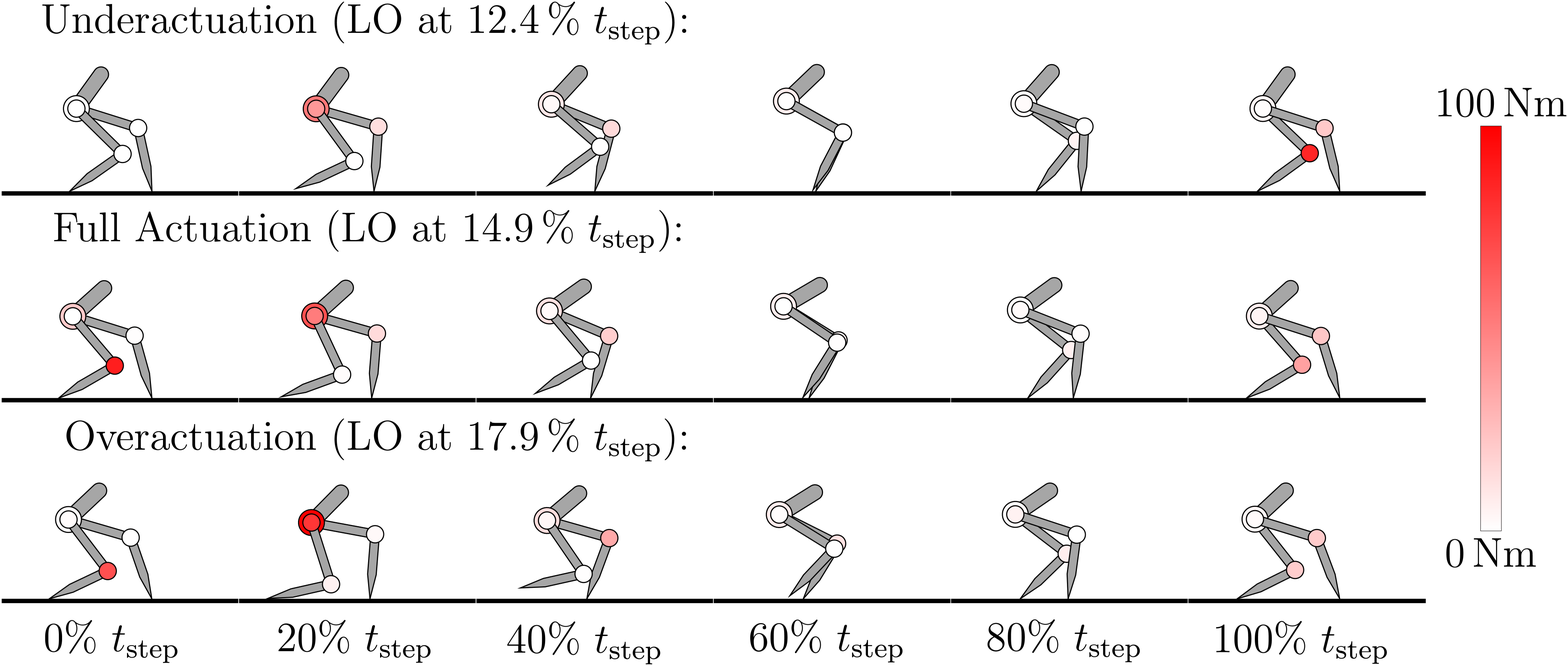}
	\caption{The optimal gait at $1.4\,\mathrm{m/s}$ starting with the DSP followed by the SSP}
	\label{figSnapshots_v140}
	\end{figure}

    The optimized walking gaits are depicted in Figure \ref{figSnapshots_v020}, Figure \ref{figSnapshots_v080} and Figure \ref{figSnapshots_v140} for walking speed $0.2\,\mathrm{m/s}$, $0.8\,\mathrm{m/s}$ and $1.4\,\mathrm{m/s}$, respectively. All results have in common that the upper body tends to tilt forwards with a large angle, which also lowers the center of the mass position. Furthermore, the flexion in the knee joint is large. Indeed, such gaits are caused by the inequality constraints required during the non-instantaneous DSP, especially the non-slipping condition according to \eqref{eqDspConstraints1}. The body configuration with the low center of mass and the bent knee joint makes it simpler for the controller to influence the contact forces at the foot in order to avoid slipping. The necessary actuation from the electric motor for this control task can be in fact observed from the simulated torque, which is indicated by the coloring of the joints. For instance, in the DSP with $v=1.4\,\mathrm{m/s}$, two motors in the trailing leg joints are active in all controllers. Especially in order to fulfill the collinearity of the contact forces in the overactuated controller, peak torques are required in the trailing leg's knee, since the contact force on the back stance foot solely depends on the motor input in the trailing leg, cf.\ \eqref{eqForce2}. 
    
    Furthermore, large actuator torques are required mostly right before the impact at the end of the SSP, where the robot tries to slow down the movement to reduce the impact losses. Also, during the transition from the DSP to the SSP, the actuator needs to create large torques to lift off the trailing stance leg. During the swing phase in the SSP however, the two motors in the swing leg are rarely used, meaning the robot is still able to utilize the passive dynamics to move forwards. The seemingly unnatural gaits are actually the consequence of the model assumption that a point foot is modeled at the lower end of the leg. Humans, on the other hand, have actuated ankle joints and feet, which are able to roll over the ground to ensure the contact conditions during the non-instantaneous DSP with a high efficiency. 
 
\begin{figure}[ht]
	\centering
	\includegraphics[width=0.7\linewidth]{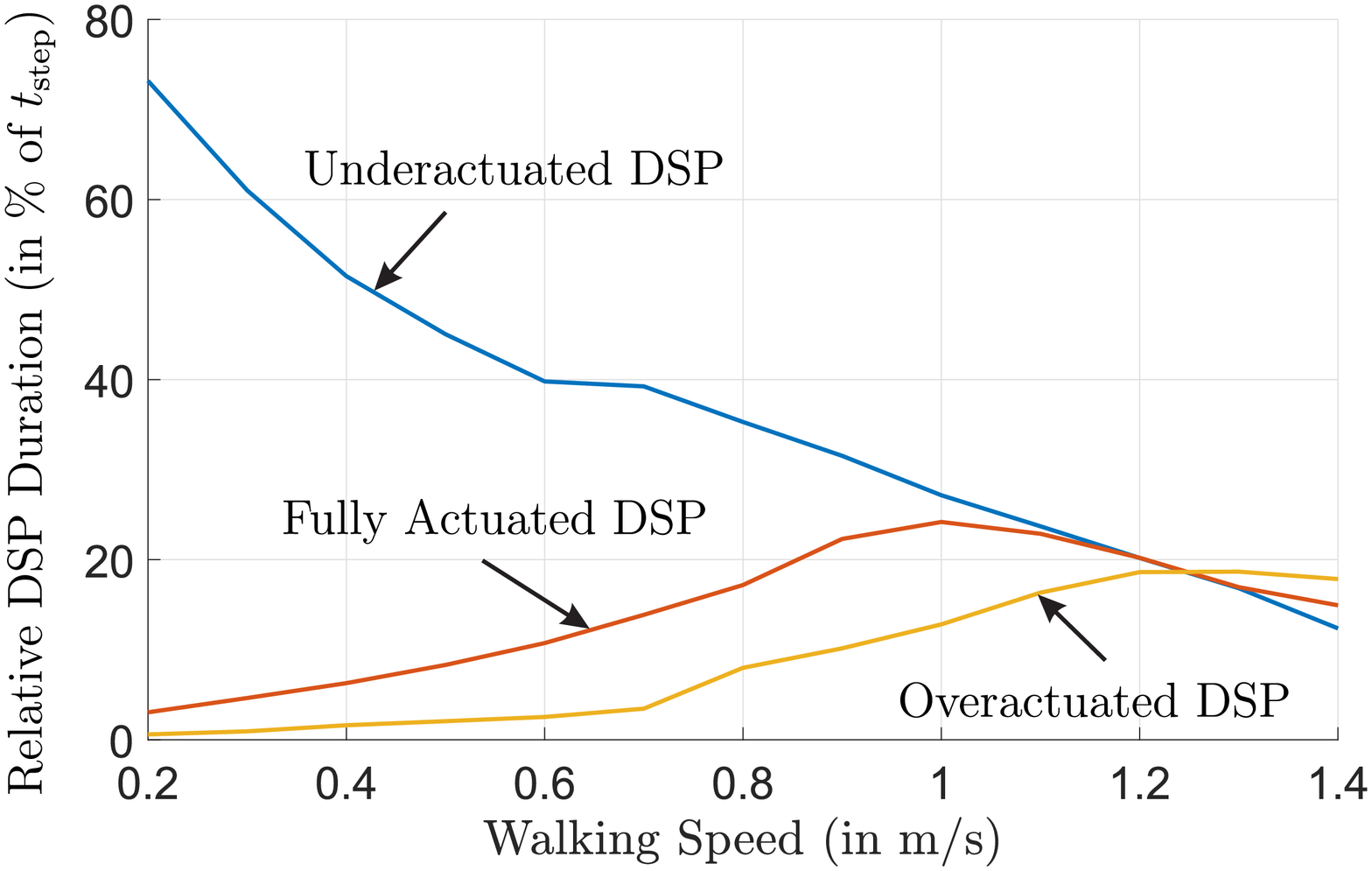}
	\caption{Relative duration of the DSP in a step period}
	\label{figDSPDuration}
	\end{figure}

    Although the optimum step lengths remain within a similar magnitude, the relative duration of the DSP (Figure \ref{figDSPDuration}) in a step period significantly varies depending on the speed and the used controller. The controller with underactuated DSP delays the step utilizing the non-instantaneous DSP to match with a lower speed. While the controller based on the fully actuated or the overactuated DSP actively shapes the generalized momentum, the relative DSP duration is implicitly reduced for slower movements. 
 
\subsection{Stability}\label{labSubsectionStability}
	Besides efficiency, another important aspect to assess the controller's performance is the related stability. In fact this has two perspectives: On the one hand we analyze the stability of the periodic solution that results from a specific HZD controller. In this manner, the solution with any small initial deviations asymptotically converges towards the reference limit cycle if the solution is stable. On the other hand, the study regards the stabilization of the control error due to the PD control during the SSP or DSP, which is necessary to render the previously mentioned asymptotic stability of the periodic solution. This behavior is validated by the closed loop simulation in section \ref{labSubsectionSim}. 
 
\begin{figure}[ht]
	\centering
	\includegraphics[width=0.7\linewidth]{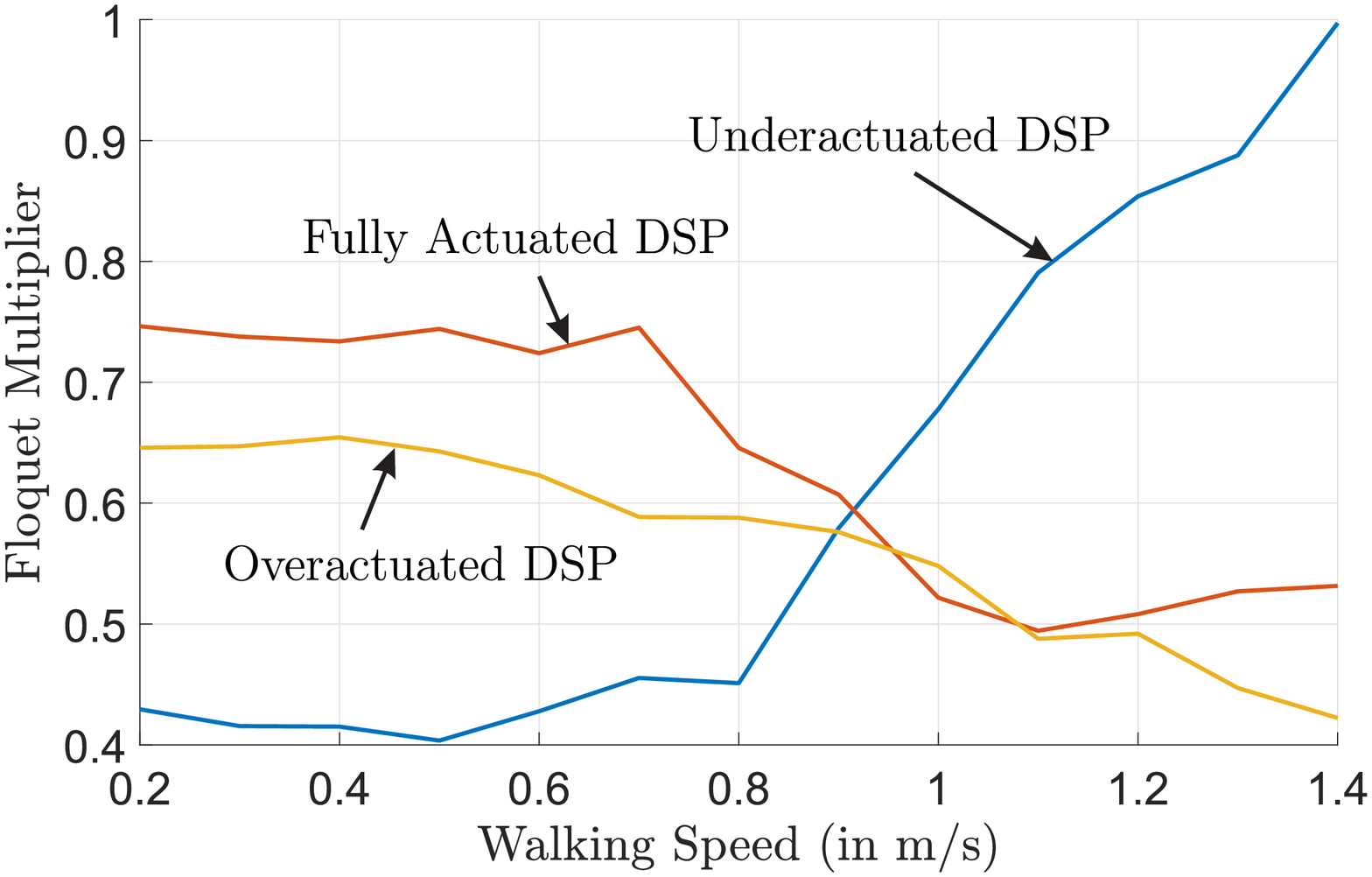}
	\caption{Floquet multipliers resulting from different control strategies}
	\label{figMultiplier}
	\end{figure}

    The first perspective can be evaluated by the Floquet multiplier $\Lambda$, which is determined by means of the Poincaré map method in \ref{labDSPunderactHZDPoincare} and \ref{labDSPoveractHZDPoincare}, as depicted in Figure \ref{figMultiplier}. Here no control error is considered and the feedback gain $K_{\zeta}=1$ is assumed\footnote{Indeed the Floquet multiplier $\Lambda$ could be enforced to be approximately equal to zero, if the control gain $K_{\zeta}$ due to the expression \eqref{eqFloquetFull} is large enough. For the sake of comparison, however, only a small value is considered in this stability study. }. The necessary condition for the asymptotic stability of the limit cycle's periodic solution is $0 < \Lambda < 1$, which is fulfilled in all but one of the investigated scenarios. In comparison to our extended HZD control concepts, the gait with an instantaneous DSP mostly results in a Floquet multiplier $0.8 < \Lambda < 1$, according to the study in \cite{OptimalEcoupling2}. The improved stability property is in fact one of the most important advantages of including a non-instantaneous DSP into the periodic gait. 
	
\subsection{Closed Loop Simulation}\label{labSubsectionSim}
    The optimization study introduced above is based on the assumption that there is no control error in the simulation. This allows us to reduce the entire dynamics of the controlled system to the remaining hybrid dynamics using the reference trajectory. It is a meaningful assumption in the optimization in order to generate highly efficient gaits and the corresponding reference trajectories in each joint. After the gait planning process, it is indispensable to consider disturbances to examine the performance of the feedback controller. For instance, it is barely avoidable that the initial velocity in each joint diverges from its reference at the beginning of a step, due to the impact mapping. In order to maintain the steady walking motion, such deviations must be compensated by the feedback control. Furthermore, one of the most challenging tasks in conducting experiments with a bipedal robot is to set up the proper initial conditions, which include the initial body configuration and the velocity in each joint. The robot usually starts at incorrect initial conditions, and the transient behavior before reaching a steady periodic gait must be stabilized by the feedback. 
    
    Thus this section considers the scenario that the robot starts from rest using different control concepts at the speed of $0.8\,\mathrm{m/s}$. Specifically, the initial velocity of all joints and the absolute angle is set to be zero ($\dot{\mathbf{q}}_\mathrm{d}(t=0) = \mathbf{0}$), the initial angles are however set up correctly according to the reference trajectories. Discontinuous event functions of the ode-solver\footnote{The solver \textit{Tsit5} is a Tsitouras $5/4$ Runge-Kutta method, provided by the \textit{Julia} package \textit{DifferentialEquations.jl}, with absolute tolerance $10^{-9}$.} are used to modify the numerical integration due to the transition events in the hybrid model \eqref{eqHybridSystem}, such as impact and lifting-off one of the stance legs. 

\begin{figure}[ht]
	\centering
	\includegraphics[width=0.7\linewidth]{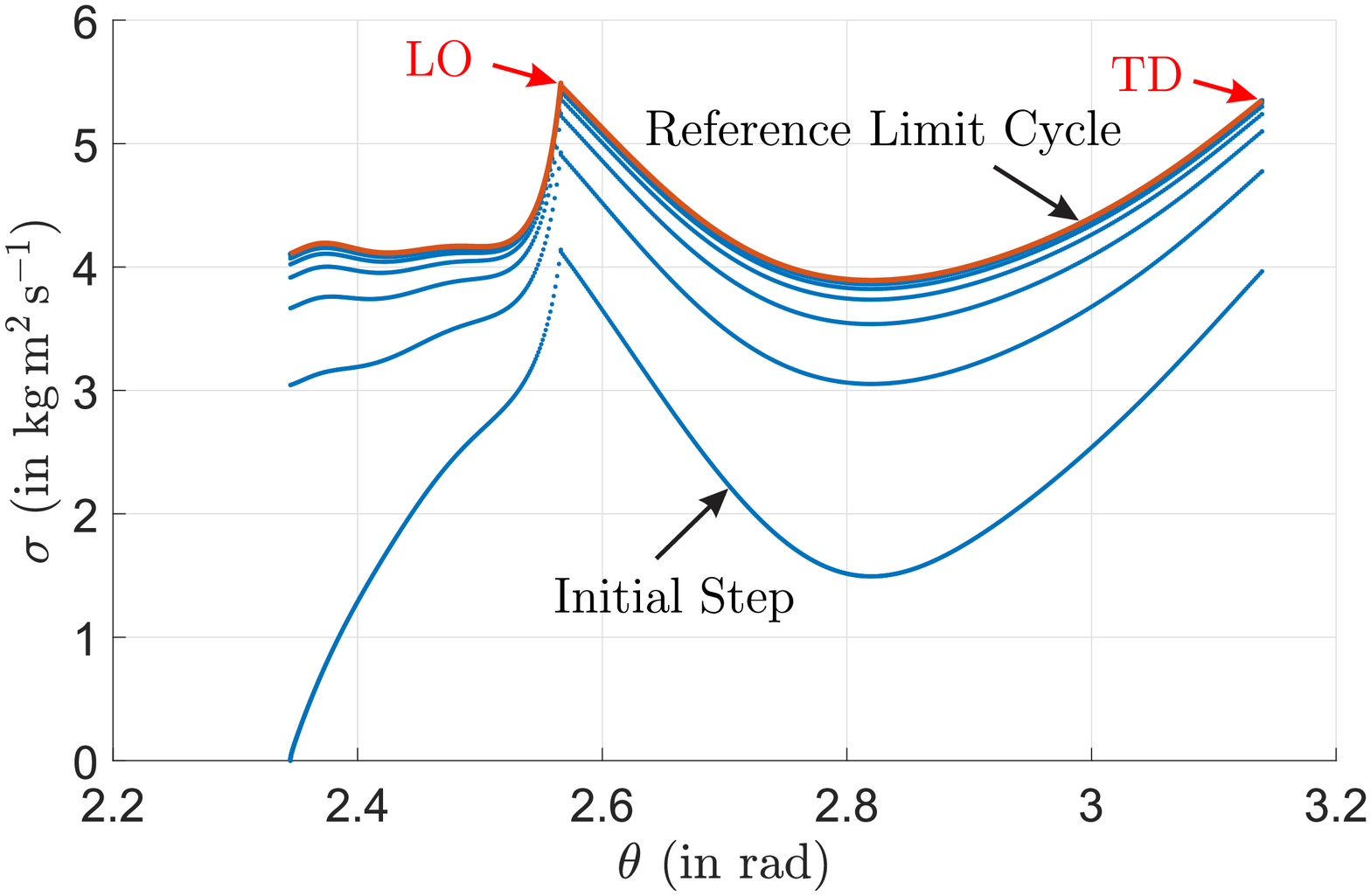}
	\caption{The reference limit cycle (red) and the actual solution (blue) for all steps with the underactuated DSP controller}
	\label{figValidateUnderactuated}
	\end{figure}

    As plotted in Figure \ref{figValidateUnderactuated}, coordinates\footnote{For an unified visualization, the coordinate $\sigma$ according to \eqref{eqMomentumDspOver} is used to display the evolution of all solutions using three different controllers, as $\tilde{\sigma}_\mathrm{d}$ defined in \eqref{eqMomentumDsp} and \eqref{eqMomentumDspFull} can have positive or negative value, depending on the projection of the virtual actuators.} $[\theta, \sigma]^\top$ of the internal dynamics---since the control error is not zero, before the periodic gait reaches the reference limit cycle---start at the initial step with a large deviation from the reference limit cycle of the hybrid zero dynamics. Due to the feedback control, the solution asymptotically converges to the limit cycle after many steps. The PD feedback gains are defined as $\mathbf{K}_\mathrm{P,d} = 10^3\mathbf{I}_2$, $\mathbf{K}_\mathrm{P,s} = 10^3\mathbf{I}_4$, $\mathbf{K}_\mathrm{D,d} = 10^2\mathbf{I}_2$, and $\mathbf{K}_\mathrm{D,s} = 10^2\mathbf{I}_4$. 
 
\begin{figure}[ht]
	\centering
	\includegraphics[width=0.7\linewidth]{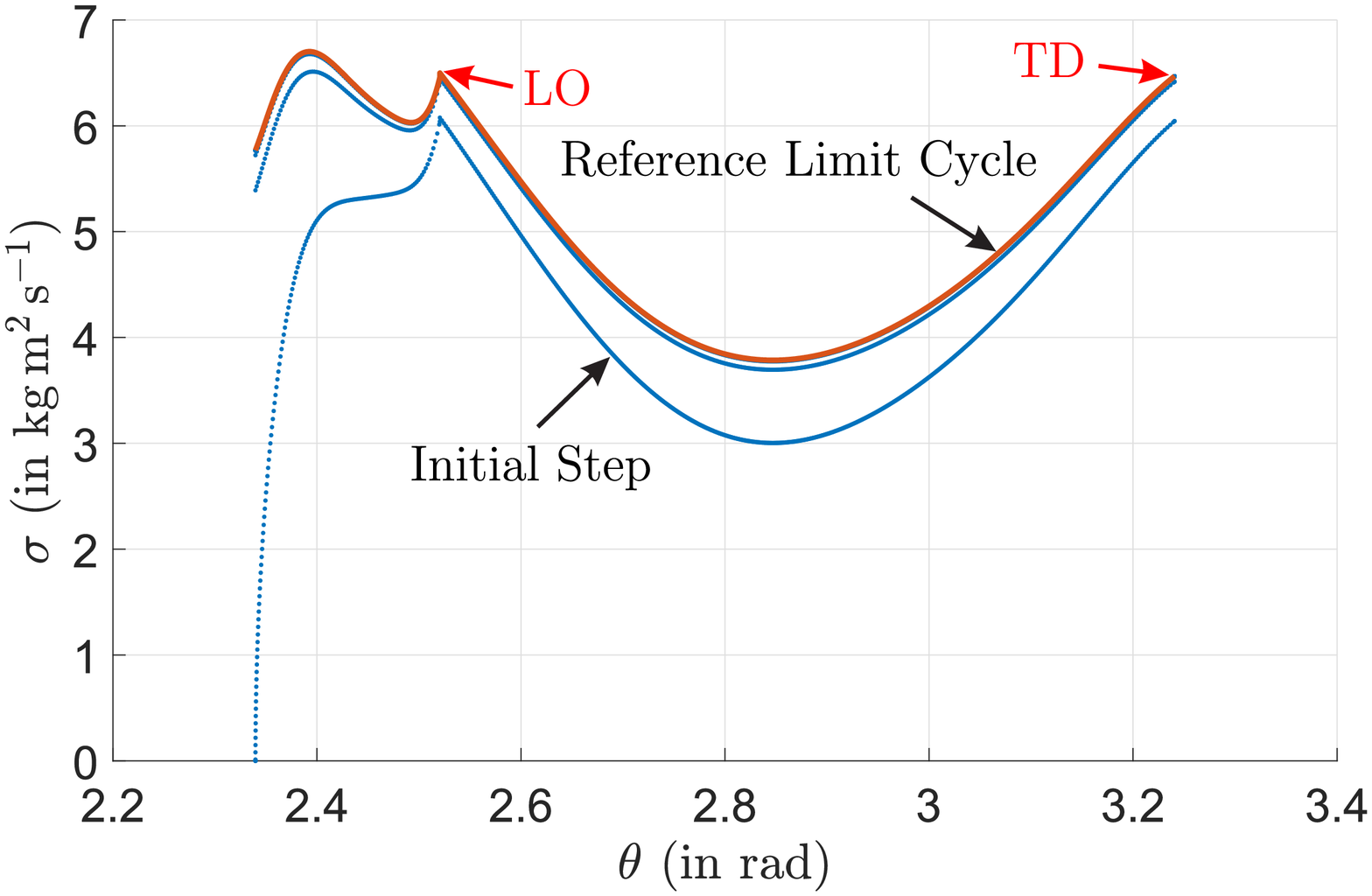}
	\caption{The reference limit cycle (red) and the actual solution (blue) for all steps with the fully actuated DSP controller}
	\label{figValidateFullactuated}
	\end{figure}

    The controller with fully actuated DSP is capable of actively shaping the limit cycle's stability during the DSP, and thus modifying the Floquet multiplier. In order to  demonstrate the controller's capability, the control gain $K_{\zeta}=10$ is used in the simulation in Figure \ref{figValidateFullactuated}, where the corresponding Floquet multiplier is $\Lambda = 0.13$. According to the result, the steady periodic gait is achieved after only two steps starting from rest. 
 
\begin{figure}[ht]
	\centering
	\includegraphics[width=0.7\linewidth]{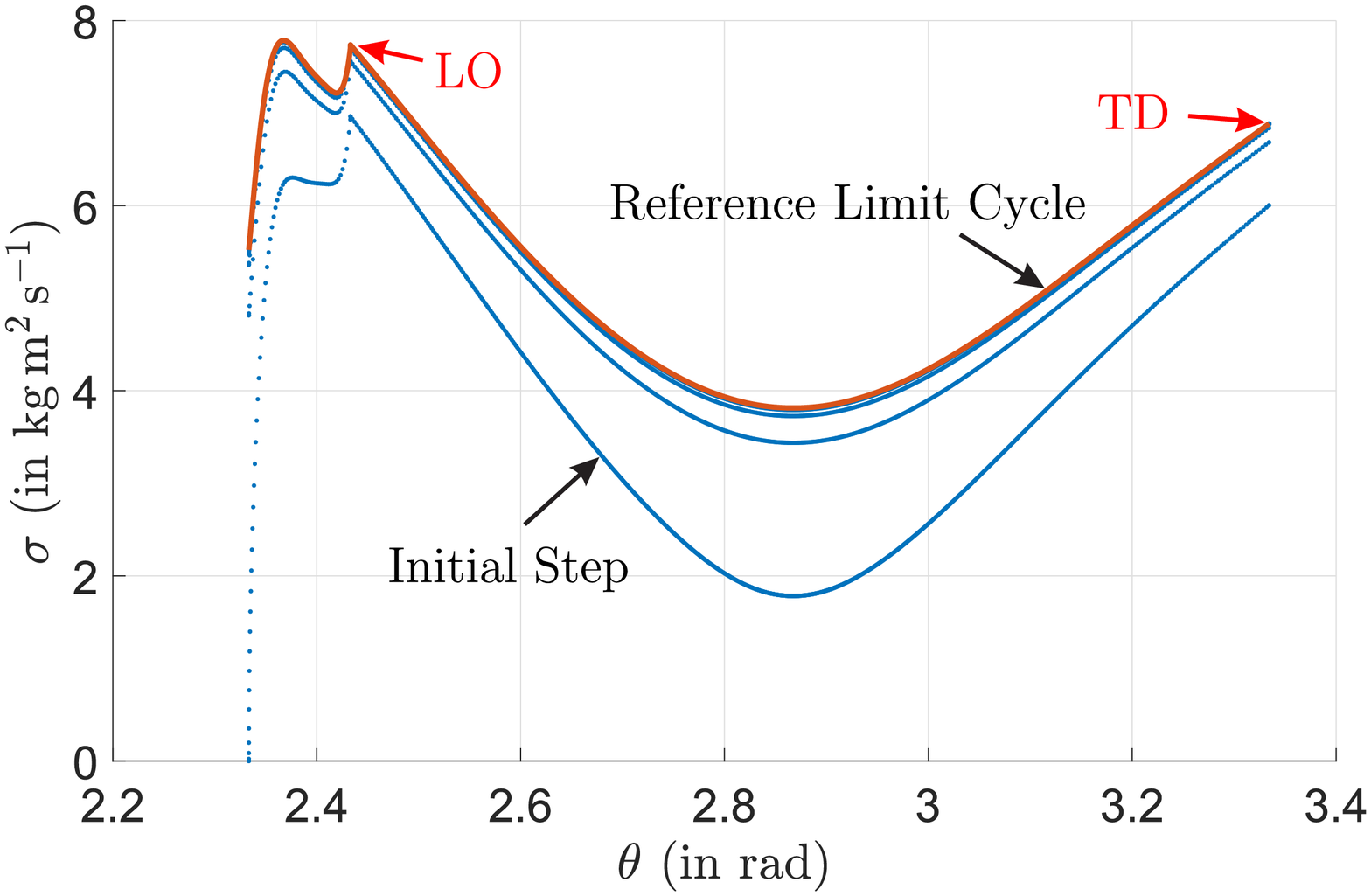}
	\caption{The reference limit cycle (red) and the actual solution (blue) for all steps with the overactuated DSP controller}
	\label{figValidateOveractuated}
	\end{figure}

    Similar performance is found by the control design with overactuated DSP, cf.\ Figure \ref{figValidateOveractuated}, with the control gain $K_{\zeta}=10$ and the Floquet multiplier $\Lambda = 0.24$. The other feedback control gains are also the same as the fully actuated controller. Although a lot of limitations in the experiment with a real robot prototype, such as saturation and delay in the control input or model deviation, are not considered in the simulation, one can still summarize that the extended controller with a non-instantaneous DSP performs well against initial errors. 
		
\section{Conclusion}\label{labSectionConclusion}
    The major goal of the present research is to extend the hybrid zero dynamics control to stabilize bipedal walking gaits with non-instantaneous double support phase. The periodic gait contains two alternating non-instantaneous single and double support phases as well as two discrete transition events between them. In the double support phase, both feet remain on the ground without slipping and the two legs form a closed kinematic chain. At the end of the double support phase, the rear leg lifts off and the system transitions into the single support phase where one foot remains on the ground. At the end of the single support phase, the swing leg foot touches the ground in front of the stance leg, which is modeled as an inelastic impact. Periodic walking gaits are thus sequences of alternating single and double support phases. A nonlinear controller is designed to synchronize each joint angle to a corresponding reference trajectory. The absolute body orientation then corresponds to the remaining (zero) dynamics of the controlled system. Periodic walking gaits can thus be reduced to the limit cycle of these hybrid dynamics, and the solution is efficiently found via numerical optimization. The optimization objective is to minimize the energy consumption of locomotion while optimizing the gait parameters. 

    Since four actuators are available in the non-instantaneous double support phase which has three degrees of freedom, the model in the double support phase is overactuated. We suggest three different control concepts in order to utilize the overactuation to formulate different control tasks. The first controller uses two independent virtual inputs that are projected onto the four physical actuators to artificially create an underactuated double support phase. The second controller uses the projection of three virtual inputs and adds a new control objective to actively stabilize the limit cycle during the double support phase, aiming on improving the gait stability. The last one uses the physical actuators without virtual inputs and considers the stabilization of the limit cycle and the collinearity of the contact forces on the stance feet. 

    According to the optimization at different average walking speeds, the controller with the underactuated double support phase achieves the highest efficiency. The other control concepts have, however, better stability properties. Since a point foot is modelled at the lower end of the leg, the robot is not able to utilize its passive dynamics while walking. Humans have actuated ankle joints and extended feet that roll over the ground and are thus capable of creating more efficient gaits with a non-instantaneous double support phase. 

    In future works, we are going to validate the control design on a real robot prototype and measure the energy consumption. Also, instead of enforcing the collinearity of the contact forces, other control tasks can be investigated in the control design with overactuation, e.\,g.\ to improve the stability in the frontal plane orthogonal to the walking direction, since the robot prototype walks in a three dimensional space. Furthermore, the extended control concept could be also applied for bipedal walking robots that have actuated feet models with more degrees of freedom. 

\begin{appendices}
\section{Geometry in the Double Support Phase}\label{labAppendixClosedLoop}
\begin{figure}[ht]
	\centering
	\includegraphics[width=0.4\linewidth]{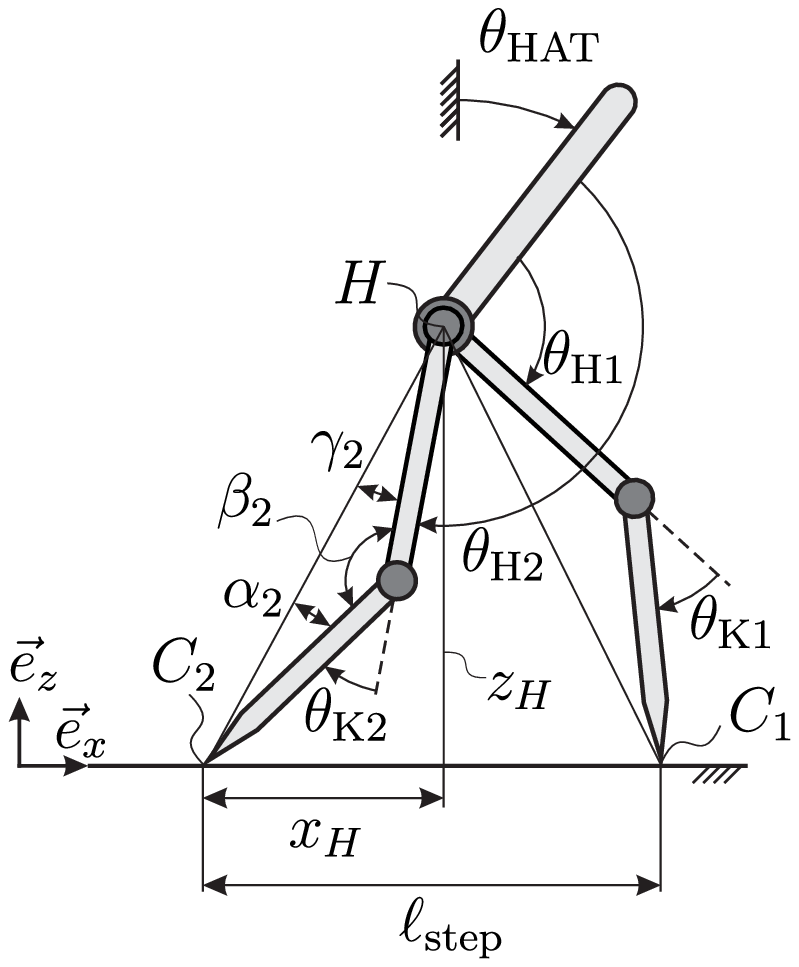}
	\caption{Geometric parameters of the closed kinematic chain formulated by two stance legs during the continuous double support phase}
	\label{figDSPgeometric}
	\end{figure}

    The holonomic constraint $\mathbf{q}_\mathrm{dd} = \hat{\mathbf{\Omega}}(\hat{\mathbf{q}}_\mathrm{d})$ in section \ref{labSubsectionDSP} follows from Figure \ref{figDSPgeometric}.

    The vector $\mathbf{r}_{HC_1}$ from hip to foot 1 is expressed using thigh $\ell_\mathrm{t}$ and shank $\ell_\mathrm{s}$ as
    \begin{equation}\label{eqGeoDSP1}
		\mathbf{r}_{HC_1} = r_{HC_1,x} \vec{e}_x + r_{HC_1,z} \vec{e}_z \text{~,} \\
	\end{equation}
	with 
	\begin{equation}\label{eqGeoDSP2}
	    \begin{aligned}
		r_{HC_1,x} &= \ell_\mathrm{t} \sin\left(\theta_{\mathrm{HAT}} + \theta_{\mathrm{H}1}\right) + \ell_\mathrm{s} \sin\left(\theta_{\mathrm{HAT}} + \theta_{\mathrm{H}1} + \theta_{\mathrm{K}1} \right) \text{~,} \\
		r_{HC_1,z} &= \ell_\mathrm{t} \cos\left(\theta_{\mathrm{HAT}} + \theta_{\mathrm{H}1}\right) + \ell_\mathrm{s} \cos\left(\theta_{\mathrm{HAT}} + \theta_{\mathrm{H}1} + \theta_{\mathrm{K}1}\right) \text{~.}
	    \end{aligned}
	\end{equation}
	With the given step length $\ell_{\mathrm{step}}$, 
	\begin{equation}\label{eqGeoDSP3}
	    \begin{aligned}
		x_{H} &= r_{HC_1,x} - \ell_{\mathrm{step}} \text{~,} \\
		z_{H} &= r_{HC_1,z} \text{~.}
	    \end{aligned}
	\end{equation}
	Thus the length of vector $\mathbf{r}_{HC_2}$ is 
	\begin{equation}\label{eqGeoDSP4}
		|\mathbf{r}_{HC_2}| = \sqrt{x_{H}^2 + z_{H}^2} \text{~.} 
	\end{equation}
	The law of cosines is applied to the triangle in leg 2 yielding
	\begin{equation}\label{eqGeoDSP5}
		|\mathbf{r}_{HC_2}|^2 = \ell_\mathrm{t}^2 + \ell_\mathrm{s}^2 - 2\ell_\mathrm{t}\ell_\mathrm{s}\cos(\beta_2) \text{~.} 
	\end{equation}
	Due to equal segment lengths $\ell_\mathrm{t} = \ell_\mathrm{s}$, 
	\begin{equation}\label{eqGeoDSPBeta}
	    \begin{aligned}
		\beta_2 & = \arccos\left(1 - \frac{|\mathbf{r}_{HC_2}|^2}{2\cdot \ell_\mathrm{t}^2} \right) \text{~,} \\
		\gamma_2 & = \alpha_2 = \frac{\pi - \beta_2}{2} \text{~.}
	    \end{aligned}
	\end{equation}
	Therefore, 
	\begin{equation}\label{eqGeoDSPFinal}
        \hat{\mathbf{\Omega}}(\hat{\mathbf{q}}_\mathrm{d}) = \begin{bmatrix}\theta_{\mathrm{H}2}\\ \theta_{\mathrm{K}2}\end{bmatrix} = \begin{bmatrix}\pi + \arctan\left(\frac{x_{H}}{z_{H}} \right) - \gamma_2 - \theta_{\mathrm{HAT}} \\ 2 \cdot \gamma_2 \end{bmatrix}\text{~.}
	\end{equation}
\end{appendices}

\begin{con}
\ctitle{Author Contributions}
Yinnan Luo, Ulrich J. Römer and Alexander Dyck contributed to the conception, method and analysis of the work. Yinnan Luo implemented and evaluated the simulation. Yinnan Luo wrote the first draft of the article. Ulrich J. Römer revised and improved the first draft. All authors read and approved the final article. 

\ctitle{Financial Support}
This work is supported by the German Research Foundation (DFG), (grant FI 1761/4-1 $|$ ZE 714/16-1).

\ctitle{Conflicts of Interest}
The authors declare no conflicts of interest exist.

\ctitle{Ethical Approval}
Not applicable.
\end{con}


\end{document}